\pdfoutput=1
\documentclass[10pt,twocolumn,letterpaper]{article}

\usepackage[pagenumbers]{paperstyle}

\makeatletter
\patchcmd{\@maketitle}{\vskip .375in}{\vskip 0in}{}{\PackageWarning{arxiv}{title space patch 1 failed}}
\patchcmd{\@maketitle}{\vspace*{24pt}}{\vspace*{12pt}}{}{\PackageWarning{arxiv}{title space patch 2 failed}}
\makeatother

\usepackage{amsmath}
\usepackage{amssymb}
\usepackage{booktabs}
\usepackage{multirow}
\usepackage{algorithm}
\usepackage{algpseudocode}
\usepackage{placeins}

\newcommand{\ksink}{\mathbf{k}_{s}}
\newcommand{\usink}{\mathbf{u}}
\newcommand{\q}{\mathbf{q}}
\newcommand{\kk}{\mathbf{k}}
\newcommand{\methodname}{\textsc{Shrink}\xspace}
\newcommand{\sinkvar}{\textsc{Sink}\xspace}
\newcommand{\allvar}{\textsc{All}\xspace}

\definecolor{linkblue}{rgb}{0.21,0.49,0.74}
\usepackage[pagebackref,breaklinks,colorlinks,allcolors=linkblue]{hyperref}

\usepackage[most]{tcolorbox}
\newtcolorbox{keyquestion}{
    enhanced,
    colback=gray!5,
    colframe=black!50,
    boxrule=0.6pt,
    arc=2pt,
    left=6pt,
    right=6pt,
    top=7pt,
    bottom=5pt,
    before skip=8pt,
    after skip=6pt,
    title={Key Question},
    coltitle=black,
    fonttitle=\bfseries,
    attach boxed title to top left={
        xshift=6pt,
        yshift=-3pt
    },
    boxed title style={
        colback=gray!15,
        colframe=black!50,
        boxrule=0.6pt,
        arc=2pt,
        left=4pt,
        right=4pt,
        top=2pt,
        bottom=2pt
    }
}

\title{Beyond the Sink Token: Sink-Guided Key Editing in Vision-Language Models}
\author{Hui Zhu \quad Saleh Zare Zade \quad Rafi Ibn Sultan \quad Xiangyu Zhou \quad Dongxiao Zhu\\
Wayne State University\\
{\tt\small \{hq2197, hr3331, hm4013, hp6438, ct4442\}@wayne.edu}}
\begin{document}
\maketitle
\begin{abstract}
In vision-language models (VLMs), a few sink tokens attract a disproportionate share of attention. While previous studies have examined why these tokens attract attention, we investigate how the sink-key direction affects attention to image and question tokens. Across four probed VLMs, boosting the magnitude of positive sink-aligned components increases attention on average, whereas boosting negative components decreases it; these responses persist through later layers. This identifies the native sink key as a reference for controlling attention beyond its own position. Motivated by this finding, we propose \methodname, a training-free method that partially reduces sink-aligned components across visible keys during generation. On nine benchmarks under sampled reasoning, it improves mean accuracy by 5.1 and 2.7 percentage points on LLaVA-OneVision-7B and Qwen2.5-VL-3B, with smaller gains on the other two main models, Qwen2.5-VL-7B and Gemma-3-4B. Beyond \methodname, rescaling sink-aligned key components can replace the attention operations of two existing attention-steering methods, GIFT and ACG, with comparable accuracy on hallucination benchmarks.
\end{abstract}

\section{Introduction}
\label{sec:intro}

Vision-language models (VLMs) rely on attention to process images and questions when generating answers~\cite{vaswani2017attention,bai2025qwen25vl}. Yet some heads assign substantial attention to a few \emph{sink tokens}, known as \emph{attention sinks}, regardless of the input~\cite{xiao2024streamingllm}. Previous studies have investigated why these tokens attract attention~\cite{sun2024massive,gu2025emergence}. Attention redistribution can improve VLM predictions~\cite{liu2024pai,kang2025var}, and structural attention biases can extend beyond individual sinks~\cite{spar2026decoupling}. Editing only the sink's score changes its attention share but preserves attention ratios among the other tokens~\cite{han2026zerotuning}. We ask:

\begin{keyquestion}
\centering
Does the sink's influence extend to how image and question tokens compete with one another for attention?
\end{keyquestion}

We investigate this broader influence through the sink's key vector. Self-attention scores tokens by comparing their \emph{keys} with the current token's \emph{query}~\cite{vaswani2017attention}. Prior work links sink attraction to its key~\cite{gu2025emergence}, while analyses of query-key geometry reveal shared directional structure~\cite{godey2024anisotropy}. We study one sink per model, the most-attended template token before the image and usually the first token; its key is identical across inputs. Other keys can contain a part pointing along or against it: a positive or negative \emph{sink-aligned component}.

\begin{figure}[t]
  \centering
  \includegraphics[width=\linewidth]{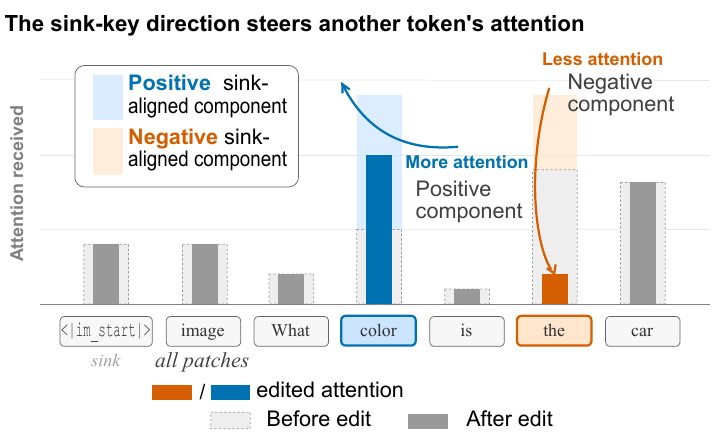}
  \caption{\textbf{Boosting the sink-aligned component.} For a query positively aligned with the sink-key direction, illustrative bars show attention increasing when a positive component is amplified and decreasing when a negative component is amplified.}
  \label{fig:sink}
\end{figure}

We find that the sink-key direction can control how much attention image and question tokens receive. As illustrated in \cref{fig:sink}, boosting the magnitude of a positive sink-aligned component increases attention on average, whereas boosting a negative one decreases it. These responses persist through later layers, extending the sink-key direction's role beyond attraction to the sink itself.

We then ask whether reducing the sink-key direction's contribution improves answer accuracy. Shrinking sink-aligned components makes attention scores depend more on the remaining key components. We introduce \methodname, a training-free decoding method that applies this shrinkage across all visible keys in heads attending to the image or question. Because the components differ across tokens, the edit changes their competition with one another as well as with the sink.

Under sampled reasoning, \methodname improves mean accuracy on the four main VLMs, with gains of $5.1$ and $2.7$ percentage points on LLaVA-OneVision-7B and Qwen2.5-VL-3B. Rescaling sink-aligned key components can also replace the attention operations of GIFT~\cite{qi2026gift} and ACG~\cite{jo2026acg} with comparable accuracy on hallucination benchmarks.

Our contributions are threefold:
\begin{itemize}
    \item We show that existing signed components along the native sink-key direction control attention to image and question tokens. Boosting positive components increases average attention; boosting negative components decreases it, with both effects persisting through later layers.

    \item We introduce \methodname, a training-free intervention that changes relative attention among content tokens by shrinking their sink-aligned components, extending the scope of sink-only editing.

    \item We show that rescaling sink-aligned key components can replace the attention operations of GIFT and ACG, with comparable accuracy on hallucination benchmarks.
\end{itemize}

\section{Related Work}
\label{sec:related}

\paragraph{Origins and function of sinks.}
Studies connect sink attraction to massive activations~\cite{sun2024massive}, representation geometry~\cite{gu2025emergence,cancedda-2024-spectral,ruscio2025sinking,li2026structural}, and circuits~\cite{sun2026spike,ranmilo2026mechanistic}. Related analyses examine concentrated query-key activations~\cite{jin2025massive} and visual re-sinking~\cite{kan2026rar}. Shared geometry connects tokens beyond the primary sink: queries and keys can drift along a common axis~\cite{godey2024anisotropy}, and middle-layer MLPs align secondary-sink hidden states with the primary sink~\cite{wong2026secondary}. SinkRouter places BOS keys near the key mean in PCA~\cite{liu2026sinkrouter}; \mbox{ValueDiff} reports anti-alignment with the content-key mean~\cite{park2026valuediff}, and KeyDiff uses key similarity to the mean for attention prediction and cache eviction~\cite{park2025keydiff}. These measurements motivate our mean-key comparison. Sink representations also serve computational roles: they support dormant heads and stable biases~\cite{guo2024active,chen2026biasvectors}, limit over-mixing~\cite{barbero2025llms}, and enable streaming~\cite{xiao2024streamingllm}, compression~\cite{queipo2026compression}, and visual aggregation~\cite{darcet2024registers,jiang2026vision,luo2026sink}. Catch, Tag, Release studies shared tags carried through sink values~\cite{zhang2026attention}; SinkTrack and OutRo exploit sink information~\cite{liu2026sinktrack,yoo2026outro}, with OutRo finding that zeroing sink-key coordinates can hurt performance. We therefore preserve token values at the edited layer when studying attention attraction.

\paragraph{Steering attention during inference.}
Inference-time methods reweight visual or textual attention~\cite{liu2024pai,yin2025clearsight,chen2025adaptvis,zhang2024pasta} and counter distracting attention~\cite{gong-etal-2024-damro,huang2024opera}. Sink-aware methods calibrate attention~\cite{yu2024act}, edit initial-token logits~\cite{han2026zerotuning}, redistribute visual-sink attention~\cite{kang2025var}, or use sink-triggered~\cite{shukla2026sage} and position-aware decoding~\cite{zhang2026tokenpenalty}. AttnReal redirects attention from output sinks to image tokens~\cite{tu2025attnreal}, while FarSight uses mask-based registers to absorb attention diverted to outliers~\cite{tang2025farsight}. SPAR broadens the focus beyond isolated sinks, correcting textual bias over visual features with an attention template~\cite{spar2026decoupling}. We study this broader competition through the sink-key direction. Native attention also guides where to intervene~\cite{sarkar2025spin,kang2025localization,chen2026havae,tang2025visflow}; we select heads reading the image, question, or options. We also test key edits within GIFT's saliency reweighting and ACG's contrastive guidance~\cite{qi2026gift,jo2026acg}.

\paragraph{Directional editing of keys and queries.}
Directional editing is established in steering and concept removal~\cite{mu2018allbut,li2023iti,rimsky2024steering,arditi2024refusal,ravfogel2020null,belrose2023leace,liu2025reducing,yang2025nullu,qiu2024sea}. Related interventions include learned key scaling~\cite{choi2026sinks}, TAME's query-key eigenspectrum regulation~\cite{tang2025tame}, spectral edits of the query-key weight product~\cite{tiwari2026qksteering}, and cached-state steering~\cite{belitsky2025kv}. Single-channel scaling targets positional bias~\cite{yu2025positionbias}. OrthoRank instead uses sink-relative hidden-state geometry to select tokens~\cite{shin2025orthorank}. Q-Filters projects keys onto a calibrated, context-agnostic dominant query axis to predict attention and rank cache entries, without editing those components~\cite{godey2025qfilters}. SEKA edits highlighted keys using contrastively estimated relevance subspaces; AdaSEKA combines expert projections~\cite{li2026seka}. We build on projection and key-query equivalence as established tools. Concurrent work~\cite{zhang2026borrowed} also studies sink-key alignment and copying, but moves visual-sink components to an image-key median for crop selection. We use the native sink key as the reference, without estimating a query subspace, and test how editing existing signed components changes attention and accuracy.

\begin{figure*}[t]
 \centering
 \includegraphics[width=0.49\textwidth]{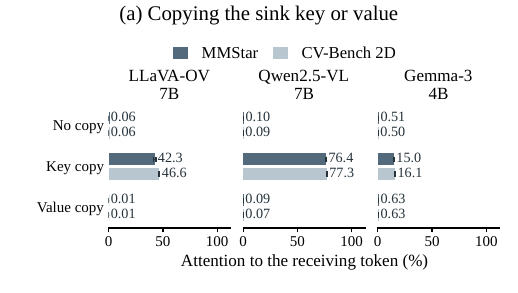}\hfill
 \includegraphics[width=0.49\textwidth]{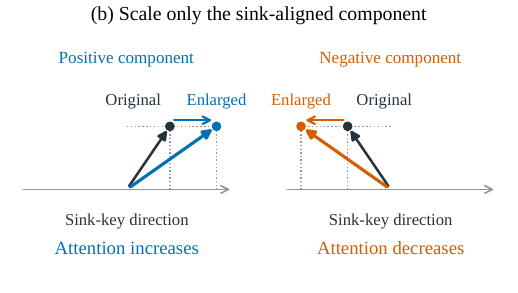}\par
 \includegraphics[width=\textwidth]{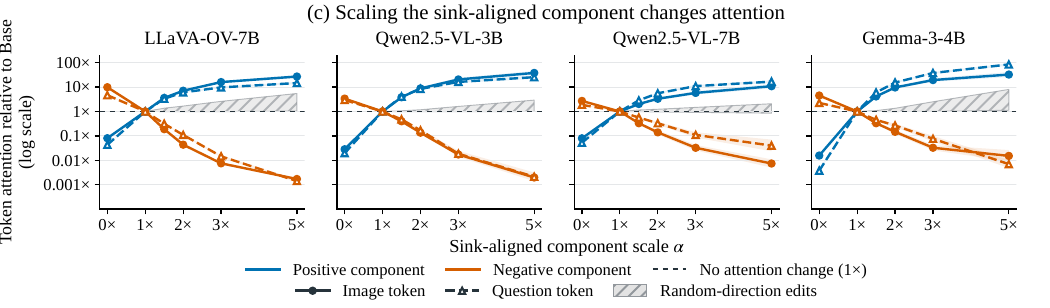}
 \caption{\textbf{Attraction follows the key, and its direction gives signed control.} (a) Attention to a receiving token after copying the sink's key or value, for one model per family; the original sink is blocked. (b) Boosting positive and negative components for a positively aligned query. (c) Attention relative to Base for all four models; multipliers above 1 boost the component's magnitude and 0 removes it. Whiskers and colored bands: 95\% confidence intervals; hatched grey band: range of four size-matched random-direction means.}
 \label{fig:key-control}
\end{figure*}

\section{Finding: Signed Components in Input Keys}
\label{sec:sink-dose}

We test whether sink attraction transfers to another token and whether the sink-key direction can control attention to image and question tokens. \emph{Key copying} replaces only a target token's key with the sink's key; \emph{value copying} replaces only its value with the sink's value. We then isolate the directional contribution by editing the component already present in a target key (\cref{fig:key-control}).

\paragraph{Attention notation.}
To isolate the sink direction's contribution to attention, we separate it from the rest of each key. Within an attention head, $\q$ denotes the current token's query, with its token index omitted for simplicity; $\kk_j$ and $\mathbf v_j$ denote the key and value at position $j$. These vectors are taken after positional encoding. Scores are $z_j=c_l\q^\top\kk_j$ with scale $c_l>0$, and $a_j=\exp z_j/\sum_{i\in\mathcal V_t}\exp z_i$ over visible positions $\mathcal V_t$. Layer and head indices are suppressed unless needed. The nonzero key at sink position $s$ defines the \emph{sink-key direction} $\usink=\ksink/\lVert\ksink\rVert_2$.

An input key's \emph{sink-aligned component} is $(\kk_j^\top\usink)\usink$. Writing its remaining part as $\kk_j^\perp$ separates the score:
\begin{equation}
 z_j=c_l\q^\top\kk_j^\perp+
 \underbrace{(\kk_j^\top\usink)c_l\q^\top\usink}_{\text{sink-aligned contribution}}.
 \label{eq:main-component-score}
\end{equation}
The component can point with or against $\usink$; we call $\kk_j^\top\usink$ its \emph{sink-aligned coefficient}, which is positive or negative accordingly.

\paragraph{Attraction transfers through the key.}
To test whether sink attraction follows the key or value, we copy either vector from the sink token to another prompt token positioned before the image and question. We then measure attention to this receiving token from the final prompt query, before answer generation. The original sink is blocked as an attention target, and a matched no-copy control preserves the receiving token's original key and value. Copying the sink key substantially increases attention to the receiving token across all four models, whereas copying its value leaves attention near the control (\cref{fig:key-control}a). On Qwen2.5-VL-3B, key copying raises it from about 0.02\% to 3.0\% and 5.1\% (Appendix~\ref{sec:supp-transfer}). We next examine the sink-aligned component already present in ordinary input keys.

\paragraph{Boosting sink-aligned components.}
To test directional control without replacing the whole key, we boost the magnitude of a token's existing sink-aligned component while preserving its sign. We edit one image or question token during prompt processing and measure attention to it from later prompt positions. We vary the component's strength using
\begin{equation}
 \kk_j(\alpha)=\kk_j+(\alpha-1)(\kk_j^\top\usink)\usink.
 \label{eq:main-component-scaling}
\end{equation}
A boost uses $\alpha>1$: a positive coefficient becomes more positive, and a negative coefficient becomes more negative (\cref{fig:key-control}b). We also include the unchanged key ($\alpha=1$) and component removal ($\alpha=0$). The remaining key component and value are unchanged at the edited layer.

To test whether the component's sign predicts the attention response, we group measurements by its sign in the unedited key. Figure~\ref{fig:key-control}c reports the geometric mean of edited-to-Base attention ratios, with $1$ indicating unchanged attention. Across all four models, boosting positive components raises the curves above $1$, while boosting negative components lowers them below $1$. Removing the component reverses this pattern. Equal-size random edits orthogonal to $\usink$ separate the two groups much less, indicating that the signed response is tied to the sink-key direction.

\paragraph{Why boosting can raise or lower attention.}
To explain the two responses in Figure~\ref{fig:key-control}b, we hold the query and all other keys fixed. Boosting changes the target's score by
\[
 \Delta z_j
 = (\alpha-1)c_l\,
   (\kk_j^\top\usink)(\q^\top\usink).
\]
For a boost ($\alpha>1$) and a positively aligned query, the score change has the same sign as the sink-aligned component. A positive component therefore raises the target's score and attention; a negative component lowers both. Negative query alignment reverses these effects.

The calculation predicts the immediate response at the edited layer. The full-pass measurements in Figure~\ref{fig:key-control}c show that the average increase for positive components and decrease for negative components persist through later layers. Most final prompt queries are positively aligned with the sink-key direction. Appendix~\ref{sec:supp-key-boost} gives the alignment statistics, measurement protocol, and full results.

These interventions identify a shared sink-related contribution to attention among image and question tokens. Reducing it across keys can adjust their competition while preserving the remaining key components, which motivates \methodname.

\section{Method}
\label{sec:method}

\methodname reduces the shared sink-related contribution identified in \cref{sec:sink-dose} to improve accuracy. It partially shrinks sink-aligned components across visible keys, reducing their contribution to attention scores while preserving the remaining key components at the edited layer. Applying the edit across keys changes competition among content tokens. We focus on heads currently reading the image or question.

\subsection{Preliminaries}
\label{sec:method-preliminaries}

We use the attention notation of \cref{sec:sink-dose} and make head indices explicit. At decoding step $t$, $\mathcal V_t$ contains the visible prompt and generated positions. Visibility follows the model's original attention mask; keys outside this set receive zero attention. Query head $h$ uses query $\q^h$ and the keys $\kk_j^{g(h)}$ and values $\mathbf v_j^{g(h)}$ from its key/value (KV) head $g(h)$. Under grouped-query attention, several query heads can share a KV head. Before editing, attention weights and the head output are
\begin{equation}
 \begin{aligned}
 \mathbf a_h&=\operatorname{softmax}_{j\in\mathcal V_t}
       \!\left(c_l(\q^h)^\top\kk_j^{g(h)}\right),\\
 \mathbf o_h&=\sum_{j\in\mathcal V_t}a_{h,j}\mathbf v_j^{g(h)},\\
 \mathbf o&=\operatorname{Concat}_h(\mathbf o_h)W_O^l.
 \end{aligned}
 \label{eq:method-prelim-attention}
\end{equation}
Here $W_O^l$ is the layer's output projection. The weights determine how the values are combined. \methodname changes the query-key interaction while keeping these values unchanged at the edited layer.

The sink occupies a fixed prefix position $s$ before the image and question. Causality makes its key independent of those subsequent inputs. Its unit key direction $\usink_g$, defined as in \cref{sec:sink-dose} for each layer and KV head, therefore supplies an input-independent reference. No labels are needed to estimate this direction.
\begin{algorithm}[t]
  \caption{\methodname Attention for One Layer and Query}
  \label{alg:shrink-attention}
  \small
  \algrenewcommand{\algorithmicrequire}{\textbf{Input:}}
  \algrenewcommand{\algorithmicensure}{\textbf{Output:}}
  \begin{algorithmic}[1]
    \Require Pre-edit $\q^h,\kk_j^g,\mathbf v_j^g$; map $g(h)$; directions $\usink_g$.
    \Require Visible positions $\mathcal V_t$; image/question sets $\mathcal I,\mathcal T$; sink $s$.
    \Require Layer $l$, eligible layers $\mathcal L_{\mathrm{edit}}$, answer-query flag $b$;
      strength $\beta\in(0,1]$, scale $c_l$, output map $W_O^l$.
    \Ensure Attention-layer output $\mathbf o$.
    \For{each query head $h$}
      \State $g\gets g(h)$
      \State $\mathbf a_h\gets\operatorname{softmax}_{j\in\mathcal V_t}
        \{c_l(\q^h)^\top\kk_j^g\}$ \Comment{pre-edit attention}
      \If{$b$ and $l\in\mathcal L_{\mathrm{edit}}$ and $s\in\mathcal V_t$}
        \State $m_I\gets\sum_{j\in\mathcal I\cap\mathcal V_t}a_{h,j}$;
          $m_T\gets\sum_{j\in\mathcal T\cap\mathcal V_t}a_{h,j}$
        \If{$m_I\geq0.2$ or $m_T\geq0.2$}
          \State $\kk'_j\gets\kk_j^g-\beta((\kk_j^g)^\top\usink_g)\usink_g$
            for all $j\in\mathcal V_t$
          \State $\mathbf a_h\gets\operatorname{softmax}_{j\in\mathcal V_t}
            \{c_l(\q^h)^\top\kk'_j\}$
        \EndIf
      \EndIf
      \State $\mathbf o_h\gets\sum_{j\in\mathcal V_t}a_{h,j}\mathbf v_j^g$
    \EndFor
    \State \Return $\mathbf o=\operatorname{Concat}_h(\mathbf o_h)W_O^l$
  \end{algorithmic}
\end{algorithm}

\ifdefined\pairedperformancetables\else
\begin{table*}[t]
\fi
  \centering\small
  \renewcommand{\arraystretch}{1.12}
  \setlength{\tabcolsep}{3pt}
  \newcommand{\accdelta}[2]{\ensuremath{#1^{#2}}}
  \caption{\textbf{Main four-model comparison.} Test accuracy (\%) on nine benchmarks under sampled reasoning with strict scoring and validation-selected strengths. Base: unedited decoding; $\Delta$: \methodname minus Base in percentage points, computed before rounding. Bold: higher accuracy in each pair. $\ast$/$\dagger$: gains/losses whose pointwise paired-bootstrap 95\% CI excludes zero. Benchmark averages are equally weighted.}
  \label{tab:main-results}
  \begin{tabular*}{\textwidth}{@{\extracolsep{\fill}}@{}l*{3}{rrc@{\hspace{9pt}}}rrc@{}}
    \toprule
    Benchmark & \multicolumn{3}{c}{\shortstack{LLaVA-OneVision\\7B}} & \multicolumn{3}{c}{\shortstack{Qwen2.5-VL\\3B}} & \multicolumn{3}{c}{\shortstack{Qwen2.5-VL\\7B}} & \multicolumn{3}{c}{\shortstack{Gemma-3\\4B}} \\
    \cmidrule(lr){2-4}\cmidrule(lr){5-7}\cmidrule(lr){8-10}\cmidrule(lr){11-13}
     & Base & \methodname & $\Delta$ & Base & \methodname & $\Delta$ & Base & \methodname & $\Delta$ & Base & \methodname & $\Delta$ \\
    \midrule
    \multicolumn{13}{l}{\textit{Core benchmarks}} \\
    MMBench & 64.2 & \textbf{70.9} & \accdelta{+6.7}{\ast} & 73.8 & \textbf{74.5} & \accdelta{+0.7}{} & \textbf{87.3} & 87.1 & \accdelta{-0.2}{} & \textbf{82.6} & 82.4 & \accdelta{-0.2}{} \\
    POPE & 61.2 & \textbf{63.1} & \accdelta{+1.9}{\ast} & 80.9 & \textbf{81.9} & \accdelta{+1.0}{\ast} & 86.3 & \textbf{86.5} & \accdelta{+0.2}{} & \textbf{84.3} & 84.2 & \accdelta{-0.1}{} \\
    CV-Bench & 51.2 & \textbf{54.9} & \accdelta{+3.6}{\ast} & 58.2 & \textbf{61.0} & \accdelta{+2.8}{\ast} & 73.7 & \textbf{74.2} & \accdelta{+0.4}{} & 59.4 & \textbf{60.8} & \accdelta{+1.5}{\ast} \\
    EmbSpatial & 51.1 & \textbf{53.0} & \accdelta{+1.9}{\ast} & 47.3 & \textbf{49.4} & \accdelta{+2.1}{\ast} & 66.7 & \textbf{66.9} & \accdelta{+0.2}{} & 53.9 & \textbf{55.0} & \accdelta{+1.2}{\ast} \\
    MMStar & 38.1 & \textbf{43.3} & \accdelta{+5.2}{\ast} & 44.7 & \textbf{45.4} & \accdelta{+0.7}{} & 59.4 & \textbf{60.7} & \accdelta{+1.3}{} & 52.4 & \textbf{52.5} & \accdelta{+0.1}{} \\
    \midrule
    \multicolumn{13}{l}{\textit{Spatial relations}} \\
    VSR & 42.6 & \textbf{53.4} & \accdelta{+10.8}{\ast} & 63.8 & \textbf{69.1} & \accdelta{+5.3}{\ast} & 84.0 & \textbf{85.2} & \accdelta{+1.2}{\ast} & \textbf{67.6} & 67.4 & \accdelta{-0.2}{} \\
    WhatsUp & 68.9 & \textbf{71.5} & \accdelta{+2.6}{\ast} & 67.2 & \textbf{74.5} & \accdelta{+7.4}{\ast} & \textbf{88.7} & 88.2 & \accdelta{-0.5}{} & 58.8 & \textbf{64.7} & \accdelta{+5.9}{\ast} \\
    COCO-QA & 37.5 & \textbf{42.1} & \accdelta{+4.6}{\ast} & 56.2 & \textbf{57.2} & \accdelta{+1.0}{} & \textbf{64.8} & 64.6 & \accdelta{-0.2}{} & 50.6 & \textbf{51.6} & \accdelta{+0.9}{\ast} \\
    VG-QA & 34.7 & \textbf{43.0} & \accdelta{+8.3}{\ast} & 51.3 & \textbf{54.8} & \accdelta{+3.5}{\ast} & 63.3 & \textbf{65.7} & \accdelta{+2.4}{\ast} & \textbf{48.0} & 46.5 & \accdelta{-1.5}{\dagger} \\
    \midrule
    Core-5 avg. & 53.2 & \textbf{57.1} & \accdelta{+3.9}{\ast} & 61.0 & \textbf{62.4} & \accdelta{+1.5}{\ast} & 74.7 & \textbf{75.1} & \accdelta{+0.4}{} & 66.5 & \textbf{67.0} & \accdelta{+0.5}{\ast} \\
    Spatial-4 avg. & 45.9 & \textbf{52.5} & \accdelta{+6.6}{\ast} & 59.6 & \textbf{63.9} & \accdelta{+4.3}{\ast} & 75.2 & \textbf{75.9} & \accdelta{+0.7}{\ast} & 56.3 & \textbf{57.5} & \accdelta{+1.3}{\ast} \\
    \addlinespace[2pt]
    All-nine avg. & 49.9 & \textbf{55.0} & \accdelta{+5.1}{\ast} & 60.4 & \textbf{63.1} & \accdelta{+2.7}{\ast} & 74.9 & \textbf{75.5} & \accdelta{+0.5}{\ast} & 62.0 & \textbf{62.8} & \accdelta{+0.8}{\ast} \\
    \bottomrule
  \end{tabular*}
\ifdefined\pairedperformancetables\else
\end{table*}
\fi

\subsection{All-Key Shrinkage}

For each eligible query head (Sec.~\ref{sec:method-head-selection}), \methodname applies the following edit, with head indices omitted:
\begin{equation}
 \kk'_j=\kk_j-\beta(\kk_j^\top\usink)\usink,
 \quad j\in\mathcal V_t,\quad 0<\beta\leq1.
 \label{eq:key-projection}
\end{equation}
The fraction removed is $\beta=1-\alpha$ in \cref{eq:main-component-scaling}; $\beta=1$ removes it entirely. Validation selects the strength. The remaining key component and value are unchanged at the edited layer.

Applying the edit to every visible key, including the sink, input, and generated keys, has an equivalent query-side interpretation:
\begin{equation}
 \q'^h=\q^h-\beta((\q^h)^\top\usink_{g(h)})\usink_{g(h)}.
 \label{eq:method-query-projection}
\end{equation}
It reduces the query's interaction with the sink-key direction at every visible position. The resulting change in relative attention is
\begin{equation}
 \Delta\log\frac{a_j}{a_i}
 =-\beta c_l(\q^\top\usink)((\kk_j-\kk_i)^\top\usink).
 \label{eq:method-attention-ratio}
\end{equation}
The softmax denominator cancels when comparing two input tokens, so a sink-only edit preserves their attention ratio. \Cref{eq:method-attention-ratio} shows why all-key shrinkage changes it: keys with different sink-aligned coefficients receive different score changes. For a positively aligned query, a token with a smaller sink-aligned coefficient gains attention relative to one with a larger coefficient; for a negatively aligned query the order reverses. This allows \methodname to adjust competition among image and question tokens alongside their competition with the sink. Appendix~\ref{sec:supp-method} gives the derivation. By the query-side form, the edit itself needs only one projection per eligible query head and leaves the KV cache unchanged.

\subsection{Head Selection and Generation}
\label{sec:method-head-selection}

We focus the edit on heads already reading the task inputs. Let $\mathcal I$ contain image positions and $\mathcal T$ question and answer-option positions. A head qualifies when at least $20\%$ of its pre-edit attention goes to visible image positions \emph{or} at least $20\%$ goes to visible question or option positions. The image threshold follows VAR's initial image-attention filter~\cite{kang2025var}, with related attention-based head selection in SPIN~\cite{sarkar2025spin}.

A label-free calibration pass over a small set of unedited prompts fixes the sink position and eligible layers once per model. During generation, only head eligibility is recomputed at each step.

Editing begins with the query predicting the first generated token; earlier prompt positions are left unedited. At each step, eligible heads recompute attention with projected keys and the original values, skipping the edit if the sink is not visible. Query heads sharing a KV head use the same direction but are assessed independently. Algorithm~\ref{alg:shrink-attention} summarizes the procedure; Appendix~\ref{sec:supp-method} specifies calibration, layer schedules, and implementation.

\flushbottom
\section{Experiments}
\label{sec:experiments}

We evaluate \methodname's accuracy, test directional key edits within GIFT~\cite{qi2026gift} and ACG~\cite{jo2026acg}, identify what \methodname's gains depend on, and examine the resulting attention allocation.

\subsection{Experimental Setup}
\label{sec:exp-setup}

\paragraph{Models and benchmarks.}
We evaluate LLaVA-OneVision-7B~\cite{li2024onevision}, Qwen2.5-VL-3B/7B (Qwen-3B/7B)~\cite{bai2025qwen25vl}, and Gemma-3-4B~\cite{gemmateam2025gemma3}; InternVL3-8B~\cite{zhu2025internvl3} results appear in Appendix~\ref{sec:supp-full-accuracy}. Table~\ref{tab:main-results} reports the four models on Core-5 (MMBench-EN~\cite{liu2024mmbench}, POPE~\cite{li2023pope}, CV-Bench~\cite{tong2024cambrian}, EmbSpatial-Bench~\cite{du2024embspatial}, MMStar~\cite{chen2024mmstar}) and Spatial-4 (VSR~\cite{liu2023vsr}, WhatsUp, COCO-QA, VG-QA~\cite{kamath2023whatsup,chen2025adaptvis}). Additional benchmark results are reported in the appendix.

\paragraph{Decoding and scoring.}
Base denotes unedited decoding. Base and \methodname use identical reasoning-then-answer prompts, three paired sampling seeds, temperature $0.7$, and top-$p=1$. We refer to this protocol as sampled reasoning. For each model and benchmark, $\beta$ is chosen by mean validation accuracy and evaluated on a disjoint test split. Accuracy counts only naturally completed responses with a correct, unambiguous final answer; all others count as wrong. Benchmark means are equally weighted. Appendix~\ref{sec:supp-setup} specifies preprocessing, including LLaVA-OneVision's $384\times384$ resizing.

\paragraph{Uncertainty.}
Table markers denote pointwise 95\% paired-bootstrap intervals over test questions, conditioned on the selected strength.

\subsection{Main Results}
\label{sec:main-results}

\methodname improves nine-benchmark mean accuracy on all four main models (\cref{tab:main-results}). LLaVA-OneVision and Qwen-3B gain $5.1$ and $2.7$ percentage points and improve on every benchmark. Qwen-7B and Gemma gain $0.5$ and $0.8$ points, with more variation across tasks. On ten additional benchmarks, LLaVA-OneVision and Qwen-3B again improve on every benchmark, significantly on four and six (Appendix~\ref{sec:supp-additional}). All four mean gains survive Holm correction over the five evaluated models (Appendix~\ref{sec:supp-multiplicity}).

The improvement is strongest on spatial relations: Spatial-4 gains exceed Core-5 gains for every model. LLaVA-OneVision gains $6.6$ versus $3.9$ points across the two groups; Qwen-3B gains $4.3$ versus $1.5$. Spatial relations respond most to the edit, while Core-5 means remain positive for every model.

The gains also persist with deterministic reasoning: greedy Core-5 accuracy rises by $3.7$ and $1.5$ points on LLaVA-OneVision and Qwen-3B (Appendix~\ref{sec:supp-greedy}).
\ifdefined\hosttableinline\else
\begin{table}[!t]
\fi
\centering\small
\renewcommand{\arraystretch}{1.02}
\setlength{\tabcolsep}{0.9pt}
\caption{\textbf{Key edits inside GIFT and ACG.} Accuracy (\%) with greedy yes/no decoding and matched tuning. Bold: higher in each pair; $\ast$/$\dagger$: Key minus tuned original has a 95\% CI above/below zero.}
\label{tab:host_swap}
\begin{tabular*}{\linewidth}{@{\extracolsep{\fill}}@{}p{45.2pt}lrrr@{\hspace{3.5pt}}r@{}}
\toprule
Model & Method & \shortstack{POPE-\\A-OKVQA} & \shortstack{POPE-\\GQA} & AMBER & Avg. \\
\midrule
\multirow{5}{45.2pt}{\shortstack{LLaVA-1.5\\7B}} & Base & 78.35 & 76.08 & 72.41 & 75.62 \\
\cmidrule(lr){2-6}
 & GIFT (tuned) & \textbf{82.26} & \textbf{81.99} & \textbf{79.70} & \textbf{81.32} \\
 & GIFT (Key) & 81.76\rlap{$^{\dagger}$} & 81.27\rlap{$^{\dagger}$} & 77.24\rlap{$^{\dagger}$} & 80.09\rlap{$^{\dagger}$} \\
\cmidrule(lr){2-6}
 & ACG (tuned) & 80.73 & 79.73 & 76.55 & 79.00 \\
 & ACG (Key) & \textbf{81.51}\rlap{$^{\ast}$} & \textbf{80.77}\rlap{$^{\ast}$} & \textbf{77.53}\rlap{$^{\ast}$} & \textbf{79.93}\rlap{$^{\ast}$} \\
\midrule[0.7pt]
\multirow{5}{45.2pt}{\shortstack{LLaVA-1.5\\13B}} & Base & 76.85 & 74.66 & 72.04 & 74.52 \\
\cmidrule(lr){2-6}
 & GIFT (tuned) & 79.12 & 78.62 & 77.15 & 78.30 \\
 & GIFT (Key) & \textbf{79.46} & \textbf{79.18} & \textbf{78.51}\rlap{$^{\ast}$} & \textbf{79.05}\rlap{$^{\ast}$} \\
\cmidrule(lr){2-6}
 & ACG (tuned) & 80.37 & 79.78 & \textbf{76.47} & 78.87 \\
 & ACG (Key) & \textbf{80.92} & \textbf{80.49}\rlap{$^{\ast}$} & 76.21 & \textbf{79.20} \\
\midrule[0.7pt]
\multirow{5}{45.2pt}{\shortstack{BakLLaVA\\7B}} & Base & 76.29 & 74.93 & 68.62 & 73.28 \\
\cmidrule(lr){2-6}
 & GIFT (tuned) & 81.76 & \textbf{81.00} & 76.88 & 79.88 \\
 & GIFT (Key) & \textbf{81.94} & 80.44\rlap{$^{\dagger}$} & \textbf{77.57}\rlap{$^{\ast}$} & \textbf{79.98} \\
\cmidrule(lr){2-6}
 & ACG (tuned) & 79.52 & 78.91 & 75.66 & 78.03 \\
 & ACG (Key) & \textbf{80.42}\rlap{$^{\ast}$} & \textbf{79.71}\rlap{$^{\ast}$} & \textbf{76.46}\rlap{$^{\ast}$} & \textbf{78.87}\rlap{$^{\ast}$} \\
\bottomrule
\end{tabular*}
\ifdefined\hosttableinline\else
\end{table}
\fi

\subsection{Directional Edits within Existing Methods}
\label{sec:host-methods}

We next test directional key edits within existing methods, retaining their selection or guidance. Each variant rescales sink-aligned key components (\cref{eq:main-component-scaling}) with token-specific strengths. \textbf{GIFT (Key)} replaces GIFT's saliency reweighting with saliency-weighted key edits; \textbf{ACG (Key)} replaces ACG's image-masked contrast branch with a key-edited branch. Both use greedy yes/no decoding. For matched tuning, GIFT's own exponent is also retuned on the same validation pools. ACG keeps validation-chosen guidance ($\gamma=2.4$ on 7B, $4$ on 13B/BakLLaVA), tuning only the new branch's strength.

In Table~\ref{tab:host_swap}, five of six key variants have higher benchmark means, three with positive pointwise intervals; GIFT on LLaVA-1.5-7B favors the original by $1.22$ points. Appendix~\ref{sec:supp-host-methods} gives results on further benchmarks. Directional key edits can thus replace the attention operations of both methods with comparable accuracy on hallucination benchmarks.

\begin{figure*}[t]
  \centering
  \includegraphics[width=\textwidth]{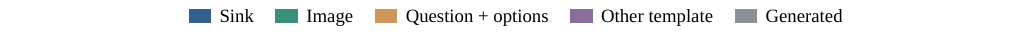}\par
  \includegraphics[width=0.49\textwidth]{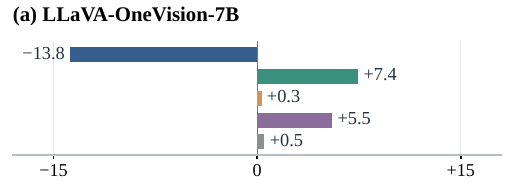}\hfill
  \includegraphics[width=0.49\textwidth]{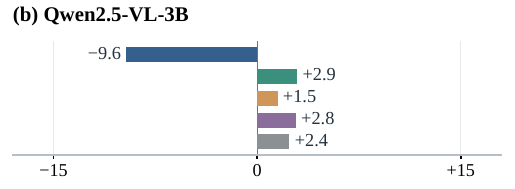}\par
  \includegraphics[width=0.49\textwidth]{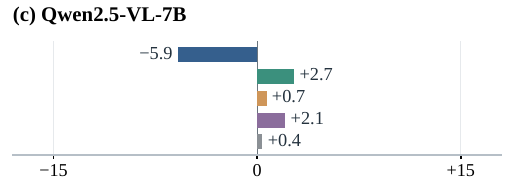}\hfill
  \includegraphics[width=0.49\textwidth]{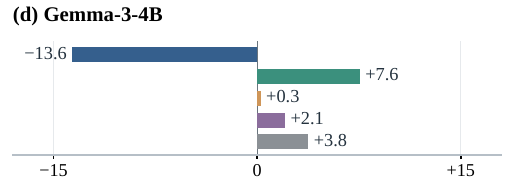}\par
  \includegraphics[width=\textwidth]{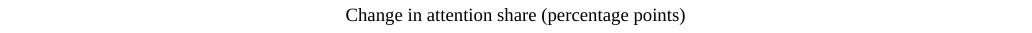}
  \caption{\textbf{Attention changes under \methodname.} Attention leaving the sink goes to image, template, and generated positions. Bars show \methodname minus Base in percentage points on a common scale; the five groups partition visible attention. Both runs follow the same answer continuation, with heads selected separately in each. Means weight the nine benchmarks equally.}
  \label{fig:attention-drift}
\end{figure*}

\subsection{What the Gains Depend On}
\label{sec:exp-analysis}

\paragraph{Editing beyond the sink.}
\methodname changes competition among content tokens. We first compare it with sink-only editing, which preserves their attention ratios. With head and layer selection fixed and strengths tuned separately for each scope, all-key editing gains $3.9$ Core-5 points on LLaVA-OneVision versus $2.4$ for sink-only editing. It is also favored on Qwen-3B ($1.5$ vs.\ $0.6$) and Gemma ($0.5$ vs.\ $-0.3$), with the 95\% CI of the difference above zero on all three models; the two scopes perform similarly on Qwen-7B (Appendix~\ref{sec:supp-breadth}).

\paragraph{The reference direction.}
We next test whether the gains depend on the reference direction, fixing scope, head and layer selection, and one model-level strength, in a separate run (Table~\ref{tab:direction-controls}; Appendix~\ref{sec:supp-directions}). Mean query tests whether stronger attraction predicts greater benefit from shrinkage. Mean key and the leading key axis test shared key geometry; a fixed template key tests another input-independent reference, and random directions test generic projection.

The sink reference gives the largest gains at the tested strengths ($+4.57$ and $+3.55$). Mean query is more aligned with the queries than the sink key (mean absolute alignment $0.76$ and $0.75$ versus $0.47$ and $0.43$) and attracts more strongly when added to a key at equal edit size (Appendix~\ref{sec:supp-key-boost}), yet gains less when shrunk ($+2.49$ and $+2.33$). Stronger attraction therefore does not imply a larger benefit from shrinkage. Template and random edits change little; the leading key axis reduces accuracy ($-11.08$ and $-5.03$). The mean key helps Qwen-3B ($+2.26$) but not LLaVA-OneVision ($-0.39$), so shared key geometry alone does not account for the gains. In paired comparisons on both models, every alternative direction falls below the sink key, with each 95\% CI below zero.

\ifdefined\designcontroltablesinline\else
\begin{table}[t]
\fi
\centering\small
\setlength{\tabcolsep}{3pt}
\renewcommand{\arraystretch}{1.18}
\caption{\textbf{Reference direction.} Nine-benchmark accuracy change (pp) from a separate run at fixed model-level strengths. $\ast$/$\dagger$: 95\% CIs above/below zero for individual directions.}
\label{tab:direction-controls}
\label{tab:direction-strength}
\begin{tabular*}{\linewidth}{@{\extracolsep{\fill}}@{}lrr@{}}
\toprule
Reference & \shortstack{LLaVA-OV\\7B} & \shortstack{Qwen2.5-VL\\3B} \\
\midrule
Random (range of 3) & $-0.24$ to $+0.40$ & $-0.76$ to $+0.41$ \\
Leading key axis & $-11.08^{\dagger}$ & $-5.03^{\dagger}$ \\
Fixed template key & $+0.14$ & $+0.15$ \\
Mean key & $-0.39$ & $+2.26^{\ast}$ \\
Mean query & $+2.49^{\ast}$ & $+2.33^{\ast}$ \\
\midrule
\textbf{Sink key} & $\mathbf{+4.57}^{\ast}$ & $\mathbf{+3.55}^{\ast}$ \\
\bottomrule
\end{tabular*}
\ifdefined\designcontroltablesinline\else
\end{table}
\fi

\paragraph{One global strength.}
We finally test whether the gains require adapting shrinkage strength to each dataset. Table~\ref{tab:strength-transfer} compares dataset-specific, model-specific, and global validation selection. Model-specific selection stays within $0.33$ points of dataset-specific selection on every model, so per-dataset tuning adds little. One global $\beta=0.3$ retains positive means on all four main models, with pointwise intervals above zero. Model-specific selection can still help: LLaVA-OneVision gains $3.88$ points globally versus $5.14$ with its model-specific choice. For Qwen-3B and Qwen-7B, the global strength coincides with the model-specific one. For Qwen-3B, the global choice is above dataset-specific selection ($3.05$ vs.\ $2.72$); for Qwen-7B and Gemma, it stays within $0.13$ points. A single strength therefore transfers across these models and benchmarks.

\newpage
\ifdefined\designcontroltablesinline\else
\begin{table}[t]
\fi
\centering\small
\setlength{\tabcolsep}{3pt}
\renewcommand{\arraystretch}{1.18}
\caption{\textbf{Strength sharing.} Nine-benchmark mean accuracy change (pp). Bold: row maximum. Global uses $\beta=0.3$; $\ast$: 95\% CI above zero.}
\label{tab:strength-transfer}
\begin{tabular*}{\linewidth}{@{\extracolsep{\fill}}@{}lrrr@{}}
\toprule
Model & \shortstack{Per\\dataset} & \shortstack{Per\\model} & Global \\
\midrule
LLaVA-OV-7B & $+5.10^{\ast}$ & $\mathbf{+5.14}^{\ast}$ & $+3.88^{\ast}$ \\
Qwen2.5-VL-3B & $+2.72^{\ast}$ & $\mathbf{+3.05}^{\ast}$ & $\mathbf{+3.05}^{\ast}$ \\
Qwen2.5-VL-7B & $\mathbf{+0.53}^{\ast}$ & $+0.48^{\ast}$ & $+0.48^{\ast}$ \\
Gemma-3-4B & $\mathbf{+0.84}^{\ast}$ & $+0.72^{\ast}$ & $+0.71^{\ast}$ \\
\bottomrule
\end{tabular*}
\ifdefined\designcontroltablesinline\else
\end{table}
\fi

\subsection{Attention Allocation}
\label{sec:attention-analysis}

The experiments above measure accuracy. We now examine how the edits move attention: group-level redistribution under \methodname (Figure~\ref{fig:attention-drift}) and token-level examples from \methodname, ACG (Key), and GIFT (Key) (Figure~\ref{fig:token-attention}).

Image attention gains exceed question/option gains in all four models. Together, these two groups receive 46\% to 58\% of the attention leaving the sink; template and generated positions receive the rest. Image tokens gain most in LLaVA-OneVision and Gemma. Qwen2.5-VL-7B shows both the smallest sink reduction (5.9 pp) and the smallest image gain (2.7 pp). Redistribution magnitude does not track accuracy: Gemma's image increase exceeds Qwen-3B's despite its smaller gain, and its sink reduction resembles LLaVA-OneVision's. Head eligibility is recomputed in each pass, so the averages cover the heads that each run selects rather than one fixed set.

\par\newpage
\twocolumn[{%
 \def\tokenfigureinline{}%
 \ifdefined\tokenfigureinline
\begin{minipage}{\textwidth}
\captionsetup{type=figure}
\else
\begin{figure*}[t]
\fi
 \centering
 \includegraphics[width=\textwidth]{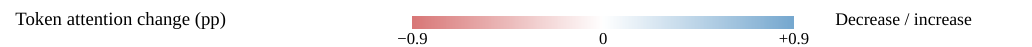}\par
 \includegraphics[width=0.44\textwidth]{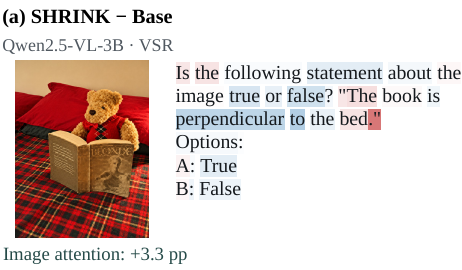}\hfill
 \includegraphics[width=0.44\textwidth]{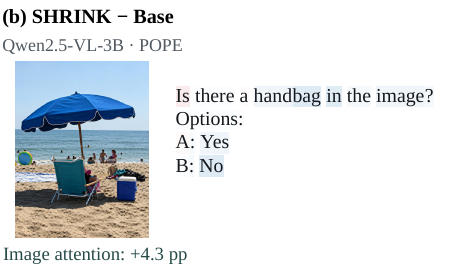}\par\smallskip
 \includegraphics[width=0.44\textwidth]{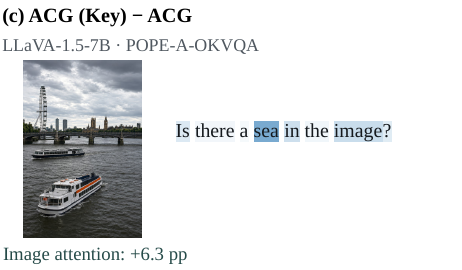}\hfill
 \includegraphics[width=0.44\textwidth]{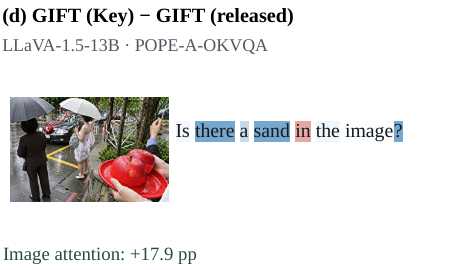}
 \caption{\textbf{Selected token-level attention changes.} Blue/red: increases/decreases, saturated at $\pm0.9$ pp. Numbers give summed image-attention changes. \methodname panels compare runs on the same response continuation; ACG and GIFT are measured before the first answer token. ACG guidance is matched; GIFT uses released settings.}
 \label{fig:token-attention}
\ifdefined\tokenfigureinline
\end{minipage}
\else
\end{figure*}
\fi
 \par\vspace{\textfloatsep}%
}]
\raggedbottom
\paragraph{Changes within an input.}
To examine attention changes within an input, we look beyond the group totals to individual tokens. In Figure~\ref{fig:token-attention}(a), attention rises on ``perpendicular'' and falls on ``bed'' while image attention increases. This example shows how a small net text-attention change can contain opposing token-level shifts. Appendix~\ref{sec:supp-host-attention} specifies the trace protocols.

\section{Discussion}
\label{sec:discussion}

\subsection{Implications}
\label{sec:qualitative-discussion}

The finding extends sink analysis to competition among input tokens. The sink token's key direction contributes to their scores (Eq.~(\ref{eq:main-component-score})), and all-key editing can change their relative attention (Eq.~(\ref{eq:method-attention-ratio})). Attraction strength alone, however, does not determine which direction is useful to shrink. Mean query attracts more strongly when added to a key yet gives smaller shrinkage gains at the tested strengths (Tab.~\ref{tab:direction-controls}). Reference selection must therefore account for the effect of reducing existing components.

\subsection{Limitations and Future Directions}
\label{sec:limitations-future}

Our evaluation covers closed-form tasks, in which the model selects among given options (including yes/no), and models of 3B to 13B parameters. Extending it to open-ended answering, captioning, long-form generation, and larger models remains future work.

\begingroup
\clubpenalty=10000
More broadly, our results suggest that attention sinks matter beyond the attention they absorb: the sink token's key direction also shapes how attention is shared among the remaining input tokens. Sink-aligned components may thus offer an inexpensive indicator of token salience, complementing fixed-direction scoring for cache compression~\cite{godey2025qfilters}, and could be regulated during training rather than edited at inference. Extending this analysis to video, multi-image, and text-only models is a natural next step.
\par\endgroup

\section{Conclusion}
\label{sec:conclusion}

The sink-key direction controls attention beyond the sink position: boosting positive sink-aligned components increases average attention to image and question tokens, while boosting negative components decreases it. \methodname uses this finding to reduce the direction's contribution across visible keys during generation. It improves mean accuracy on the four main VLMs under sampled reasoning. Together, these results show that the key direction of the sink token offers a handle for controlling attention among task-input tokens.

\label{sec:main-end}
\par\pagebreak
\flushbottom
\section*{LLM Usage Statement}
\label{sec:llm-usage}
OpenAI Codex was used during manuscript preparation to assist with drafting and revising text, reviewing related literature, preparing figures and tables, analyzing experimental results, and discussing research design and interpretation. Fable 5.2 was also used to help identify and explore relevant literature, and Claude was used to help write and revise the manuscript text. All AI-assisted content was reviewed by the authors, who remain responsible for the research methodology, scientific claims, experimental results, and final manuscript.

\label{sec:references-start}
{
    \small
    \bibliographystyle{ieeenat_fullname}
    \bibliography{main}
}

\clearpage
\setcounter{page}{1}
\renewcommand{\thepage}{S\arabic{page}}
\maketitlesupplementary
\appendix
\setcounter{topnumber}{4}
\setcounter{dbltopnumber}{4}
\renewcommand{\topfraction}{0.95}
\renewcommand{\dbltopfraction}{0.95}
\renewcommand{\textfraction}{0.03}
\setlength{\floatsep}{10pt plus 2pt minus 2pt}
\setlength{\dblfloatsep}{10pt plus 2pt minus 2pt}
\setlength{\textfloatsep}{12pt plus 2pt minus 2pt}
\setlength{\dbltextfloatsep}{12pt plus 2pt minus 2pt}
\section{Experimental Details}
\label{sec:supp-setup}

\paragraph{Model checkpoints and inputs.}
\Cref{tab:checkpoints} lists the checkpoints. Each uses its own chat template with one user turn: image, then text. Qwen2.5-VL receives the system prompt ``You are a helpful assistant.'' LLaVA-OneVision images are resized to $384\times384$ with bicubic interpolation. LLaVA-1.5 images are padded to a square with the CLIP mean color.

\begin{table*}[t]
  \centering\small
  \caption{Model checkpoints. The first five models form the main evaluation; the remaining four extend the architecture sweep.}
  \label{tab:checkpoints}
  \begin{tabular}{ll}
    \toprule
    Model & Hugging Face checkpoint \\
    \midrule
    LLaVA-OneVision-7B & \texttt{llava-hf/llava-onevision-qwen2-7b-ov-hf} \\
    Qwen2.5-VL-3B & \texttt{Qwen/Qwen2.5-VL-3B-Instruct} \\
    Qwen2.5-VL-7B & \texttt{Qwen/Qwen2.5-VL-7B-Instruct} \\
    Gemma-3-4B & \texttt{google/gemma-3-4b-it} \\
    InternVL3-8B & \texttt{OpenGVLab/InternVL3-8B-hf} \\
    \midrule
    LLaVA-1.5-7B & \texttt{llava-hf/llava-1.5-7b-hf} \\
    Qwen3-VL-4B & \texttt{Qwen/Qwen3-VL-4B-Instruct} \\
    InternVL3.5-8B & \texttt{OpenGVLab/InternVL3\_5-8B-HF} \\
    Gemma-3-12B & \texttt{google/gemma-3-12b-it} \\
    \bottomrule
  \end{tabular}
\end{table*}

\paragraph{Reasoning instruction.}
Every arm receives the following instruction after the question and lettered options:
\begin{quote}\small
Reason step by step before choosing an option. Explain the evidence and check the candidate options. Do not give only an option letter. Write your reasoning inside \texttt{<think>} and \texttt{</think>} tags (ordinary text formatting). Finish with a separate final line exactly like ``\texttt{<answer>A</answer>}'', outside those tags, replacing A with your chosen option letter.
\end{quote}
MMStar retains its original question text, including options. Other benchmarks use a question followed by an explicit options list. Strict scoring accepts an answer tag or an explicit final-answer line; a sentence such as ``Therefore, option D is the correct choice'' alone is rejected. Conflicting explicit decisions are also rejected.

\paragraph{Benchmark splits.}
\Cref{tab:benchmark-splits} lists validation and test sizes. The first four core benchmarks use fixed 20/80 hash splits; POPE is grouped by image. MMStar assigns all 18 calibration questions to validation, then hashes the remaining questions. VSR uses its official zero-shot splits, and WhatsUp, COCO-QA, and VG-QA use AdaptVis's validation/test splits. We retain those splits when converting each subset's open-ended answers into lettered options over fixed option words.

M3CoT uses an image-grouped 20/80 hash split of its released validation set. The nine further benchmarks also use grouped hash splits: questions sharing an image stay together, as do MMVP's question pairs and HallusionBench's figure sets. The validation and test groups are disjoint. All reported scores use only test questions, with $\beta$ selected on validation.

\begin{table}[t]
  \centering\small
  \caption{Validation and test questions. Additional benchmarks use the same strength-selection and strict-scoring protocol as the main table.}
  \label{tab:benchmark-splits}
  \begin{tabular}{lrr}
    \toprule
    Benchmark & Validation & Test \\
    \midrule
    MMBench-EN dev & 866 & 3,463 \\
    POPE & 1,800 & 7,200 \\
    CV-Bench & 528 & 2,110 \\
    EmbSpatial-Bench & 728 & 2,912 \\
    MMStar & 278 & 1,222 \\
    \midrule
    VSR & 340 & 1,222 \\
    WhatsUp & 163 & 657 \\
    COCO-QA & 537 & 2,150 \\
    VG-QA & 290 & 1,161 \\
    \midrule
    M3CoT & 196 & 912 \\
    AI2D & 744 & 2,344 \\
    ScienceQA-IMG & 374 & 1,643 \\
    A-OKVQA & 226 & 919 \\
    MMVP & 66 & 234 \\
    RealWorldQA (MC) & 85 & 343 \\
    MME & 454 & 1,920 \\
    HallusionBench & 189 & 762 \\
    BLINK (single-image) & 153 & 640 \\
    MMMU (single-image MC) & 161 & 640 \\
    \bottomrule
  \end{tabular}
\end{table}

\paragraph{Dataset composition and metric.}
MMBench-EN uses its development questions in a single pass without circular evaluation. POPE contains 3,000 questions in each of its adversarial, popular, and random subsets over 500 COCO val2014 images. CV-Bench contains 788 count, 650 relation, 600 depth, and 600 distance questions. EmbSpatial uses its released test set; MMStar contains 1,500 questions before splitting. WhatsUp includes Controlled Images A+B; COCO-QA and VG-QA include the one- and two-object subsets. POPE, MME, and HallusionBench present Yes/No as lettered choices. We report strict per-question accuracy on all benchmarks, including MMVP, MME, and HallusionBench; their scores here therefore use this common metric rather than benchmark-specific pair or aggregate scores.

\paragraph{Inference and stopping.}
The main sampled evaluation uses vLLM 0.31~\cite{kwon2023vllm}, bf16, and H100 GPUs, with at most 2,048 new tokens. A loop guard stops generation if the last 96 tokens repeat with a period of at most 24 tokens. Sampling uses temperature 0.7 and top-$p=1$, without top-$k$ truncation or repetition penalties. Seeds 1--3 are shared between arms for each question and replicate. Greedy evaluation uses one run per arm and selects strengths independently (Appendix~\ref{sec:supp-greedy}).

\paragraph{Strict scoring.}
A response counts as correct only if it ends naturally and gives an unambiguous explicit decision matching the label. We accept an answer tag or a final line such as ``Final answer: A'' or ``The answer is A.'' Cap or loop-guard stops, conflicting decisions, and prose without an accepted decision count as wrong. The instruction and accepted answer forms above apply to every arm of the main evaluation.

\paragraph{Uncertainty and averaging.}
Accuracy is averaged over the three sampling seeds. Confidence intervals use 10,000 paired bootstrap resamples over test questions, seed 0, applied to each question's seed-mean edited-minus-unedited change. For Core-5, Spatial-4, and All-nine averages, questions are resampled independently within each benchmark and the resulting changes are averaged with equal benchmark weights. Pointwise 95\% intervals condition on the validation-selected strength. A gain ($\ast$) or loss ($\dagger$) is marked only when its interval excludes zero; boldface in the main tables indicates a numerical comparison, not significance. All-19 averages also weight benchmarks equally but are descriptive, without aggregate significance tests. Changes are computed before rounding.

\section{Calibration and Native Key Geometry}
\label{sec:supp-calibration}

Sink-position and emergence profiling uses 16 label-free questions (eight each from MMStar and M3CoT) for LLaVA-OneVision-7B, Qwen2.5-VL-7B, InternVL3-8B, InternVL3.5-8B, and both Gemma models. It uses 20 questions (ten each) for Qwen2.5-VL-3B, Qwen3-VL-4B, and LLaVA-1.5-7B. The extreme-logit diagnostic uses 20 questions per model. All MMStar calibration questions are assigned to its validation split, so none is reported. M3CoT calibration uses test questions, disjoint from the validation-split evaluation used here. Labels are never read. \Cref{tab:native-geometry} gives the native projection and sign statistics.

\begin{table*}[t]
  \centering\footnotesize
  \setlength{\tabcolsep}{4pt}
  \caption{Native geometry on the 20-question diagnostic. Shares are percentages. Negative sink-score shares are over query head-steps selected at $\tau=0.2$; negative projection ratios exclude the sink. Image-key cosines are means within early/middle/late layer thirds.}
  \label{tab:native-geometry}
  \begin{tabular}{lccc}
    \toprule
    Model & $z_s<0$ & $r_j<0$: image / text / generated & Mean image cosine: early / middle / late \\
    \midrule
    LLaVA-OneVision-7B & $1.6$ & $41.3 / 36.1 / 35.9$ & $+0.46 / +0.17 / +0.01$ \\
    Qwen2.5-VL-3B & $2.1$ & $16.3 / 10.0 / 13.3$ & $+0.30 / +0.32 / +0.36$ \\
    Qwen2.5-VL-7B & $1.9$ & $7.4 / 3.9 / 4.9$ & $+0.56 / +0.35 / +0.26$ \\
    Gemma-3-4B & $0.1$ & $12.7 / 4.0 / 2.9$ & $+0.26 / +0.38 / +0.37$ \\
    InternVL3-8B & $1.1$ & $46.5 / 45.1 / 47.1$ & $+0.41 / +0.13 / -0.04$ \\
    LLaVA-1.5-7B & $15.5$ & $96.2 / 95.9 / 95.9$ & $-0.52 / -0.60 / -0.62$ \\
    Qwen3-VL-4B & $0.4$ & $16.1 / 11.2 / 16.4$ & $+0.39 / +0.24 / +0.24$ \\
    InternVL3.5-8B & $0.3$ & $24.9 / 12.8 / 19.5$ & $+0.28 / +0.16 / +0.16$ \\
    Gemma-3-12B & $0.2$ & $20.8 / 8.8 / 6.3$ & $+0.17 / +0.37 / +0.22$ \\
    \bottomrule
  \end{tabular}
\end{table*}

In the 20-question label-free diagnostic, we teacher-force unedited chain-of-thought answers through the fp32 Hugging Face reference. Among decoding head-steps selected by \cref{eq:head-selection}, the sink score is positive in 97.9--99.9\% for eight of the nine VLMs, and in 84.5\% for LLaVA-1.5-7B. Mean image-key cosines with $\usink$ reach 0.6 in magnitude. Negative projection ratios occur in 41.3\% of LLaVA-OneVision image keys, 7.4\% of Qwen2.5-VL-7B image keys, and 96.2\% of LLaVA-1.5 image keys. These decoding diagnostics differ from the all-head prompt-pass measurements in Appendix~\ref{sec:supp-key-boost}.

\section{\texorpdfstring{\methodname}{SHRINK} Formulation and Implementation}
\label{sec:supp-method}

This appendix gives the full key-set formulation, derivations, calibration rules, and implementation details of \cref{sec:method}. The main method applies the projection to all visible keys in selected query heads; the sink-only variant is a comparison.

\subsection{Sink Identification and Native Direction}

Attention notation and the key decomposition are given in Appendices~\ref{sec:supp-additive-transfer} and~\ref{sec:supp-key-boost}. \methodname shrinks the sink-aligned component of every visible key in selected query heads. The sink-only variant serves as a comparison that changes competition with the sink without changing relative non-sink attention.

The sink position $s$ is a template token before the image. We identify it once per model using 16--20 calibration questions from MMStar and M3CoT. Calibration reads only the model's own unedited chain-of-thought answers and attention, never labels. We run an fp32 teacher-forced pass and choose the pre-image template position with maximum attention averaged over decoding rows, heads, questions, and the last third of layers. We assign all MMStar calibration questions to its validation split, so none of them is reported (\cref{sec:exp-setup}). The M3CoT calibration items come from its test split and are disjoint from our validation-split evaluation. Appendix~\ref{sec:supp-calibration} gives the per-model allocation.

Because the prefix through $s$ and its position are fixed, causality makes $\ksink$ identical for every input. It defines a unit vector for each layer and KV head,
\begin{equation}
 \usink^{l,g}=\frac{\ksink^{l,g}}{\lVert\ksink^{l,g}\rVert_2}.
 \label{eq:sink-direction}
\end{equation}
All query heads in a GQA group share this direction. To measure another key's component along it, we define the projection ratio
\begin{equation}
 r_j=\frac{\kk_j^\top\usink}{\lVert\ksink\rVert_2}.
 \label{eq:projection-ratio}
\end{equation}
This is the coefficient in \cref{eq:dose-decomposition}: $r_s=1$, whereas $r_j=0$ for a key orthogonal to $\usink$. Content keys need not be orthogonal to $\usink$. Their components along this direction make directional shrinkage distinct from changing only the sink logit.

\subsection{Sink-Direction Key Projection}

For strength $\beta\in(0,1]$ and key set $\mathcal{S}_t\subseteq\mathcal{V}_t$, \methodname applies
\begin{equation}
 \kk'_j=
 \begin{cases}
 \kk_j-\beta(\kk_j^\top\usink)\usink,&j\in\mathcal{S}_t,\\
 \kk_j,&j\notin\mathcal{S}_t.
 \end{cases}
 \label{eq:key-projection-general}
\end{equation}
The corresponding logit change is
\begin{equation}
 \Delta z_{t,j}=-\beta c_l(\q_t^\top\usink)(\kk_j^\top\usink),
 \qquad j\in\mathcal{S}_t.
 \label{eq:logit-change}
\end{equation}
Equivalently, \methodname replaces $r_j$ by $(1-\beta)r_j$ in \cref{eq:native-direction-dose}, leaving $\kk_j^{\perp}$ unchanged. At $\beta=1$ the component is removed completely; smaller strengths retain part of it.

\paragraph{\sinkvar: project the sink key.}
For the sink-only comparison, setting $\mathcal{S}_t=\{s\}$ gives $\kk'_s=(1-\beta)\ksink$ and $z'_{t,s}=(1-\beta)z_{t,s}$; every other logit is unchanged. Hence every non-sink key's margin over the sink changes by the same $\beta z_{t,s}$. The conditional attention distribution among non-sink keys remains unchanged; when $z_{t,s}>0$, their total mass increases by a shared renormalization.

\paragraph{\allvar: project every visible key.}
Our main method sets $\mathcal{S}_t=\mathcal{V}_t$, covering the sink, input, and generated keys visible to the current query. This yields an exact query-side interpretation. Since $\usink\usink^\top$ is symmetric,
\begin{equation}
 \q_t^\top(\mathbf{I}-\beta\usink\usink^\top)\kk_j
 =\big[(\mathbf{I}-\beta\usink\usink^\top)\q_t\big]^\top\kk_j.
 \label{eq:query-equivalence}
 \csname ltx@label\endcsname{{eq:main-query-equivalence}}
\end{equation}
Thus projecting all keys is equivalent to using $\q'_t=(\mathbf{I}-\beta\usink\usink^\top)\q_t$. Combining \cref{eq:logit-change,eq:projection-ratio}, the change in key $j$'s margin over the sink is
\begin{equation}
 \Delta(z_{t,j}-z_{t,s})
 =\beta c_l(\q_t^\top\usink)\lVert\ksink\rVert_2(1-r_j).
 \label{eq:margin-change}
\end{equation}
When $\q_t^\top\usink>0$, keys with $r_j<1$ gain on the sink, and keys with smaller projection ratios gain more. More generally, the change in the log attention ratio of any two visible positions is
\begin{equation}
 \Delta\log\frac{a_{t,j}}{a_{t,i}}
 =-\beta z_{t,s}(r_j-r_i).
 \label{eq:shrink-pairwise}
\end{equation}
Thus the all-key operation changes competition among content tokens whenever their projection ratios differ. This is not equivalent to subtracting a common bias from every logit, which would cancel under softmax. It does not guarantee that every content token gains attention, and no positive-sink-score gate is used in the \methodname implementation.

\subsection{Selecting Heads at Each Token}

Let $\mathcal{I}$ be the image positions and $\mathcal{T}$ the question and answer-option positions. Selection uses the native attention from \cref{eq:attention}, before any edit. At every generated token, we select
\begin{equation}
 \mathcal{H}^{l}_t=\left\{h:\ \sum_{j\in\mathcal{I}}a^{\mathrm{nat},l,h}_{t,j}\geq\tau
 \ \lor\
 \sum_{j\in\mathcal{T}}a^{\mathrm{nat},l,h}_{t,j}\geq\tau\right\}.
 \label{eq:head-selection}
 \csname ltx@label\endcsname{{eq:main-head-selection}}
\end{equation}
We use $\tau=0.2$. The image condition follows VAR's initial image-attention filter~\cite{kang2025var}; VAR additionally selects by the fraction of visual attention assigned to non-sink tokens. SPIN likewise uses native image attention to select heads~\cite{sarkar2025spin}. Our text condition extends selection to heads reading the question or choices. Because we implement \cref{eq:logit-change} per query head, an edited query head does not alter unselected siblings that share its KV head under GQA. The head-selection rule is recomputed from the native attention of the current row, before that row's edit.

On unedited calibration decoding, this rule selects an average of 8--19\% of query heads per step in the edited layers.

\subsection{Scheduling Layers}

Let $\bar a_s^l$ denote mean calibration-set sink attention in layer $l$. Within the first half of an $L$-layer decoder, we define the sink emergence layer as the largest one-layer rise,
\begin{equation}
 \ell_{\mathrm{emg}}
 =\underset{1\leq l<L/2}{\arg\max}\;\big(\bar a_s^l-\bar a_s^{l-1}\big).
 \label{eq:emergence}
\end{equation}
We also exclude any layer in which more than 1\% of selected calibration head-steps have an extreme native sink logit,
\begin{equation}
 \mathcal{L}_{\mathrm{excl}}
 =\left\{l:\Pr\big(|z_{t,s}^{l,h}|>50\mid h\in\mathcal{H}_t^l\big)>0.01\right\}.
 \label{eq:excluded-layers}
\end{equation}
Let $\ell_{\mathrm{start}}=\ell_{\mathrm{emg}}$ by default. LLaVA-OneVision, the first model we configured, keeps its initial schedule: edits start at layer 0, and layers 0 and 27 are excluded. Layer 0 meets the extreme-logit rule, and layer 27 has 0.8\% extreme head-steps; its measured emergence layer is 4. The edited layer set is
\begin{equation}
 \mathcal{L}_{\mathrm{edit}}
 =\{l:l\geq\ell_{\mathrm{start}}\}\setminus\mathcal{L}_{\mathrm{excl}}.
 \label{eq:edited-layers}
\end{equation}
The sink and emergence profiles use 16--20 questions; the extreme-logit diagnostic uses 20 questions. No labels tune these choices.

\begin{table}[t]
  \centering
  \small
  \setlength{\tabcolsep}{3.2pt}
  \caption{Frozen sink and layer configuration. A dash denotes no excluded layer.}
  \label{tab:schedule}
  \begin{tabular}{lccc}
    \toprule
    Model & $s$ & $\ell_{\mathrm{start}}$ & $\mathcal{L}_{\mathrm{excl}}$ \\
    \midrule
    LLaVA-OneVision-7B & 0 & 0 & 0, 27 \\
    LLaVA-1.5-7B & 0 & 2 & -- \\
    Qwen2.5-VL-3B  & 0 & 3 & 0 \\
    Qwen2.5-VL-7B  & 2 & 4 & 0, 1 \\
    Qwen3-VL-4B    & 1 & 7 & -- \\
    InternVL3-8B   & 2 & 4 & 0, 1 \\
    InternVL3.5-8B & 0 & 7 & -- \\
    Gemma-3-4B     & 0 & 5 & -- \\
    Gemma-3-12B    & 0 & 1 & 4 \\
    \bottomrule
  \end{tabular}
\end{table}

The sink token is \texttt{<bos>} for both Gemma~3 models, \texttt{<s>} for LLaVA-1.5, \texttt{<|im\_start|>} for LLaVA-OneVision, Qwen2.5-VL-3B, and InternVL3.5-8B, \texttt{user} for Qwen3-VL-4B, and the newline after the role tag for Qwen2.5-VL-7B and InternVL3-8B.

\subsection{Implementation Details}

We edit the query that predicts the first answer token and every generated-token query thereafter; earlier prompt rows remain native. The implementation reads the native post-RoPE sink key for each layer and KV head and computes the projection in vLLM~\cite{kwon2023vllm}. The stored keys and values are never modified. An edit is skipped whenever the sink is outside the query's attention window. All reported results use the direct path: transient projected keys rounded to the key dtype. An optimized query-side path also exists, based on \cref{eq:query-equivalence}, but produces none of the reported results. The equivalence is exact in real arithmetic.

Against an independent fp32 Hugging Face implementation, next-token log-probabilities differ by at most $2.5\times10^{-4}$ over 18 key-projection test cases per model, and by at most $4\times10^{-5}$ for six of the nine VLMs. Every job must pass this check before running. The intervention requires no training or additional model forward pass. We report shrinkage strengths and evaluation choices in \cref{sec:experiments}.

Head selection and attention recomputation add inference work, although no additional full model forward pass is required.

\section{Controlled Key and Value Transfer}
\label{sec:supp-transfer}

\begin{table*}[t]
  \centering\small
  \setlength{\tabcolsep}{5pt}
  \caption{Recipient attention (\%) under all six key/value-transfer conditions, with 240 questions per model and dataset. The last column gives the paired key-copy increase over the matched no-copy control, in percentage points, with its 95\% interval. Figure~\ref{fig:key-control}a shows the first three models.}
  \label{tab:key-transfer-models}
  \label{tab:key-transfer}
  \begin{tabular}{llrrrrrrr}
    \toprule
    Model & Dataset & STD & BLOCK & RR & SR & RS & SS & SR$-$RR [95\% CI] \\
    \midrule
    LLaVA-OneVision-7B & MMStar & 0.006 & 49.858 & 0.062 & 42.335 & 0.008 & 49.754 & +42.27 [+41.19, +43.36] \\
     & CV-Bench 2D & 0.005 & 51.556 & 0.058 & 46.602 & 0.007 & 51.430 & +46.54 [+46.02, +47.04] \\
    \midrule
    Qwen2.5-VL-7B & MMStar & 0.072 & 42.114 & 0.099 & 76.433 & 0.089 & 43.546 & +76.33 [+76.18, +76.49] \\
     & CV-Bench 2D & 0.061 & 41.975 & 0.087 & 77.290 & 0.069 & 43.863 & +77.20 [+77.11, +77.29] \\
    \midrule
    Gemma-3-4B & MMStar & 0.279 & 0.003 & 0.511 & 15.043 & 0.627 & 46.365 & +14.53 [+14.36, +14.71] \\
     & CV-Bench 2D & 0.297 & 0.002 & 0.504 & 16.106 & 0.632 & 45.025 & +15.60 [+15.45, +15.76] \\
    \midrule
    Qwen2.5-VL-3B & MMStar & 0.024 & 26.093 & 0.028 & 2.997 & 0.074 & 26.289 & +2.97 [+2.81, +3.14] \\
     & CV-Bench 2D & 0.019 & 26.025 & 0.026 & 5.055 & 0.064 & 26.207 & +5.03 [+4.86, +5.19] \\
    \bottomrule
  \end{tabular}
\end{table*}

\Cref{fig:key-control}a shows LLaVA-OneVision-7B, Qwen2.5-VL-7B, and Gemma-3-4B, one model per family. \Cref{tab:key-transfer-models} retains Qwen2.5-VL-3B and all six conditions. Each model uses the same 240 MMStar and 240 CV-Bench 2D questions, with 24 disjoint calibration questions per dataset to identify the sink and recipient before evaluation. Recipient attention is measured from the last prompt query, averaged over all heads and the final third of decoder layers, then over questions. The 95\% intervals use 10,000 question-bootstrap resamples within category strata, seed 20261009, pairing conditions within each resample. Each question has a distinct image.

Qwen2.5-VL-7B and Gemma use the original reference runs; LLaVA-OneVision and Qwen2.5-VL-3B use the same protocol in the fp32 reference. A 24-question check per dataset reproduces Qwen2.5-VL-7B's sink, recipient, and prompt lengths, with mean attention differences at most 0.0151 percentage points and maximum individual difference 0.250 points. Gemma's backend supports bidirectional image attention. The probe questions overlap the main evaluation splits (107 validation and 373 test questions), but this probe does not select \methodname strengths. The final-third means include propagation through earlier edited layers.

The recipient is the first natively eligible pre-content template position after the sink. All transplanted vectors come from the same prompt's unedited source pass. The original sink is blocked as a key for other queries, with its native self-row retained; all other native permissions remain. RR fixes the ordinary recipient key and value, SR transfers the sink key only, RS the sink value only, and SS both. BLOCK allows natural relocation without clamping; the matched key-copy effect is SR minus RR. Joint key/value transfer can differ from key-only transfer after propagation, particularly on Gemma, so key-mediated attraction does not imply that values are functionally interchangeable. The VLM CV-Bench sample contains count and relation questions. This fp32 direct-answer probe is separate from the bf16 reasoning-generation evaluation.

For Qwen2.5-VL-7B, accuracy is measured with direct-answer option probabilities on the same 240 questions per dataset. Key-only transfer reduces accuracy by 35.0 points on MMStar (95\% CI $[-43.3,-26.7]$) and 40.8 on CV-Bench (2D) ($[-47.9,-33.3]$). Joint transfer changes accuracy by $+1.2$ and $+0.8$ points from STD accuracies of 64.6\% and 78.3\%. Its recipient attention is also different (43.5--43.9\%), so this is not a fixed-attention comparison of values. Intervals use 10,000 paired category-stratified question-bootstrap resamples, seed 20261009; the original planned contrast family uses Holm correction. These propagated outcomes show why transferable attraction need not imply functional equivalence.

\section{Attention Response to Additive Key Transfer}
\label{sec:supp-additive-transfer}

The transplant in \cref{sec:sink-dose} tests full key replacement. Here we derive the response to a fixed additive transfer. For a query $\q_t$, post-positional-encoding keys $\kk_i$, and visible positions $\mathcal V_t$, scaled dot-product attention~\cite{vaswani2017attention} gives
\begin{equation}
 z_{t,i}=c_l\q_t^\top\kk_i,\qquad
 a_{t,i}=\frac{\exp z_{t,i}}{\sum_{m\in\mathcal V_t}\exp z_{t,m}},
 \label{eq:attention}
\end{equation}
where $c_l>0$ is the layer's attention scale. Hold the query and all other keys fixed, omit the query index, and write $z_s=c_l\q_t^\top\ksink$. The analytical additive response below is distinct from the empirical native-component sweep in Appendix~\ref{sec:supp-key-boost}.

\paragraph{Key addition transfers part of the sink's score.}
For a recipient $j\ne s$, add a fraction $\delta\geq 0$ of the sink key while retaining the recipient's original value:
\begin{equation}
 \begin{aligned}
 \kk_j(\delta)&=\kk_j+\delta\ksink,\\
 z_j(\delta)&=c_l\q_t^\top(\kk_j+\delta\ksink)
             =z_j+\delta z_s.
 \end{aligned}
 \label{eq:additive-dose}
\end{equation}
The recipient therefore gains $\delta$ times the sink's native score. Its value remains unchanged: the intervention changes how strongly the token is retrieved, while preserving the content it supplies.

\paragraph{The score change produces a graded attention response.}
Only the recipient's score changes. Substituting \cref{eq:additive-dose} into the softmax gives
\begin{equation}
 \begin{aligned}
 a_j(\delta)
 &=\frac{\exp(z_j+\delta z_s)}
 {\exp(z_j+\delta z_s)+\sum_{i\ne j}\exp z_i}\\
 &=\frac{a_j\exp(\delta z_s)}
 {1+a_j[\exp(\delta z_s)-1]},
 \end{aligned}
 \label{eq:dose-probability}
\end{equation}
where the sum is over the other visible positions and $a_j=a_j(0)$ is the native attention. Differentiating yields
\begin{equation}
 \frac{\partial a_j(\delta)}{\partial\delta}
 =z_s\,a_j(\delta)[1-a_j(\delta)].
 \label{eq:dose-derivative}
\end{equation}
When $z_s>0$, the derivative is positive: increasing the transfer strength continuously increases recipient attention, approaching one as $\delta$ grows. The recipient's attention ratio against any unedited token is multiplied by $\exp(\delta z_s)$. The sink key therefore serves as an \emph{attention control signal}: $\delta$ sets the transfer strength, and the query's alignment with the sink key determines the signal's effect. If $z_s=0$ the edit has no effect; if $z_s<0$ its effect reverses. High sink attention alone does not establish this sign condition.

This is an exact fixed-query response, not an additional empirical sweep. Appendix~\ref{sec:supp-key-boost} instead tests scaling an existing native component through the full network.

\section{Native Sink-Component Probe}
\label{sec:supp-key-boost}

This probe tests whether changing a token's existing sink-direction component moves its attention as predicted by \cref{eq:native-scaling-prob}. It is distinct from the full key/value replacement in Appendix~\ref{sec:supp-transfer} and the fixed additive edit in \cref{eq:additive-dose}. Only prompt attention is measured; the probe does not evaluate accuracy or attention during generation.

\paragraph{Exact local response.}
For a nonzero sink key, decompose an ordinary key into an orthogonal component $\kk_j^\perp$ and a parallel component:
\begin{equation}
 \kk_j=\kk_j^\perp+r_j\ksink,\qquad
 r_j=\frac{\kk_j^\top\ksink}{\lVert\ksink\rVert_2^2},
 \label{eq:dose-decomposition}
\end{equation}
Its native score separates as
\begin{equation}
 z_j=c_l\q_t^\top\kk_j^\perp+r_jz_s.
 \label{eq:native-direction-dose}
\end{equation}
The probe scales the parallel component while retaining the orthogonal component and value locally. With $\usink=\ksink/\lVert\ksink\rVert_2$,
\begin{equation}
 \begin{aligned}
 \kk_j(\eta)
 &=\kk_j+\eta(\kk_j^\top\usink)\usink\\
 &=\kk_j^\perp+(1+\eta)r_j\ksink.
 \end{aligned}
 \label{eq:native-scaling}
\end{equation}
With the query and all other keys fixed, \cref{eq:native-direction-dose,eq:native-scaling} give
\begin{equation}
 \Delta z_j=\eta r_jz_s,
 \label{eq:native-scaling-score}
\end{equation}
Substituting this score change into the softmax,
\begin{equation}
 \operatorname{logit}a_j(\eta)
 =\operatorname{logit}a_j+\eta r_jz_s,
 \label{eq:native-scaling-prob}
\end{equation}
where $\operatorname{logit}p=\log[p/(1-p)]$. Equivalently, the probability follows \cref{eq:dose-probability} with score increment $\eta r_jz_s$. This increment can have either sign even for $\eta>0$, whereas fixed additive transfer increases the coefficient by $\alpha$ irrespective of its original sign.

\subsection{Protocol and Controls}

\paragraph{Models and questions.}
We probe the four main models: LLaVA-OneVision-7B, Qwen2.5-VL-3B/7B, and Gemma-3-4B. Each receives the same 100 validation questions: 50 MMStar and 50 CV-Bench (2D), sampled separately with seed 0. No test question or answer label is used. Prompts contain the image, the model's chat template, and the same reasoning instruction as the main evaluation. Each condition is a single full prompt forward pass through the Hugging Face reference, with fp32 computation on bf16-rounded weights. This differs from the bf16 vLLM decoding evaluation.

\begin{table*}[t]
 \centering\small
 \caption{Native-component probe configuration. Layers and positions are zero-indexed. Every KV head is edited from sink emergence through the final layer; no head selection or layer exclusion is applied. The last column counts question-layer-query-head units per recipient kind, over 100 questions.}
 \label{tab:native-probe-setup}
 \begin{tabular}{lrrrrr}
 \toprule
 Model & Edited layers & Query / KV heads & Sink position & Mean prompt length & Units \\
 \midrule
 LLaVA-OneVision-7B & 4--27 & 28 / 4 & 0 & 1,612 & 67,200 \\
 Qwen2.5-VL-3B & 3--35 & 16 / 2 & 0 & 400 & 52,800 \\
 Qwen2.5-VL-7B & 4--27 & 28 / 4 & 2 & 400 & 67,200 \\
 Gemma-3-4B & 5--33 & 8 / 4 & 0 & 390 & 23,200 \\
 \bottomrule
 \end{tabular}
\end{table*}

\paragraph{Recipients and edit schedule.}
The primary run draws one image token uniformly per question and, separately, one question-text token, seeded by question identifier. Question text excludes answer options and instructions. Each recipient is edited alone, after positional encoding, in every KV head and every layer from sink emergence onward (\cref{tab:native-probe-setup}). This schedule deliberately differs from \methodname's selected heads and excluded layers. At each layer the operation leaves the orthogonal key component and value tensor unchanged locally. Edits to earlier layers can nevertheless change the hidden states, queries, other keys, and values arriving at later layers. Thus this probe does not hold the entire forward pass fixed or test the simultaneous editing of many tokens.

\paragraph{Conditions.}
The unedited pass provides native attention and projection coefficients. We apply \cref{eq:native-scaling} at $\eta\in\{0.5,1,2,4\}$, and remove the component at $\eta=-1$. For each positive strength, additional conditions use perturbation magnitude
\begin{equation}
 d=|\eta|\,|\kk_j^\top\usink|.
 \label{eq:probe-perturbation-size}
\end{equation}
RAND adds $d\mathbf{w}$, where $\mathbf{w}$ is a random unit vector orthogonal to $\usink$, fixed by question, recipient kind, layer, and KV head across strengths. NORM adds $d\kk_j/\lVert\kk_j\rVert_2$, lengthening the key in its own direction. Because this also scales its sink coefficient, NORM is not a clean null for the component hypothesis. MQ adds $d\bar{\q}/\lVert\bar{\q}\rVert_2$, where $\bar{\q}$ averages post-positional-encoding queries over later prompt rows and query heads in the recipient's GQA group. This uses prompt queries as a reference direction, rather than as a proposed decoding intervention. Each arm computes its perturbation from the tensors arriving at that layer; those tensors can differ between arms after earlier edits propagate.

The sweep parameter satisfies $\alpha=1+\eta$ in the main text's notation. \methodname corresponds to $\eta=-\beta$ and edits all visible keys simultaneously; this single-recipient sweep tests controllability, not its chosen decoding strengths. Post-positional components can reflect position as well as content; the probe does not separate those sources.

\paragraph{Measurement and intervals.}
For each question, edited layer, and query head, we first average attention to the recipient over all prompt rows after it. We then form edited-to-unedited ratios and report their geometric mean over these units. Sign strata use $r_j$ from the unedited pass at each layer and KV head, repeated across its query heads. A token can belong to opposite strata in different layers or heads. The final prompt row, which predicts the first answer token, is also measured separately. To handle zeros in the saved fp16 attention summaries, ratios use $(a^{\mathrm{edit}}+\epsilon)/(a^{\mathrm{native}}+\epsilon)$ with $\epsilon=10^{-12}$. The 95\% intervals use 2,000 bootstrap resamples of questions, seed 0, moving all units of a question together because layer-head units within a question are correlated. These original tables retain $\epsilon=10^{-12}$. Figure~\ref{fig:key-control}(c) is recomputed with $\epsilon=6\times10^{-8}$, near the smallest positive fp16 value, using the same question bootstrap. The signed-stratum responses persist, but the magnitudes of extreme decreases depend on storage precision. For example, the LLaVA-OneVision negative image stratum at $\eta=4$ changes from $1.8\times10^{-6}$ to $1.7\times10^{-3}$; restricting to nonzero summaries gives $1.3\times10^{-2}$ and excludes a substantial fraction of units. None resolves probabilities rounded to zero. These large amplification settings lie outside \methodname's partial-removal range.

\subsection{Signed Response and Alternative Directions}

\begin{figure*}[t]
 \centering
 \includegraphics[width=\textwidth]{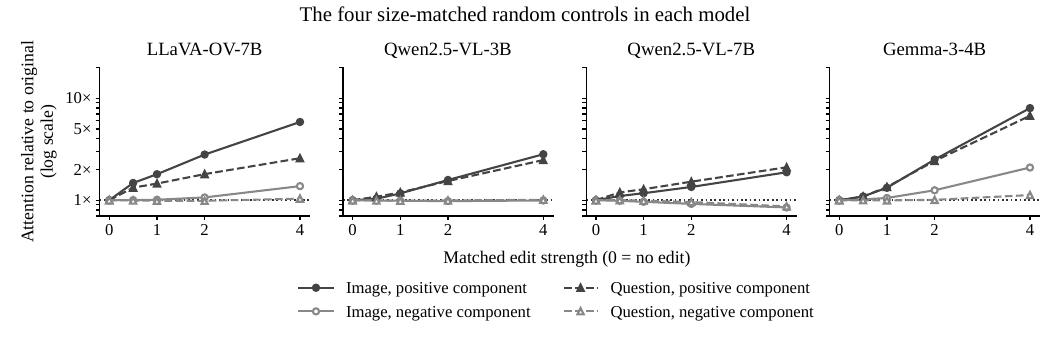}
 \caption{The four random-control means summarized by the grey range in Figure~\ref{fig:key-control}c. Filled and open markers distinguish positive and negative native components; circles and triangles distinguish image and question recipients. Perturbations match the local key-change size of the sink-direction edit. A ratio of $1$ means unchanged attention.}
 \label{fig:random-control-curves}
\end{figure*}

The grey range in \cref{fig:key-control}c spans four RAND means at each matched edit size: image and question recipients, each split by the native coefficient's sign. It is a descriptive minimum-to-maximum range, not a confidence interval. \Cref{fig:random-control-curves} shows the four curves separately. A random direction orthogonal to the sink direction need not be orthogonal to a query, so its score change need not vanish.

\begin{table*}[t]
 \centering\small
 \setlength{\tabcolsep}{4pt}
 \caption{Random recipients, units with $r_j>0$: geometric-mean recipient-attention ratios and 95\% question-bootstrap intervals. Boost doubles the native coefficient ($\eta=1$); removal sets it to zero ($\eta=-1$). RAND uses a random orthogonal direction at $\eta=1$. $N$ counts question-layer-query-head units from 100 questions per model. Share is the percentage of all units in the stratum.}
 \label{tab:native-positive}
 \begin{tabular}{llrrccc}
 \toprule
 Model & Recipient & $N$ & Share (\%) & Boost & Removal & RAND \\
 \midrule
 LLaVA-OneVision-7B & image & 34,797 & 51.8 & $5.18\;[4.82,5.58]$ & $0.0135\;[0.0128,0.0142]$ & $1.8\;[1.74,1.86]$ \\
  & question & 39,522 & 58.8 & $5.59\;[5.31,5.89]$ & $0.0112\;[0.0106,0.0119]$ & $1.46\;[1.42,1.49]$ \\
 \midrule
 Qwen2.5-VL-3B & image & 42,560 & 80.6 & $8.6\;[7.72,9.62]$ & $0.00705\;[0.0067,0.00742]$ & $1.17\;[1.12,1.21]$ \\
  & question & 49,360 & 93.5 & $8.37\;[7.97,8.81]$ & $0.00671\;[0.00634,0.00712]$ & $1.19\;[1.16,1.22]$ \\
 \midrule
 Qwen2.5-VL-7B & image & 60,599 & 90.2 & $3.14\;[2.78,3.57]$ & $0.0333\;[0.0316,0.0352]$ & $1.17\;[1.14,1.2]$ \\
  & question & 64,960 & 96.7 & $5.57\;[5.2,5.98]$ & $0.022\;[0.0208,0.0231]$ & $1.28\;[1.26,1.3]$ \\
 \midrule
 Gemma-3-4B & image & 21,360 & 92.1 & $9.63\;[8.5,11]$ & $0.0111\;[0.00981,0.0125]$ & $1.32\;[1.27,1.38]$ \\
  & question & 22,814 & 98.3 & $15.4\;[14.3,16.5]$ & $0.00216\;[0.00195,0.00241]$ & $1.35\;[1.32,1.37]$ \\
 \bottomrule
 \end{tabular}
\end{table*}

\begin{table*}[t]
 \centering\small
 \setlength{\tabcolsep}{4pt}
 \caption{Random recipients, units with $r_j<0$: geometric-mean recipient-attention ratios and 95\% question-bootstrap intervals. Boost doubles the native coefficient ($\eta=1$); removal sets it to zero ($\eta=-1$). RAND uses a random orthogonal direction at $\eta=1$. $N$ counts question-layer-query-head units from 100 questions per model. Share is the percentage of all units in the stratum.}
 \label{tab:native-negative}
 \begin{tabular}{llrrccc}
 \toprule
 Model & Recipient & $N$ & Share (\%) & Boost & Removal & RAND \\
 \midrule
 LLaVA-OneVision-7B & image & 32,403 & 48.2 & $0.0245\;[0.0204,0.0291]$ & $9.92\;[9.23,10.7]$ & $1.01\;[0.994,1.03]$ \\
  & question & 27,678 & 41.2 & $0.105\;[0.0981,0.112]$ & $4.67\;[4.45,4.91]$ & $0.988\;[0.977,1]$ \\
 \midrule
 Qwen2.5-VL-3B & image & 10,240 & 19.4 & $0.129\;[0.118,0.142]$ & $3.37\;[3.18,3.58]$ & $0.988\;[0.973,1]$ \\
  & question & 3,440 & 6.5 & $0.146\;[0.126,0.166]$ & $3.03\;[2.77,3.33]$ & $0.989\;[0.962,1.01]$ \\
 \midrule
 Qwen2.5-VL-7B & image & 6,601 & 9.8 & $0.127\;[0.101,0.162]$ & $2.71\;[2.5,2.93]$ & $0.964\;[0.94,0.987]$ \\
  & question & 2,240 & 3.3 & $0.326\;[0.285,0.372]$ & $1.89\;[1.78,2]$ & $0.989\;[0.959,1.02]$ \\
 \midrule
 Gemma-3-4B & image & 1,840 & 7.9 & $0.128\;[0.102,0.16]$ & $4.6\;[4.31,4.93]$ & $1.05\;[1.01,1.1]$ \\
  & question & 386 & 1.7 & $0.268\;[0.23,0.317]$ & $2.26\;[2.08,2.45]$ & $0.996\;[0.969,1.02]$ \\
 \bottomrule
 \end{tabular}
\end{table*}

Both signs occur in the unedited keys. Pooling layers and heads, negative components account for $48.2\%$, $19.4\%$, $9.8\%$, and $7.9\%$ of image-key measurements on LLaVA-OneVision, Qwen2.5-VL-3B, Qwen2.5-VL-7B, and Gemma-3-4B, respectively. We abbreviate the last three as Qwen-3B, Qwen-7B, and Gemma.

\Cref{tab:native-positive,tab:native-negative} give the primary results with stratum sizes and intervals. At $\eta=1$, attention moves in the coefficient-predicted direction in 67--92\% of positive-coefficient units and 85--99\% of negative-coefficient units. Removal reverses these directions in 86--100\% and 89--97\%, respectively. At the same strength, RAND agrees with the coefficient-based direction in 46--62\% of units, with a modest increase for positive coefficients.

\begin{figure*}[t]
 \centering
 \includegraphics[width=\textwidth]{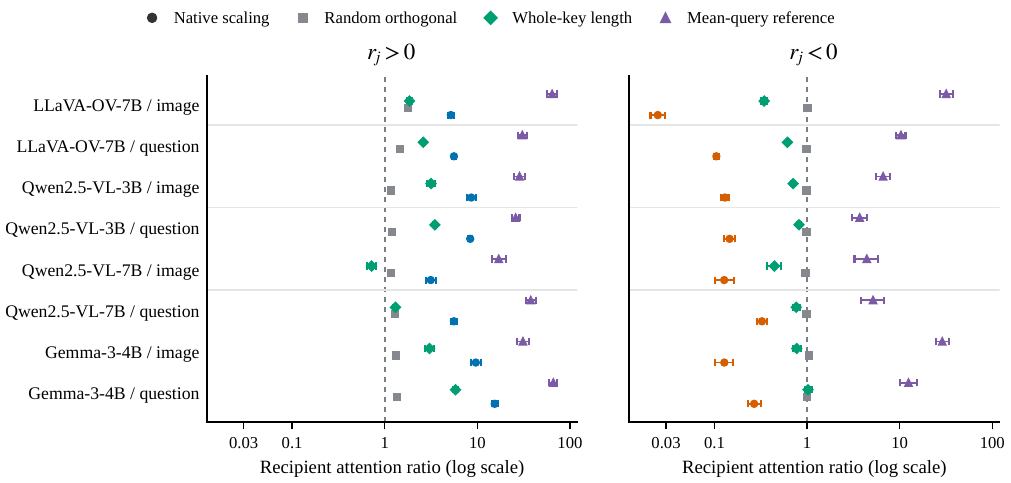}
 \caption{\textbf{The sink direction is a signed control direction, not a maximum-attraction direction.} Random-recipient results at $\eta=1$, split by the native coefficient. Points are geometric-mean attention ratios over later prompt rows; bars are 95\% question-bootstrap intervals. All conditions use the local perturbation norm in \cref{eq:probe-perturbation-size}. NORM also scales the sink component. The mean-query reference (MQ) attracts more strongly and raises attention even in the negative-coefficient stratum.}
 \label{fig:direction-controls}
\end{figure*}

\Cref{fig:direction-controls} compares the native-component edit with RAND, NORM, and MQ. MQ raises attention by $14.9$--$64.3$ across all units at $\eta=1$, and by $3.7$--$31.8$ in negative-coefficient units where native-component scaling lowers it. Across strengths up to four, the all-unit ratio reaches 1,288. Query alignment can therefore attract more attention than the sink direction at the same local perturbation norm. The property studied here is that ordinary keys already carry a signed component along the native sink direction, which can be isolated and scaled.

\begin{table*}[t]
 \centering\small
 \setlength{\tabcolsep}{4pt}
 \caption{Random recipients: whole-population and first-answer-query results. The first two numeric columns aggregate all units using attention averaged over later prompt rows (95\% intervals). MQ is the matched-norm mean-query reference. The last two columns use only the final prompt row, split by native coefficient sign, and give boost / removal ratios; these last-row ratios are shown without intervals. Boost and MQ use $\eta=1$; removal uses $\eta=-1$.}
 \label{tab:native-scope}
 \begin{tabular}{llcccc}
 \toprule
  & & \multicolumn{2}{c}{All units, mean over later rows} & \multicolumn{2}{c}{First-answer query: boost / removal} \\
 \cmidrule(lr){3-4}\cmidrule(lr){5-6}
 Model & Recipient & Boost & MQ & $r_j>0$ & $r_j<0$ \\
 \midrule
 LLaVA-OneVision-7B & image & $0.392\;[0.346,0.442]$ & $45.7\;[40.5,52.1]$ & $6.1 / 0.0109$ & $0.002 / 12.8$ \\
  & question & $1.09\;[1.02,1.15]$ & $19.6\;[17.7,21.5]$ & $4.95 / 0.0054$ & $0.0588 / 5.39$ \\
 \midrule
 Qwen2.5-VL-3B & image & $3.81\;[3.49,4.19]$ & $21.5\;[18.9,24.4]$ & $22.2 / 0.00272$ & $0.0553 / 3.8$ \\
  & question & $6.43\;[6.11,6.75]$ & $22.8\;[21,24.9]$ & $9.44 / 0.0037$ & $0.115 / 3.33$ \\
 \midrule
 Qwen2.5-VL-7B & image & $2.29\;[2.05,2.58]$ & $14.9\;[12.8,17.6]$ & $4.54 / 0.0146$ & $0.0262 / 3.72$ \\
  & question & $5.07\;[4.7,5.46]$ & $35.3\;[31.1,40.2]$ & $7.68 / 0.0109$ & $0.203 / 1.98$ \\
 \midrule
 Gemma-3-4B & image & $6.83\;[6.04,7.81]$ & $30.9\;[27.1,35.2]$ & $18.8 / 4.66\!\times\!10^{-4}$ & $0.0539 / 8.05$ \\
  & question & $14.4\;[13.4,15.4]$ & $64.3\;[58.5,70.2]$ & $24.4 / 2.41\!\times\!10^{-4}$ & $0.141 / 2.23$ \\
 \bottomrule
 \end{tabular}
\end{table*}

The all-unit averages in \cref{tab:native-scope} show how the sign mix shapes whole-token responses: LLaVA-OneVision image-recipient attention falls to $0.39$ at $\eta=1$, because 48.2\% of its units have negative coefficients and their attention falls sharply. Its question-recipient ratio is $1.09$. In Qwen and Gemma, all-unit ratios rise by $2.3$--$14.4$, while negative-coefficient units account for only 1.7--19.4\% of the primary-run strata. The final prompt row shows the same positive/negative response pattern. No confidence intervals were computed for those last-row ratios.

\subsection{Selecting Recipients with More Negative Coefficients}

An additional unedited pass measures $r_j$ for every image or question-text candidate across edited layers and KV heads. We choose the candidate with the largest fraction of negative coefficients, breaking ties by the lowest mean coefficient, and repeat the same conditions except MQ. This selection increases the fraction of negative units but does not make a recipient negative in every layer and head: selected image/question recipients average 61/51\% negative units in LLaVA-OneVision, 34/14\% in Qwen-3B, 31/11\% in Qwen-7B, and 21/4\% in Gemma. The exact query-head-weighted shares appear in \cref{tab:negative-recipients}.

\begin{table*}[t]
 \centering\small
 \setlength{\tabcolsep}{4pt}
 \caption{Recipients selected for the largest fraction of negative native coefficients: results within their $r_j<0$ units. Ratios and 95\% intervals use mean attention over later prompt rows. All conditions use $\eta=1$ except removal ($\eta=-1$). Share gives the fraction of the selected recipient\textquotesingle s units with $r_j<0$; Fall is the percentage of those units whose attention decreases under the boost.}
 \label{tab:negative-recipients}
 \begin{tabular}{llrcccr}
 \toprule
 Model & Recipient & Share (\%) & Boost & Removal & RAND & Fall (\%) \\
 \midrule
 LLaVA-OneVision-7B & image & 61.3 & $0.0123\;[0.0104,0.0145]$ & $14.3\;[13.3,15.3]$ & $1.02\;[0.999,1.05]$ & 96 \\
  & question & 50.6 & $0.105\;[0.0989,0.112]$ & $6.47\;[6.16,6.8]$ & $1.01\;[0.998,1.02]$ & 93 \\
 \midrule
 Qwen2.5-VL-3B & image & 34.1 & $0.118\;[0.106,0.131]$ & $4.81\;[4.57,5.08]$ & $1.01\;[0.99,1.02]$ & 95 \\
  & question & 13.6 & $0.169\;[0.152,0.187]$ & $3.97\;[3.63,4.3]$ & $1.03\;[1.01,1.06]$ & 95 \\
 \midrule
 Qwen2.5-VL-7B & image & 30.7 & $0.19\;[0.16,0.228]$ & $3.83\;[3.63,4.06]$ & $0.978\;[0.948,1.01]$ & 82 \\
  & question & 11.4 & $0.28\;[0.261,0.302]$ & $2.19\;[2.1,2.28]$ & $1\;[0.976,1.03]$ & 84 \\
 \midrule
 Gemma-3-4B & image & 21.5 & $0.366\;[0.292,0.466]$ & $4.18\;[3.99,4.37]$ & $1.05\;[1.03,1.08]$ & 72 \\
  & question & 3.9 & $0.205\;[0.171,0.253]$ & $2.81\;[2.7,2.92]$ & $1.01\;[0.986,1.04]$ & 90 \\
 \bottomrule
 \end{tabular}
\end{table*}

\begin{figure*}[t]
 \centering
 \includegraphics[width=\textwidth]{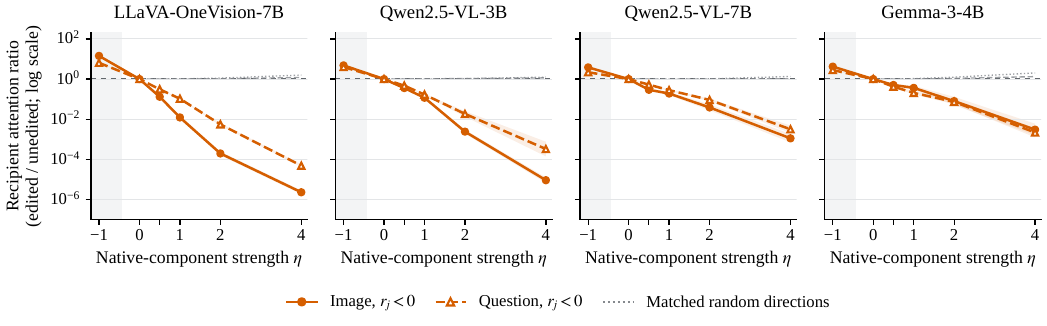}
 \caption{\textbf{Recipients selected for negative coefficients reproduce the signed response.} Curves show only units with native $r_j<0$, using the recipient-selection procedure in the text. Increasing the component lowers attention; removing it raises attention. The statistic, intervals, and grey RAND curves follow \cref{fig:key-control}c. A selected recipient can still have positive coefficients in other units, which are excluded from these curves.}
 \label{fig:negative-recipients}
\end{figure*}

Within negative units, $\eta=1$ gives ratios $0.012$--$0.37$, with decreases in 72--96\% of units; removal raises attention by $2.2$--$14.3$ (\cref{tab:negative-recipients,fig:negative-recipients}). RAND gives $0.98$--$1.05$. In LLaVA-OneVision, the selected image recipient also loses attention over all units: $0.143$ (95\% CI $[0.127,0.161]$) at $\eta=1$ and $0.016$ ($[0.014,0.019]$) at $\eta=2$. In Qwen and Gemma, the selected recipients remain positive in most units, and their all-unit ratios rise by $2.4$--$13.6$. For these models, the decrease appears in the negative-coefficient units.

\subsection{Exact Local Identity and Propagation through Depth}

At the first edited layer, queries and other keys are unchanged from the native pass, so \cref{eq:native-scaling-prob} is an identity. Maximum absolute attention errors are $3.9\times10^{-6}$ for random recipients and $3.5\times10^{-6}$ for negative-selected recipients; the run rejects any question with error above $10^{-3}$. Deeper layers no longer satisfy the fixed-query assumption. Across positive-strength arms and recipient kinds in the primary run, correlations between predicted and measured log attention ratios range from 0.91--0.97 for LLaVA-OneVision, 0.65--0.93 for Qwen-3B, 0.70--0.93 for Qwen-7B, and 0.56--0.87 for Gemma. Median measured-to-predicted changes range from 0.97--1.00, 0.53--0.99, 0.45--0.96, and 0.11--0.96, respectively. The local identity explains the edit, while the full-pass measurements establish the response after propagation.

The last-row sink score is positive in 93--100\% of question-layer-query-head units across these four models. This all-head prompt statistic is distinct from the selected-head decoding diagnostic in Appendix~\ref{sec:supp-calibration}. The sink's mean share falls by 1.7--6.8 percentage points in positive-coefficient strata at $\eta=1$, exceeding the recipient's absolute gain, so the released mass also reaches other keys. These probes measure attention movement under a controlled key edit; \cref{sec:experiments} evaluates task accuracy.

\section{Full Sampled Accuracy Results}
\label{sec:supp-full-accuracy}

\begin{table*}[t]
  \centering
  \small
  \renewcommand{\arraystretch}{1.15}
  \setlength{\tabcolsep}{4pt}
  \caption{Full core-five evaluation: test accuracy (\%). Core avg. weights the five benchmarks equally. All strengths use the validation-selected settings of the main evaluation. Superscripts on \methodname accuracies mark changes from unedited decoding: $\ast$/$\dagger$ indicate a pointwise paired-bootstrap 95\% CI above/below zero.}
  \label{tab:full-core-results}
  \begin{tabular*}{\textwidth}{@{\extracolsep{\fill}}@{}llrrrrrr@{}}
    \toprule
    \textbf{Model} & \textbf{Method} & \textbf{MMBench} & \textbf{POPE} & \textbf{CV-Bench} & \textbf{EmbSpatial} & \textbf{MMStar} & \textbf{Core avg.} \\
    \midrule
    LLaVA-OneVision-7B & Unedited & 64.2 & 61.2 & 51.2 & 51.1 & 38.1 & 53.2 \\
     & +\methodname & 70.9$^{\ast}$ & 63.1$^{\ast}$ & 54.9$^{\ast}$ & 53.0$^{\ast}$ & 43.3$^{\ast}$ & 57.1$^{\ast}$ \\
    \addlinespace[2pt]
    \midrule
    \addlinespace[2pt]
    Qwen2.5-VL-3B & Unedited & 73.8 & 80.9 & 58.2 & 47.3 & 44.7 & 61.0 \\
     & +\methodname & 74.5 & 81.9$^{\ast}$ & 61.0$^{\ast}$ & 49.4$^{\ast}$ & 45.4 & 62.4$^{\ast}$ \\
    \addlinespace[2pt]
    \midrule
    \addlinespace[2pt]
    Qwen2.5-VL-7B & Unedited & 87.3 & 86.3 & 73.7 & 66.7 & 59.4 & 74.7 \\
     & +\methodname & 87.1 & 86.5 & 74.2 & 66.9 & 60.7 & 75.1 \\
    \addlinespace[2pt]
    \midrule
    \addlinespace[2pt]
    Gemma-3-4B & Unedited & 82.6 & 84.3 & 59.4 & 53.9 & 52.4 & 66.5 \\
     & +\methodname & 82.4 & 84.2 & 60.8$^{\ast}$ & 55.0$^{\ast}$ & 52.5 & 67.0$^{\ast}$ \\
    \addlinespace[2pt]
    \midrule
    \addlinespace[2pt]
    InternVL3-8B & Unedited & 89.0 & 88.2 & 79.5 & 70.3 & 64.0 & 78.2 \\
     & +\methodname & 88.9 & 87.6$^{\dagger}$ & 78.7 & 69.8 & 66.6$^{\ast}$ & 78.3 \\
    \bottomrule
  \end{tabular*}
\end{table*}

\begin{table*}[t]
  \centering
  \small
  \renewcommand{\arraystretch}{1.15}
  \setlength{\tabcolsep}{4pt}
  \caption{Full spatial evaluation: test accuracy (\%). Spatial avg. weights these four benchmarks equally; All avg. weights all nine core and spatial benchmarks equally. Superscripts on \methodname accuracies mark changes from unedited decoding: $\ast$/$\dagger$ indicate a pointwise paired-bootstrap 95\% CI above/below zero.}
  \label{tab:full-spatial-results}
  \begin{tabular*}{\textwidth}{@{\extracolsep{\fill}}@{}llrrrrrr@{}}
    \toprule
    \textbf{Model} & \textbf{Method} & \textbf{VSR} & \textbf{WhatsUp} & \textbf{COCO-QA} & \textbf{VG-QA} & \textbf{Spatial avg.} & \textbf{All avg.} \\
    \midrule
    LLaVA-OneVision-7B & Unedited & 42.6 & 68.9 & 37.5 & 34.7 & 45.9 & 49.9 \\
     & +\methodname & 53.4$^{\ast}$ & 71.5$^{\ast}$ & 42.1$^{\ast}$ & 43.0$^{\ast}$ & 52.5$^{\ast}$ & 55.0$^{\ast}$ \\
    \addlinespace[2pt]
    \midrule
    \addlinespace[2pt]
    Qwen2.5-VL-3B & Unedited & 63.8 & 67.2 & 56.2 & 51.3 & 59.6 & 60.4 \\
     & +\methodname & 69.1$^{\ast}$ & 74.5$^{\ast}$ & 57.2 & 54.8$^{\ast}$ & 63.9$^{\ast}$ & 63.1$^{\ast}$ \\
    \addlinespace[2pt]
    \midrule
    \addlinespace[2pt]
    Qwen2.5-VL-7B & Unedited & 84.0 & 88.7 & 64.8 & 63.3 & 75.2 & 74.9 \\
     & +\methodname & 85.2$^{\ast}$ & 88.2 & 64.6 & 65.7$^{\ast}$ & 75.9$^{\ast}$ & 75.5$^{\ast}$ \\
    \addlinespace[2pt]
    \midrule
    \addlinespace[2pt]
    Gemma-3-4B & Unedited & 67.6 & 58.8 & 50.6 & 48.0 & 56.3 & 62.0 \\
     & +\methodname & 67.4 & 64.7$^{\ast}$ & 51.6$^{\ast}$ & 46.5$^{\dagger}$ & 57.5$^{\ast}$ & 62.8$^{\ast}$ \\
    \addlinespace[2pt]
    \midrule
    \addlinespace[2pt]
    InternVL3-8B & Unedited & 81.6 & 90.3 & 69.2 & 70.6 & 77.9 & 78.1 \\
     & +\methodname & 82.4 & 86.9$^{\dagger}$ & 69.1 & 72.6$^{\ast}$ & 77.7 & 78.0 \\
    \bottomrule
  \end{tabular*}
\end{table*}

\begin{table}[t]
  \centering\small
  \setlength{\tabcolsep}{4pt}
  \caption{Full five-model evaluation: test accuracy (\%) averaged equally over all 19 benchmarks. $\Delta$ is \methodname minus Base in points, computed before rounding. Bold marks the higher mean. No significance test is applied to these averages.}
  \label{tab:all19-summary}
  \begin{tabular*}{\linewidth}{@{\extracolsep{\fill}}@{}lrrc@{}}
    \toprule
    Model & Base & \methodname & $\Delta$ \\
    \midrule
    LLaVA-OneVision-7B & 52.4 & \textbf{55.9} & $+3.5$ \\
    Qwen2.5-VL-3B & 57.5 & \textbf{60.1} & $+2.7$ \\
    Qwen2.5-VL-7B & 72.7 & \textbf{73.1} & $+0.4$ \\
    Gemma-3-4B & 63.1 & \textbf{63.6} & $+0.6$ \\
    InternVL3-8B & \textbf{75.2} & 74.8 & $-0.4$ \\
    \bottomrule
  \end{tabular*}
\end{table}

\Cref{tab:full-core-results,tab:full-spatial-results} give the core and spatial results for all five sampled models, using the same validation-selected strengths and three-seed test evaluation as \cref{tab:main-results}. Together with the additional ten benchmarks in \cref{tab:additional-results}, they retain all 19 benchmark results for every model. \Cref{tab:all19-summary} gives the five models' All-19 averages. On the nine benchmarks of \cref{tab:main-results}, InternVL3-8B's mean is essentially unchanged ($-0.03$ points).

\section{Greedy Decoding}
\label{sec:supp-greedy}

\begin{table*}[t]
  \centering\footnotesize
  \setlength{\tabcolsep}{3pt}
  \caption{Greedy test-split strict accuracy (\%): unedited $\to$ \methodname, with change and independently validation-selected $\beta$ beneath each score. $\ast$/$\dagger$ denote paired-bootstrap 95\% CIs above/below zero. The comparison covers Core-5, with benchmarks weighted equally.}
  \label{tab:greedy-results}
  \begin{tabular}{lccccc}
    \toprule
    Benchmark & \shortstack{LLaVA-OV\\7B} & \shortstack{Qwen2.5-VL\\3B} & \shortstack{Qwen2.5-VL\\7B} & \shortstack{Gemma-3\\4B} & \shortstack{InternVL3\\8B} \\
    \midrule
    MMBench & \shortstack{$74.1\!\to\!81.8$\\[-1pt]{\scriptsize$(+7.7)^{\ast}\;[0.5]$}} & \shortstack{$79.4\!\to\!81.4$\\[-1pt]{\scriptsize$(+2.0)^{\ast}\;[0.5]$}} & \shortstack{$87.7\!\to\!87.8$\\[-1pt]{\scriptsize$(+0.1)\;[0.2]$}} & \shortstack{$82.8\!\to\!81.0$\\[-1pt]{\scriptsize$(-1.8)^{\dagger}\;[0.7]$}} & \shortstack{$89.3\!\to\!88.2$\\[-1pt]{\scriptsize$(-1.0)^{\dagger}\;[0.7]$}} \\
    \addlinespace[2pt]
    POPE & \shortstack{$69.5\!\to\!72.9$\\[-1pt]{\scriptsize$(+3.4)^{\ast}\;[0.3]$}} & \shortstack{$85.7\!\to\!86.4$\\[-1pt]{\scriptsize$(+0.7)^{\ast}\;[0.2]$}} & \shortstack{$86.8\!\to\!87.6$\\[-1pt]{\scriptsize$(+0.8)^{\ast}\;[0.2]$}} & \shortstack{$84.2\!\to\!84.0$\\[-1pt]{\scriptsize$(-0.2)\;[0.5]$}} & \shortstack{$88.6\!\to\!87.2$\\[-1pt]{\scriptsize$(-1.3)^{\dagger}\;[0.7]$}} \\
    \addlinespace[2pt]
    CV-Bench & \shortstack{$62.7\!\to\!61.9$\\[-1pt]{\scriptsize$(-0.8)\;[0.7]$}} & \shortstack{$66.6\!\to\!67.1$\\[-1pt]{\scriptsize$(+0.4)\;[0.3]$}} & \shortstack{$74.4\!\to\!75.8$\\[-1pt]{\scriptsize$(+1.4)\;[0.3]$}} & \shortstack{$60.3\!\to\!62.5$\\[-1pt]{\scriptsize$(+2.2)^{\ast}\;[0.2]$}} & \shortstack{$79.6\!\to\!79.5$\\[-1pt]{\scriptsize$(-0.1)\;[0.3]$}} \\
    \addlinespace[2pt]
    EmbSpatial & \shortstack{$62.9\!\to\!64.4$\\[-1pt]{\scriptsize$(+1.5)^{\ast}\;[0.5]$}} & \shortstack{$53.9\!\to\!55.9$\\[-1pt]{\scriptsize$(+2.0)^{\ast}\;[0.2]$}} & \shortstack{$69.0\!\to\!68.6$\\[-1pt]{\scriptsize$(-0.3)\;[0.2]$}} & \shortstack{$53.3\!\to\!53.8$\\[-1pt]{\scriptsize$(+0.5)\;[0.2]$}} & \shortstack{$71.3\!\to\!70.1$\\[-1pt]{\scriptsize$(-1.2)\;[0.2]$}} \\
    \addlinespace[2pt]
    MMStar & \shortstack{$43.5\!\to\!50.2$\\[-1pt]{\scriptsize$(+6.6)^{\ast}\;[0.5]$}} & \shortstack{$47.5\!\to\!49.8$\\[-1pt]{\scriptsize$(+2.2)\;[0.2]$}} & \shortstack{$61.4\!\to\!62.2$\\[-1pt]{\scriptsize$(+0.8)\;[0.2]$}} & \shortstack{$51.5\!\to\!50.6$\\[-1pt]{\scriptsize$(-0.9)\;[0.5]$}} & \shortstack{$65.5\!\to\!63.3$\\[-1pt]{\scriptsize$(-2.3)^{\dagger}\;[0.5]$}} \\
    \midrule
    Core-five avg. & \shortstack{$62.6\!\to\!66.2$\\[-1pt]{\scriptsize$(+3.7)^{\ast}$}} & \shortstack{$66.6\!\to\!68.1$\\[-1pt]{\scriptsize$(+1.5)^{\ast}$}} & \shortstack{$75.9\!\to\!76.4$\\[-1pt]{\scriptsize$(+0.6)$}} & \shortstack{$66.4\!\to\!66.4$\\[-1pt]{\scriptsize$(0.0)$}} & \shortstack{$78.8\!\to\!77.7$\\[-1pt]{\scriptsize$(-1.2)^{\dagger}$}} \\
    \bottomrule
  \end{tabular}
\end{table*}

Greedy evaluation uses one run, selecting strengths independently on each validation split. Batched bf16 inference is not bitwise deterministic, and repeated unedited greedy runs with different batch compositions differ by at most 1.0 point in per-benchmark accuracy. We therefore use greedy decoding as a complement to the three-seed sampled evaluation. The Core-5 averages improve significantly for LLaVA-OneVision and Qwen2.5-VL-3B, change within their intervals for Qwen2.5-VL-7B and Gemma-3-4B, and decrease for InternVL3-8B.

The reported greedy comparison is restricted to Core-5. The spatial results in the main evaluation use the sampled protocol.

\section{Strength Selection and Sensitivity}
\label{sec:supp-strength}

The strength grid is $\{0.3,0.5,0.7\}$ for LLaVA-OneVision, $\{0.2,0.3,0.5\}$ for Qwen2.5-VL-3B/7B, and $\{0.2,0.3,0.5,0.7\}$ for Gemma-3-4B and InternVL3-8B. Each model-benchmark pair uses the validation strength with the largest mean edited-minus-unedited accuracy change, breaking ties toward smaller strengths. Sampled selection averages three seeds; greedy selection uses one run. The test questions play no part in strength selection. Selected strengths for the greedy and additional-benchmark evaluations appear in \cref{tab:greedy-results,tab:additional-results}.

\Cref{tab:validation-strength-sensitivity} reports the original-grid validation sweeps underlying strength selection. Stronger projection can reduce accuracy: on Gemma/WhatsUp, the validation change moves from $+6.1$ points at $\beta=0.2$ to $-2.5$ at $\beta=0.7$. These sweeps support selecting partial removal on validation. Appendix~\ref{sec:supp-directions} evaluates fixed strengths across the main nine benchmarks.

\begin{table*}[p]
\centering\small
\renewcommand{\arraystretch}{0.95}
\setlength{\tabcolsep}{16pt}
\begin{tabular}{lrrrr}
\toprule
Benchmark & $\beta=0.2$ & $\beta=0.3$ & $\beta=0.5$ & $\beta=0.7$ \\
\midrule
\multicolumn{5}{l}{\textbf{LLaVA-OneVision-7B}} \\
\midrule
MMBench-EN & -- & +7.5 & +6.2 & \textbf{+8.2} \\
POPE & -- & +1.6 & +1.2 & \textbf{+2.1} \\
CV-Bench & -- & \textbf{+1.1} & +0.4 & +1.1 \\
EmbSpatial & -- & +2.6 & \textbf{+3.2} & +2.3 \\
MMStar & -- & +2.2 & \textbf{+4.7} & +0.8 \\
VSR & -- & +4.0 & +8.4 & \textbf{+11.3} \\
WhatsUp & -- & \textbf{+3.9} & +3.1 & +1.6 \\
COCO-QA & -- & \textbf{+4.7} & +3.2 & +4.3 \\
VG-QA & -- & +3.7 & +5.1 & \textbf{+7.1} \\
\midrule
\multicolumn{5}{l}{\textbf{Qwen2.5-VL-3B}} \\
\midrule
MMBench-EN & \textbf{+2.9} & +2.4 & +1.0 & -- \\
POPE & +0.9 & \textbf{+1.1} & -1.1 & -- \\
CV-Bench & +4.8 & \textbf{+5.4} & +5.1 & -- \\
EmbSpatial & \textbf{+3.1} & +1.2 & +2.0 & -- \\
MMStar & \textbf{+0.7} & +0.7 & -1.1 & -- \\
VSR & +6.9 & +6.4 & \textbf{+7.9} & -- \\
WhatsUp & +0.4 & \textbf{+4.5} & +2.9 & -- \\
COCO-QA & +2.2 & \textbf{+3.8} & +2.9 & -- \\
VG-QA & \textbf{+5.1} & +2.4 & +4.7 & -- \\
\midrule
\multicolumn{5}{l}{\textbf{Qwen2.5-VL-7B}} \\
\midrule
MMBench-EN & -0.9 & -0.8 & \textbf{-0.5} & -- \\
POPE & +0.1 & \textbf{+0.9} & +0.9 & -- \\
CV-Bench & +0.1 & \textbf{+0.4} & -0.5 & -- \\
EmbSpatial & \textbf{+1.5} & +0.6 & +0.1 & -- \\
MMStar & -0.5 & \textbf{+0.6} & +0.1 & -- \\
VSR & -0.4 & \textbf{+0.5} & +0.3 & -- \\
WhatsUp & \textbf{+1.4} & -0.8 & -0.8 & -- \\
COCO-QA & -0.2 & \textbf{+0.3} & +0.1 & -- \\
VG-QA & -1.1 & \textbf{+0.5} & +0.5 & -- \\
\midrule
\multicolumn{5}{l}{\textbf{Gemma-3-4B}} \\
\midrule
MMBench-EN & \textbf{+0.3} & -0.6 & -0.2 & -1.1 \\
POPE & +0.2 & +0.7 & \textbf{+0.7} & -0.1 \\
CV-Bench & +1.3 & \textbf{+1.9} & +1.6 & -0.3 \\
EmbSpatial & \textbf{+1.2} & +1.0 & -0.4 & -1.3 \\
MMStar & -0.6 & \textbf{+0.6} & +0.4 & -1.2 \\
VSR & +1.0 & +0.9 & +1.0 & \textbf{+1.2} \\
WhatsUp & \textbf{+6.1} & +3.9 & -0.4 & -2.5 \\
COCO-QA & \textbf{+1.1} & -0.2 & -0.6 & -4.3 \\
VG-QA & \textbf{-1.0} & -3.0 & -4.0 & -6.0 \\
\midrule
\multicolumn{5}{l}{\textbf{InternVL3-8B}} \\
\midrule
MMBench-EN & \textbf{+0.2} & -0.5 & -0.2 & -1.0 \\
POPE & +0.5 & +0.2 & +0.8 & \textbf{+0.8} \\
CV-Bench & \textbf{-0.9} & -1.5 & -1.0 & -2.9 \\
EmbSpatial & -0.3 & \textbf{+0.5} & -2.1 & -1.6 \\
MMStar & -1.0 & \textbf{-0.2} & -3.4 & -2.0 \\
VSR & -0.2 & \textbf{+0.8} & +0.0 & -0.7 \\
WhatsUp & -1.8 & \textbf{-0.4} & -3.1 & -4.7 \\
COCO-QA & \textbf{-0.1} & -0.4 & -0.8 & -2.2 \\
VG-QA & -3.1 & -3.4 & \textbf{-1.8} & -3.4 \\
\bottomrule
\end{tabular}
\caption{Original-grid validation sensitivity: mean accuracy change (percentage points) from unedited decoding. Bold: validation-selected strength. Dashes indicate strengths outside the model\textquotesingle s original grid.}
\label{tab:validation-strength-sensitivity}
\end{table*}

\section{Additional Benchmarks}
\label{sec:supp-additional}

\begin{table*}[t]
  \centering\footnotesize
  \setlength{\tabcolsep}{3pt}
  \caption{Additional test-split strict accuracy (\%): unedited $\to$ \methodname, mean $\pm$ standard deviation. Each second line gives the change in points and validation-selected $\beta$. The protocol and significance symbols match \cref{tab:main-results}. MME and HallusionBench use per-question lettered Yes/No accuracy; MMVP uses per-question accuracy. BLINK and MMMU include only single-image multiple-choice items.}
  \label{tab:additional-results}
  \begin{tabular}{lccccc}
    \toprule
    Benchmark & \shortstack{LLaVA-OV\\7B} & \shortstack{Qwen2.5-VL\\3B} & \shortstack{Qwen2.5-VL\\7B} & \shortstack{Gemma-3\\4B} & \shortstack{InternVL3\\8B} \\
    \midrule
    M3CoT & \shortstack{$41.8{\scriptstyle\pm 0.3}\!\to\!43.6{\scriptstyle\pm 1.6}$\\[-1pt]{\scriptsize$(+1.8)\;[0.3]$}} & \shortstack{$42.5{\scriptstyle\pm 0.7}\!\to\!43.6{\scriptstyle\pm 0.9}$\\[-1pt]{\scriptsize$(+1.0)\;[0.2]$}} & \shortstack{$61.4{\scriptstyle\pm 0.5}\!\to\!61.7{\scriptstyle\pm 1.0}$\\[-1pt]{\scriptsize$(+0.3)\;[0.3]$}} & \shortstack{$50.7{\scriptstyle\pm 1.6}\!\to\!50.9{\scriptstyle\pm 1.6}$\\[-1pt]{\scriptsize$(+0.2)\;[0.5]$}} & \shortstack{$56.9{\scriptstyle\pm 1.6}\!\to\!57.7{\scriptstyle\pm 0.8}$\\[-1pt]{\scriptsize$(+0.8)\;[0.3]$}} \\
    \addlinespace[2pt]
    AI2D & \shortstack{$59.7{\scriptstyle\pm 1.0}\!\to\!61.8{\scriptstyle\pm 1.2}$\\[-1pt]{\scriptsize$(+2.1)^{\ast}\;[0.3]$}} & \shortstack{$52.3{\scriptstyle\pm 1.2}\!\to\!56.4{\scriptstyle\pm 1.1}$\\[-1pt]{\scriptsize$(+4.0)^{\ast}\;[0.2]$}} & \shortstack{$75.2{\scriptstyle\pm 0.6}\!\to\!75.4{\scriptstyle\pm 1.0}$\\[-1pt]{\scriptsize$(+0.2)\;[0.2]$}} & \shortstack{$71.7{\scriptstyle\pm 0.2}\!\to\!71.5{\scriptstyle\pm 0.0}$\\[-1pt]{\scriptsize$(-0.2)\;[0.2]$}} & \shortstack{$78.6{\scriptstyle\pm 0.4}\!\to\!79.3{\scriptstyle\pm 0.3}$\\[-1pt]{\scriptsize$(+0.7)\;[0.2]$}} \\
    \addlinespace[2pt]
    ScienceQA-IMG & \shortstack{$74.1{\scriptstyle\pm 0.5}\!\to\!78.3{\scriptstyle\pm 0.8}$\\[-1pt]{\scriptsize$(+4.2)^{\ast}\;[0.5]$}} & \shortstack{$66.1{\scriptstyle\pm 1.1}\!\to\!69.0{\scriptstyle\pm 1.5}$\\[-1pt]{\scriptsize$(+2.9)^{\ast}\;[0.3]$}} & \shortstack{$87.8{\scriptstyle\pm 0.7}\!\to\!88.1{\scriptstyle\pm 0.7}$\\[-1pt]{\scriptsize$(+0.3)\;[0.2]$}} & \shortstack{$80.8{\scriptstyle\pm 0.4}\!\to\!81.3{\scriptstyle\pm 0.0}$\\[-1pt]{\scriptsize$(+0.4)\;[0.2]$}} & \shortstack{$94.7{\scriptstyle\pm 0.2}\!\to\!94.4{\scriptstyle\pm 0.5}$\\[-1pt]{\scriptsize$(-0.3)\;[0.3]$}} \\
    \addlinespace[2pt]
    A-OKVQA & \shortstack{$68.8{\scriptstyle\pm 2.0}\!\to\!72.9{\scriptstyle\pm 1.4}$\\[-1pt]{\scriptsize$(+4.1)^{\ast}\;[0.5]$}} & \shortstack{$66.0{\scriptstyle\pm 0.3}\!\to\!70.9{\scriptstyle\pm 2.2}$\\[-1pt]{\scriptsize$(+4.9)^{\ast}\;[0.5]$}} & \shortstack{$84.4{\scriptstyle\pm 0.5}\!\to\!84.8{\scriptstyle\pm 0.4}$\\[-1pt]{\scriptsize$(+0.4)\;[0.3]$}} & \shortstack{$79.8{\scriptstyle\pm 1.0}\!\to\!79.3{\scriptstyle\pm 0.9}$\\[-1pt]{\scriptsize$(-0.4)\;[0.3]$}} & \shortstack{$85.2{\scriptstyle\pm 0.6}\!\to\!85.0{\scriptstyle\pm 1.1}$\\[-1pt]{\scriptsize$(-0.2)\;[0.2]$}} \\
    \addlinespace[2pt]
    MMVP & \shortstack{$60.3{\scriptstyle\pm 3.0}\!\to\!60.8{\scriptstyle\pm 1.7}$\\[-1pt]{\scriptsize$(+0.6)\;[0.3]$}} & \shortstack{$58.0{\scriptstyle\pm 0.5}\!\to\!58.5{\scriptstyle\pm 2.2}$\\[-1pt]{\scriptsize$(+0.6)\;[0.3]$}} & \shortstack{$73.8{\scriptstyle\pm 1.3}\!\to\!74.6{\scriptstyle\pm 2.0}$\\[-1pt]{\scriptsize$(+0.8)\;[0.5]$}} & \shortstack{$68.4{\scriptstyle\pm 1.1}\!\to\!69.4{\scriptstyle\pm 1.9}$\\[-1pt]{\scriptsize$(+1.0)\;[0.2]$}} & \shortstack{$76.9{\scriptstyle\pm 1.3}\!\to\!74.4{\scriptstyle\pm 3.7}$\\[-1pt]{\scriptsize$(-2.6)\;[0.5]$}} \\
    \addlinespace[2pt]
    RealWorldQA & \shortstack{$46.1{\scriptstyle\pm 2.5}\!\to\!48.3{\scriptstyle\pm 1.9}$\\[-1pt]{\scriptsize$(+2.2)\;[0.3]$}} & \shortstack{$47.2{\scriptstyle\pm 1.1}\!\to\!49.7{\scriptstyle\pm 0.8}$\\[-1pt]{\scriptsize$(+2.4)\;[0.2]$}} & \shortstack{$60.5{\scriptstyle\pm 1.9}\!\to\!61.8{\scriptstyle\pm 3.1}$\\[-1pt]{\scriptsize$(+1.3)\;[0.2]$}} & \shortstack{$53.4{\scriptstyle\pm 1.0}\!\to\!55.5{\scriptstyle\pm 1.8}$\\[-1pt]{\scriptsize$(+2.1)\;[0.5]$}} & \shortstack{$65.5{\scriptstyle\pm 1.2}\!\to\!61.3{\scriptstyle\pm 1.5}$\\[-1pt]{\scriptsize$(-4.2)^{\dagger}\;[0.7]$}} \\
    \addlinespace[2pt]
    MME & \shortstack{$66.9{\scriptstyle\pm 0.8}\!\to\!68.3{\scriptstyle\pm 1.1}$\\[-1pt]{\scriptsize$(+1.4)\;[0.3]$}} & \shortstack{$74.4{\scriptstyle\pm 0.3}\!\to\!76.4{\scriptstyle\pm 0.7}$\\[-1pt]{\scriptsize$(+2.0)^{\ast}\;[0.3]$}} & \shortstack{$84.5{\scriptstyle\pm 1.0}\!\to\!85.1{\scriptstyle\pm 0.4}$\\[-1pt]{\scriptsize$(+0.6)\;[0.5]$}} & \shortstack{$79.4{\scriptstyle\pm 0.1}\!\to\!79.9{\scriptstyle\pm 0.7}$\\[-1pt]{\scriptsize$(+0.6)\;[0.2]$}} & \shortstack{$87.2{\scriptstyle\pm 0.2}\!\to\!87.0{\scriptstyle\pm 0.4}$\\[-1pt]{\scriptsize$(-0.2)\;[0.2]$}} \\
    \addlinespace[2pt]
    HallusionBench & \shortstack{$51.9{\scriptstyle\pm 0.5}\!\to\!52.8{\scriptstyle\pm 0.6}$\\[-1pt]{\scriptsize$(+0.8)\;[0.5]$}} & \shortstack{$56.7{\scriptstyle\pm 1.4}\!\to\!60.2{\scriptstyle\pm 0.5}$\\[-1pt]{\scriptsize$(+3.5)^{\ast}\;[0.5]$}} & \shortstack{$68.9{\scriptstyle\pm 0.1}\!\to\!68.9{\scriptstyle\pm 0.6}$\\[-1pt]{\scriptsize$(0.0)\;[0.2]$}} & \shortstack{$60.4{\scriptstyle\pm 1.0}\!\to\!62.5{\scriptstyle\pm 0.3}$\\[-1pt]{\scriptsize$(+2.1)^{\ast}\;[0.2]$}} & \shortstack{$66.2{\scriptstyle\pm 0.9}\!\to\!67.2{\scriptstyle\pm 1.0}$\\[-1pt]{\scriptsize$(+1.0)\;[0.3]$}} \\
    \addlinespace[2pt]
    BLINK & \shortstack{$40.9{\scriptstyle\pm 1.5}\!\to\!40.9{\scriptstyle\pm 0.8}$\\[-1pt]{\scriptsize$(+0.1)\;[0.7]$}} & \shortstack{$45.4{\scriptstyle\pm 1.9}\!\to\!47.0{\scriptstyle\pm 2.1}$\\[-1pt]{\scriptsize$(+1.6)\;[0.5]$}} & \shortstack{$55.9{\scriptstyle\pm 1.7}\!\to\!55.9{\scriptstyle\pm 2.6}$\\[-1pt]{\scriptsize$(-0.1)\;[0.5]$}} & \shortstack{$48.3{\scriptstyle\pm 0.5}\!\to\!47.3{\scriptstyle\pm 1.6}$\\[-1pt]{\scriptsize$(-1.0)\;[0.7]$}} & \shortstack{$55.0{\scriptstyle\pm 3.5}\!\to\!54.4{\scriptstyle\pm 1.4}$\\[-1pt]{\scriptsize$(-0.6)\;[0.2]$}} \\
    \addlinespace[2pt]
    MMMU & \shortstack{$34.9{\scriptstyle\pm 1.0}\!\to\!39.0{\scriptstyle\pm 1.5}$\\[-1pt]{\scriptsize$(+4.1)^{\ast}\;[0.3]$}} & \shortstack{$40.1{\scriptstyle\pm 1.2}\!\to\!43.1{\scriptstyle\pm 0.5}$\\[-1pt]{\scriptsize$(+3.0)^{\ast}\;[0.3]$}} & \shortstack{$55.1{\scriptstyle\pm 0.5}\!\to\!53.9{\scriptstyle\pm 0.8}$\\[-1pt]{\scriptsize$(-1.2)\;[0.5]$}} & \shortstack{$47.6{\scriptstyle\pm 1.1}\!\to\!46.3{\scriptstyle\pm 0.9}$\\[-1pt]{\scriptsize$(-1.3)\;[0.5]$}} & \shortstack{$60.4{\scriptstyle\pm 1.2}\!\to\!58.8{\scriptstyle\pm 1.6}$\\[-1pt]{\scriptsize$(-1.6)\;[0.5]$}} \\
    \bottomrule
  \end{tabular}
\end{table*}

We evaluate M3CoT~\cite{chen2024m3cot}, AI2D~\cite{kembhavi2016ai2d}, ScienceQA-IMG~\cite{lu2022scienceqa}, A-OKVQA~\cite{schwenk2022aokvqa}, MMVP~\cite{tong2024mmvp}, RealWorldQA~\cite{xai2024realworldqa}, MME~\cite{fu2023mme}, HallusionBench~\cite{guan2024hallusionbench}, BLINK~\cite{fu2024blink}, and MMMU~\cite{yue2024mmmu}. RealWorldQA uses its multiple-choice questions; BLINK and MMMU include only single-image multiple-choice items. Every cell follows the validation-to-test selection protocol of the main evaluation.

No M3CoT change is significant. Across the other nine benchmarks, LLaVA-OneVision and Qwen2.5-VL-3B improve in every cell, significantly on four and six benchmarks.

Qwen2.5-VL-7B has no significant change on these nine benchmarks. Gemma-3-4B gains $2.10$ points significantly on HallusionBench, with no significant loss. InternVL3-8B loses $4.18$ points significantly on RealWorldQA and has no significant gain. The small validation splits of MMVP and RealWorldQA make strength selection less precise; we retain their validation-selected strengths.

\section{Sink-Only Comparison}
\label{sec:supp-breadth}

This accuracy comparison covers the five main models on Core-5. The additional checkpoints (LLaVA-1.5-7B~\cite{liu2024improved}, Qwen3-VL-4B~\cite{bai2025qwen3vl}, InternVL3.5-8B~\cite{wang2025internvl35}, and Gemma-3-12B~\cite{gemmateam2025gemma3}) are included in the native-geometry diagnostics of Appendix~\ref{sec:supp-calibration}; no extended greedy accuracy comparison is reported here.

\paragraph{Sink-only comparison.}
The sink-only control applies the same projection to the sink key alone, retaining \methodname's eligible layers, native-attention head selector, and decoding onset. It can alter the sink's competition with content positions but preserves non-sink attention ratios within each edited row. Both interventions use the original strength grids, with strengths selected independently by validation accuracy before testing on the same core-five questions.

\Cref{tab:sink-only-control} reports all five models and benchmarks at the selected strengths. Each intervention's gain is measured against its own unedited baseline run; the contrast is the difference of gains. Pointwise 95\% CIs use 10,000 paired test-question bootstrap resamples (seed 0) of per-question mean gain differences. Core-five intervals resample each benchmark separately and weight their means equally.

The mean gain difference favors \methodname on LLaVA-OneVision, Qwen2.5-VL-3B, and Gemma, with intervals above zero. Qwen2.5-VL-7B and InternVL3 intervals include zero. Individual tasks also differ: the control has a supported advantage on InternVL3/POPE. These results support a benefit of all-key editing beyond editing only the sink on these three models.

\begin{table*}[p]
\centering\small
\setlength{\tabcolsep}{9pt}
\begin{tabular}{lcrrl}
\toprule
Benchmark & $\beta$: \methodname/control & \methodname $\Delta$ & Control $\Delta$ & Difference of gains [95\% CI] \\
\midrule
\multicolumn{5}{l}{\textbf{LLaVA-OneVision-7B}} \\
\midrule
MMBench-EN & 0.7/0.5 & $+6.73$ & $+4.04$ & $+2.69$ $[+1.36, +3.98]$ \\
POPE & 0.7/0.3 & $+1.94$ & $+0.94$ & $+1.01$ $[+0.12, +1.88]$ \\
CV-Bench & 0.3/0.5 & $+3.63$ & $+1.56$ & $+2.07$ $[+0.39, +3.73]$ \\
EmbSpatial & 0.5/0.7 & $+1.95$ & $+3.15$ & $-1.20$ $[-2.64, +0.23]$ \\
MMStar & 0.5/0.7 & $+5.24$ & $+2.26$ & $+2.97$ $[+0.68, +5.24]$ \\
\cmidrule(lr){1-5}
Core-5 avg. & -- & $+3.90$ & $+2.39$ & $+1.51$ $[+0.81, +2.22]$ \\
\midrule
\multicolumn{5}{l}{\textbf{Qwen2.5-VL-3B}} \\
\midrule
MMBench-EN & 0.2/0.5 & $+0.70$ & $-0.46$ & $+1.16$ $[+0.00, +2.28]$ \\
POPE & 0.3/0.5 & $+1.01$ & $+1.43$ & $-0.42$ $[-1.06, +0.23]$ \\
CV-Bench & 0.3/0.5 & $+2.80$ & $+1.04$ & $+1.75$ $[+0.11, +3.38]$ \\
EmbSpatial & 0.2/0.5 & $+2.07$ & $+1.01$ & $+1.06$ $[-0.35, +2.47]$ \\
MMStar & 0.2/0.5 & $+0.74$ & $+0.14$ & $+0.60$ $[-1.45, +2.73]$ \\
\cmidrule(lr){1-5}
Core-5 avg. & -- & $+1.46$ & $+0.63$ & $+0.83$ $[+0.17, +1.50]$ \\
\midrule
\multicolumn{5}{l}{\textbf{Qwen2.5-VL-7B}} \\
\midrule
MMBench-EN & 0.5/0.2 & $-0.20$ & $+0.02$ & $-0.22$ $[-0.90, +0.46]$ \\
POPE & 0.3/0.3 & $+0.19$ & $+0.20$ & $-0.01$ $[-0.36, +0.33]$ \\
CV-Bench & 0.3/0.3 & $+0.41$ & $+0.19$ & $+0.22$ $[-0.96, +1.42]$ \\
EmbSpatial & 0.2/0.5 & $+0.21$ & $-0.56$ & $+0.77$ $[-0.37, +1.92]$ \\
MMStar & 0.3/0.3 & $+1.34$ & $+1.55$ & $-0.22$ $[-1.96, +1.55]$ \\
\cmidrule(lr){1-5}
Core-5 avg. & -- & $+0.39$ & $+0.28$ & $+0.11$ $[-0.39, +0.62]$ \\
\midrule
\multicolumn{5}{l}{\textbf{Gemma-3-4B}} \\
\midrule
MMBench-EN & 0.2/0.7 & $-0.20$ & $-0.02$ & $-0.18$ $[-0.82, +0.46]$ \\
POPE & 0.5/0.2 & $-0.09$ & $-0.15$ & $+0.06$ $[-0.30, +0.43]$ \\
CV-Bench & 0.3/0.7 & $+1.45$ & $-0.73$ & $+2.18$ $[+0.71, +3.65]$ \\
EmbSpatial & 0.2/0.5 & $+1.18$ & $-0.50$ & $+1.68$ $[+0.58, +2.78]$ \\
MMStar & 0.3/0.5 & $+0.08$ & $+0.05$ & $+0.03$ $[-1.69, +1.75]$ \\
\cmidrule(lr){1-5}
Core-5 avg. & -- & $+0.48$ & $-0.27$ & $+0.75$ $[+0.20, +1.29]$ \\
\midrule
\multicolumn{5}{l}{\textbf{InternVL3-8B}} \\
\midrule
MMBench-EN & 0.2/0.5 & $-0.12$ & $-0.10$ & $-0.02$ $[-0.58, +0.56]$ \\
POPE & 0.7/0.7 & $-0.66$ & $+0.00$ & $-0.66$ $[-1.06, -0.25]$ \\
CV-Bench & 0.2/0.3 & $-0.84$ & $-0.27$ & $-0.57$ $[-1.77, +0.60]$ \\
EmbSpatial & 0.3/0.2 & $-0.53$ & $+0.08$ & $-0.61$ $[-1.74, +0.53]$ \\
MMStar & 0.3/0.5 & $+2.56$ & $+0.65$ & $+1.91$ $[+0.16, +3.66]$ \\
\cmidrule(lr){1-5}
Core-5 avg. & -- & $+0.09$ & $+0.07$ & $+0.01$ $[-0.49, +0.52]$ \\
\bottomrule
\end{tabular}
\caption{Sink-only control under the main sampled protocol. Values are test-accuracy changes in percentage points against each run\textquotesingle s own unedited baseline. Strengths are selected independently on validation. The final column is \methodname's gain minus the control\textquotesingle s gain, with pointwise paired-question bootstrap 95\% CIs. Core-5 means weight benchmarks equally.}
\label{tab:sink-only-control}
\end{table*}

\section{Sink-Direction Editing in Existing Methods}
\label{sec:supp-host-methods}

This study applies the directional control from \cref{sec:sink-dose} within GIFT~\cite{qi2026gift} and ACG~\cite{jo2026acg}. Their intervention rules differ from \methodname's all-visible-key projection and 20\% head selector. The results in \cref{tab:host_swap} therefore evaluate integrations of the finding, rather than the standalone \allvar method.

\subsection{Evaluation Protocol}

\paragraph{Models and inputs.}
We use LLaVA-1.5-7B/13B~\cite{liu2024improved} and BakLLaVA-7B~\cite{skunkworks2023bakllava}, whose language backbone is Mistral-7B~\cite{jiang2023mistral}. All three use the LLaVA-1.5 chat template with its system line and one user image/question turn, matching GIFT's prompt. BakLLaVA uses this template rather than its hub template. Evaluation uses bf16, greedy decoding, at most 10 new tokens, and one question at a time, with one generation per arm and question. Intervals resample test questions, not generation seeds. Unlike the letter-decision scoring in \cref{sec:exp-setup}, we use the LLaVA POPE yes/no rule: take the output's first sentence, remove commas, and predict ``no'' if its space-separated words contain \texttt{no}, \texttt{No}, or \texttt{not}; otherwise predict ``yes.'' Accuracy is the fraction matching the gold yes/no label.

\paragraph{Benchmarks and splits.}
POPE-A-OKVQA and POPE-GQA~\cite{li2023pope,schwenk2022aokvqa,hudson2019gqa} pool random, popular, and adversarial questions from the VCD release~\cite{leng2024vcd}. We split by image, assigning a group to validation when the hash of its dataset and image identifier modulo five is zero. The test sets contain 7,452 and 7,128 questions. The AMBER discriminative set~\cite{wang2023amber} contributes all 14,216 questions over 1,004 images and is never used for selection. We also evaluate POPE-COCO (7,344 test questions) and MME (1,948 test questions; per-question yes/no accuracy). Their results are retained in \cref{tab:host-swap-additional}. The same questions are used for every method and model.

\paragraph{Selection and uncertainty.}
Settings maximize mean validation accuracy over POPE-COCO, POPE-A-OKVQA, and POPE-GQA, with ties favoring smaller edits. AMBER and all test questions are excluded. We first reproduce the released GIFT configuration and independently select its key-edit strength. For the matched comparison, GIFT's saliency exponent is also tuned on the same validation pools. ACG's guidance strength is selected on validation; the matched key variant retains that strength while tuning its branch edit. The exploratory image-bootstrap comparisons use 10,000 resamples, seed 20261102; matched-comparison CIs resample paired questions within each benchmark, with 10,000 resamples and seed 0. Means are computed before rounding. All intervals are pointwise and conditional on the selected settings.

\subsection{Directional Operation and Integrations}

We identify the sink once per model on 24 POPE-COCO validation questions, taking the pre-image template position with the largest mean last-row attention over the final third of layers. All three models select position zero, \texttt{<s>}. For the relevant layer and head, its native post-RoPE key defines $\usink=\ksink/\lVert\ksink\rVert_2$. With a token-specific strength $x_j$, the local operation is
\begin{equation}
 \kk'_j=\kk_j+x_j(\kk_j^\top\usink)\usink.
 \label{eq:host-key-edit}
\end{equation}
It rescales the existing parallel component by $1+x_j$, without directly editing the query, value, or orthogonal key component. Negative strengths between $-1$ and zero shrink the component; $x_j=-1$ removes it, and $x_j<-1$ reverses its sign. The latter settings exceed \methodname's bounded shrinkage range. Attention changes depend on query-key alignment, as in \cref{sec:sink-dose}.

\paragraph{GIFT.}
We retain the released saliency calculation, enhancement layers, last-query-row scope, and image/question head and token selections. The saliency map averages positive visual-attention shifts over content words and the top half of heads, clips at three standard deviations, and normalizes to $h_j\in[0,h_{\max}]$ with $h_{\max}=5$. Its original image reweighting uses $S_j=\exp(h_j)$. Our replacement uses \cref{eq:host-key-edit} with $x_j=\lambda h_j/h_{\max}$ on image keys in the selected image heads. On question keys in the selected question heads, it uses $x_j=\lambda\log(q_s)/h_{\max}$, where $q_s$ is GIFT's ratio of enhanced to native selected image mass. We preserve these weighting rules, replacing the attention operation rather than adding an extra saliency selector. GIFT uses saliency layer 11 and enhancement layers 12--22 for the 7B-depth models; for 13B these are layer 10 and layers 14--20. BakLLaVA reuses the 7B settings. Token positions for its question are located directly to account for tokenizer differences.

\paragraph{ACG.}
We retain its conditional attention output, last-query-row scope in every layer/head, and orthogonalized contrastive guidance. Original ACG defines its contrast branch by masking all image logits. Our contrast branch instead applies \cref{eq:host-key-edit} to every image key with a shared strength $x$. Both branches use the original values. Guidance uses the difference between the conditional and modified outputs, removes its component along the modified output, and adds the residual to the conditional output with scale $\gamma$, as in ACG. The exploratory $x$ values are negative; the matched 13B configuration instead selects positive $x$. A key-edited contrast branch is therefore not generally a softer approximation to image masking.

\begin{center}
\begin{minipage}{\linewidth}
 \captionsetup{type=table}
 \centering\small
 \caption{Exploratory settings for the released-host comparisons and attention traces, distinct from the matched settings below.}
 \label{tab:host-settings}
 \setlength{\tabcolsep}{3pt}
 \begin{tabular}{lrrrr}
  \toprule
  Model & GIFT $\lambda$ & ACG $\gamma$ & Key branch $x$ & $\gamma$ \\
  \midrule
  LLaVA-1.5-7B & $-1$ & 2.4 & $-0.75$ & 2.4 \\
  LLaVA-1.5-13B & $-1.25$ & 4 & $-1$ & 2.4 \\
  BakLLaVA-7B & $-2$ & 4 & $-0.75$ & 2.4 \\
  \bottomrule
 \end{tabular}
\end{minipage}
\end{center}

The GIFT edit grid for LLaVA-1.5 includes both signs of $\{0.1,0.25,0.5,0.75,1,1.25,1.5,2,4,8\}$. BakLLaVA uses $\{0.5,1,-0.25,-0.5,-0.75,-1,-1.25,-1.5,-2\}$. ACG's original guidance grid is $\{1,1.5,2,2.4,3,4,5,6,8\}$ for LLaVA-1.5 and $\{1,1.5,2,2.4,3,4,5\}$ for BakLLaVA. Key-edited ACG explores positive and negative strengths, including $\{0.5,1,2,4,8,16,-0.5,-0.75,-1,-1.5,-2,-3,-4\}$ for LLaVA-1.5, alongside auxiliary strength/guidance pairs. Tables~\ref{tab:host-settings} and~\ref{tab:host-settings-matched} specify the reported configurations. The exploratory 13B/BakLLaVA ACG comparisons include a guidance-scale difference; matched comparisons use the same guidance on both sides.

\paragraph{Matched tuning.}
GIFT's saliency exponent is selected from 21 validation candidates spanning $0.5$--$25$, with selected values $10$, $25$, and $9$ for 7B, 13B, and BakLLaVA. The 13B pick lies at the grid boundary, with a flat validation response over $15$--$25$. The key-edited GIFT configurations are unchanged. For matched ACG, both sides use $\gamma=2.4$ on 7B and $4$ on 13B/BakLLaVA. The latter key branches select $x=2$ and $x=-0.25$ from 15 candidates spanning $-4$--$16$; 7B retains $x=-0.75$. All test arms for each matched comparison share a run; repeated released-GIFT, GIFT-Key, and ACG-$4$ accuracies agree with the exploratory runs.

\begin{center}
\begin{minipage}{\linewidth}
\captionsetup{type=table}\centering\small
\caption{Settings for the matched host comparisons. The same ACG guidance scale is used on both sides.}
\label{tab:host-settings-matched}
\setlength{\tabcolsep}{3pt}
\begin{tabular}{lrrrr}
\toprule
Model & GIFT exponent & Key $\lambda$ & ACG $\gamma$ & Key $x$ \\
\midrule
LLaVA-1.5-7B & 10 & $-1$ & 2.4 & $-0.75$ \\
LLaVA-1.5-13B & 25 & $-1.25$ & 4 & 2 \\
BakLLaVA-7B & 9 & $-2$ & 4 & $-0.25$ \\
\bottomrule
\end{tabular}
\end{minipage}
\end{center}

\paragraph{Reproduction details.}
Our GIFT implementation follows the released code at commit \texttt{e20b8a9}; ACG follows \texttt{dee49c5}, whose masking and orthogonalized-guidance construction is also described in the paper's third version. All methods use bf16 here, whereas the original LLaVA implementations use fp16. ACG's branch outputs and guidance are computed in fp32 before casting back. GIFT's prompt saliency pass is retained. Because precision, input files, and splits differ, reproduced accuracies differ from those reported in the original papers.

\subsection{Results and Selection Scope}

\begin{table*}[t]
  \centering\small
  \setlength{\tabcolsep}{3pt}
  \caption{Accuracy differences (percentage points) for the exploratory configurations in \cref{tab:host-settings}. Brackets show paired-image bootstrap 95\% intervals (10,000 resamples). The three columns compare each original method with unedited decoding, its key-edited variant with unedited decoding, and the variant with the original method.}
  \label{tab:host-swap-intervals}
  \begin{tabular*}{\textwidth}{@{\extracolsep{\fill}}@{}p{72pt}llrrr@{}}
    \toprule
    Model & Method & Benchmark & Method $-$ base & Key edit $-$ base & Key edit $-$ method \\
    \midrule
    \multirow{6}{72pt}{\shortstack{LLaVA-1.5\\7B}} & \multirow{3}{*}{GIFT} & POPE-A-OKVQA & $+2.48\,[+1.89,+3.07]$ & $+3.41\,[+2.74,+4.08]$ & $+0.93\,[+0.50,+1.37]$ \\
     &  & POPE-GQA & $+3.73\,[+3.06,+4.43]$ & $+5.19\,[+4.43,+5.96]$ & $+1.46\,[+0.95,+1.96]$ \\
     &  & AMBER & $+4.65\,[+4.17,+5.13]$ & $+4.83\,[+4.28,+5.38]$ & $+0.18\,[-0.19,+0.56]$ \\
     & \multirow{3}{*}{ACG} & POPE-A-OKVQA & $+2.38\,[+1.74,+2.99]$ & $+3.15\,[+2.44,+3.86]$ & $+0.78\,[+0.15,+1.41]$ \\
     &  & POPE-GQA & $+3.65\,[+2.95,+4.35]$ & $+4.69\,[+3.89,+5.50]$ & $+1.04\,[+0.35,+1.74]$ \\
     &  & AMBER & $+4.14\,[+3.72,+4.56]$ & $+5.11\,[+4.51,+5.72]$ & $+0.98\,[+0.46,+1.52]$ \\
    \midrule
    \multirow{6}{72pt}{\shortstack{LLaVA-1.5\\13B}} & \multirow{3}{*}{GIFT} & POPE-A-OKVQA & $+0.47\,[+0.09,+0.86]$ & $+2.60\,[+1.95,+3.29]$ & $+2.13\,[+1.50,+2.76]$ \\
     &  & POPE-GQA & $+1.16\,[+0.74,+1.60]$ & $+4.52\,[+3.82,+5.23]$ & $+3.35\,[+2.71,+3.98]$ \\
     &  & AMBER & $+2.75\,[+2.39,+3.11]$ & $+6.47\,[+5.92,+7.02]$ & $+3.72\,[+3.26,+4.18]$ \\
     & \multirow{3}{*}{ACG} & POPE-A-OKVQA & $+3.52\,[+2.59,+4.46]$ & $+5.93\,[+5.01,+6.87]$ & $+2.42\,[+1.72,+3.13]$ \\
     &  & POPE-GQA & $+5.12\,[+4.25,+6.00]$ & $+8.89\,[+7.91,+9.85]$ & $+3.77\,[+2.97,+4.57]$ \\
     &  & AMBER & $+4.43\,[+3.90,+4.96]$ & $+10.40\,[+9.78,+11.04]$ & $+5.97\,[+5.44,+6.50]$ \\
    \midrule
    \multirow{6}{72pt}{\shortstack{BakLLaVA\\7B}} & \multirow{3}{*}{GIFT} & POPE-A-OKVQA & $+4.64\,[+3.85,+5.43]$ & $+5.65\,[+4.72,+6.59]$ & $+1.01\,[+0.39,+1.64]$ \\
     &  & POPE-GQA & $+4.66\,[+3.79,+5.51]$ & $+5.51\,[+4.49,+6.51]$ & $+0.86\,[+0.13,+1.57]$ \\
     &  & AMBER & $+6.71\,[+6.24,+7.18]$ & $+8.95\,[+8.36,+9.52]$ & $+2.24\,[+1.82,+2.65]$ \\
     & \multirow{3}{*}{ACG} & POPE-A-OKVQA & $+3.23\,[+2.28,+4.21]$ & $+5.11\,[+4.19,+6.05]$ & $+1.88\,[+1.01,+2.78]$ \\
     &  & POPE-GQA & $+3.98\,[+2.89,+5.05]$ & $+5.82\,[+4.81,+6.80]$ & $+1.84\,[+0.91,+2.72]$ \\
     &  & AMBER & $+7.04\,[+6.34,+7.76]$ & $+8.12\,[+7.52,+8.73]$ & $+1.08\,[+0.41,+1.72]$ \\
    \bottomrule
  \end{tabular*}
\end{table*}

\begin{table*}[t]
  \centering\small
  \setlength{\tabcolsep}{3pt}
  \caption{Additional benchmarks for the comparison in \cref{tab:host_swap}. Values are accuracy (\%); $\Delta$ compares each key-edited variant with the corresponding original method. $\ast$/$\dagger$ indicate an image-bootstrap 95\% interval entirely above/below zero. These retain the exploratory settings in \cref{tab:host-settings}; MME uses per-question accuracy.}
  \label{tab:host-swap-additional}
  \begin{tabular*}{\textwidth}{@{\extracolsep{\fill}}@{}llrrrrrrr@{}}
    \toprule
     & & & \multicolumn{3}{c}{GIFT} & \multicolumn{3}{c}{ACG} \\
    \cmidrule(lr){4-6}\cmidrule(lr){7-9}
    Model & Benchmark & Base & Original & Key edit & $\Delta$ & Original & Key branch & $\Delta$ \\
    \midrule
    LLaVA-1.5-7B & POPE-COCO & 84.87 & 85.70 & 85.99 & $+0.29$ & 86.19 & 85.74 & $-0.45$ \\
    LLaVA-1.5-7B & MME & 77.87 & 77.05 & 76.59 & $-0.46$ & 76.39 & 73.36 & $-3.03^{\dagger}$ \\
    \midrule
    LLaVA-1.5-13B & POPE-COCO & 84.12 & 84.33 & 85.28 & $+0.95^{\ast}$ & 84.42 & 87.12 & $+2.70^{\ast}$ \\
    LLaVA-1.5-13B & MME & 80.24 & 79.26 & 75.67 & $-3.59^{\dagger}$ & 80.34 & 79.67 & $-0.67$ \\
    \midrule
    BakLLaVA-7B & POPE-COCO & 83.42 & 85.39 & 85.69 & $+0.30$ & 83.85 & 85.70 & $+1.85^{\ast}$ \\
    BakLLaVA-7B & MME & 77.82 & 78.08 & 77.52 & $-0.56$ & 76.08 & 77.93 & $+1.85^{\ast}$ \\
    \bottomrule
  \end{tabular*}
\end{table*}

The models and benchmarks in Table~\ref{tab:host_swap} were chosen based on the exploratory reproductions, and matched tuning was then applied to them. \Cref{tab:host-matched} gives the six paired mean contrasts. Three CIs lie above zero, two include zero, and one lies below zero. At matched guidance, the 13B ACG gain is $0.33$ points, compared with $4.05$ points in the exploratory comparison at different guidance scales.

The exploratory released-host comparisons appear in \cref{tab:host-swap-intervals,tab:host-swap-additional} and include 17 positive key-versus-host intervals. On the three-benchmark mean, GIFT-Key exceeds released GIFT by $0.86$, $3.07$, and $1.37$ points; tuning GIFT changes those contrasts to $-1.22$, $+0.75$, and $+0.10$.

Results on POPE-COCO and MME are mixed. With matched $\gamma=4$, 13B ACG-Key raises POPE-COCO accuracy from $84.42\%$ to $85.62\%$ and lowers MME from $80.34\%$ to $72.79\%$, giving a five-set mean contrast of $-1.07$ points (95\% CI $[-1.52,-0.64]$). BakLLaVA's matched five-set ACG contrast is $+1.26$ points ($[+0.90,+1.62]$). For GIFT, the matched five-set contrasts are $-0.97$, $+0.14$, and $-0.05$ points for 7B, 13B, and BakLLaVA.

The validation-selected 13B branch strength depends on the guidance scale, changing from $x=-1$ at $\gamma=2.4$ to $x=+2$ at $\gamma=4$.

\begin{table*}[t]
\centering\small
\setlength{\tabcolsep}{3pt}
\renewcommand{\arraystretch}{1.08}
\caption{Matched host comparisons on the selected three test benchmarks: accuracy (\%) and Key-minus-host change (points). GIFT is validation-tuned; ACG uses the same $\gamma$ on both sides. CIs are pointwise 95\% intervals.}
\label{tab:host-matched}
\begin{tabular*}{\textwidth}{@{\extracolsep{\fill}}@{}llrrrr@{}}
\toprule
Model & Host & Tuned host & Key & Change & 95\% CI \\
\midrule
LLaVA-1.5-7B & GIFT & 81.32 & 80.09 & $-1.22$ & $[-1.49,-0.95]$ \\
LLaVA-1.5-13B & GIFT & 78.30 & 79.05 & $+0.75$ & $[+0.45,+1.05]$ \\
BakLLaVA-7B & GIFT & 79.88 & 79.98 & $+0.10$ & $[-0.16,+0.36]$ \\
LLaVA-1.5-7B & ACG & 79.00 & 79.93 & $+0.93$ & $[+0.61,+1.24]$ \\
LLaVA-1.5-13B & ACG & 78.87 & 79.20 & $+0.33$ & $[-0.03,+0.67]$ \\
BakLLaVA-7B & ACG & 78.03 & 78.87 & $+0.83$ & $[+0.50,+1.17]$ \\
\bottomrule
\end{tabular*}
\end{table*}

We also ran both integrations on LLaVA-NeXT-7B/13B, with single-view inputs, and on Qwen2-VL-7B. The released methods themselves give little or no gain on LLaVA-NeXT-7B and Qwen2-VL-7B, and the key variants do not consistently improve on them across these models.

Exploratory integrations with several other attention-intervention methods did not consistently improve on those methods. Our claims concern the GIFT and ACG integrations and settings reported in \cref{tab:host_swap}.

\section{Absolute Attention Allocation}
\label{sec:supp-host-attention}

\begin{figure}[t]
 \centering
 \includegraphics[width=\linewidth]{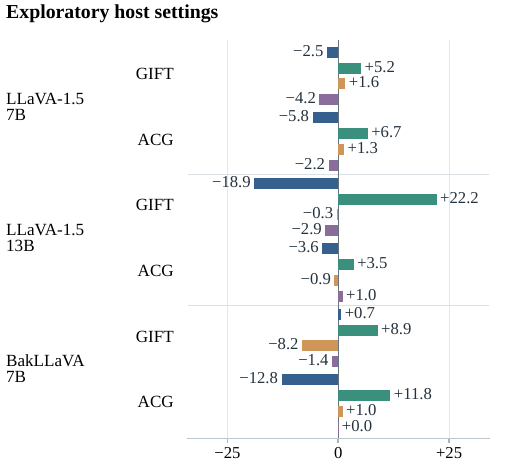}
 \caption{\textbf{Exploratory host attention changes.} Bars show attention-share changes (pp), key edit minus original at answer onset. Within each group: sink, image, question/options, other template; generated share is zero. Colors follow Figure~\ref{fig:attention-drift}. Traces use released GIFT settings and the exploratory ACG guidance, not the matched settings of Table~\ref{tab:host_swap}.}
 \label{fig:exploratory-host-attention}
\end{figure}

\Cref{fig:exploratory-host-attention} uses the exploratory host settings in \cref{tab:host-settings}, on POPE-A-OKVQA (7,452 questions), POPE-GQA (7,128), and AMBER (14,216). These configurations differ from the matched-tuning settings in \cref{tab:host_swap}. They characterize attention allocation under the exploratory settings. \Cref{tab:host-attention-absolute} reports the full absolute allocation for Base, Original, and Key edit, both per benchmark and as an equally weighted benchmark mean. The plotted changes are Key edit minus Original, computed from unrounded shares before rounding the labels.

\paragraph{Measurement position and scope.}
We measure the attention row that predicts the first answer token, averaging all query heads and the specified layers, then questions within each benchmark. GIFT measurements cover its enhancement layers: indices 12--22 for LLaVA-1.5-7B and BakLLaVA-7B, and 14--20 for LLaVA-1.5-13B (zero-based, inclusive). ACG measurements cover all 32 layers for the 7B models and all 40 for the 13B model. For ACG, the measured distribution is the normalized attention in the conditional main branch before contrastive guidance, not a signed combination of branch weights.

\paragraph{Separate Base averages.}
The same unedited forward pass provides Base for both methods, but the layer averages differ: GIFT uses its enhancement layers and ACG uses all layers. Each Original and Key edit condition is compared with Base under its own method's scope. Combining these Base averages would change the comparison. The replacement conditions retain the host methods' selection and guidance structures as described in Appendix~\ref{sec:supp-host-methods}. Original and key edit ACG both use $\gamma=2.4$ for LLaVA-1.5-7B; for LLaVA-1.5-13B and BakLLaVA, Original uses $\gamma=4$ and Key edit uses $\gamma=2.4$. Their differences therefore include the guidance-strength change as well as branch replacement.

\paragraph{Token groups and conservation.}
The sink is the first template token. Image and question/option positions are disjoint from it; the question group combines content words, function words, and any options. Remaining template and generated positions complete the visible keys, with no generated positions at answer onset. The host figure omits generated positions, which have zero mass at answer onset; the tables below retain their absolute shares with template/generated combined as Other. The partition sums to 100\% before rounding and describes attention allocation, not a causal accuracy mechanism.

\begin{table*}[p]
\centering
\small
\setlength{\tabcolsep}{5pt}
\begin{tabular}{llrrrrrrrr}
\toprule
& & \multicolumn{4}{c}{GIFT scope} & \multicolumn{4}{c}{ACG scope} \\
\cmidrule(lr){3-6}\cmidrule(lr){7-10}
Benchmark & Condition & Sink & Image & Q./opt. & Other & Sink & Image & Q./opt. & Other \\
\midrule
\multicolumn{10}{l}{\textbf{LLaVA-1.5-7B}} \\
\midrule
POPE-A-OKVQA & Base & 35.78 & 6.52 & 12.38 & 45.33 & 37.22 & 8.83 & 7.97 & 45.99 \\
 & Original & 30.14 & 14.20 & 21.48 & 34.18 & 39.62 & 6.54 & 4.64 & 49.20 \\
 & Key edit & 27.29 & 20.44 & 22.70 & 29.58 & 33.62 & 13.36 & 6.30 & 46.72 \\
\midrule
POPE-GQA & Base & 35.76 & 6.48 & 12.44 & 45.32 & 37.21 & 8.78 & 8.01 & 46.01 \\
 & Original & 30.10 & 14.24 & 21.54 & 34.11 & 39.61 & 6.52 & 4.59 & 49.28 \\
 & Key edit & 27.18 & 20.63 & 22.73 & 29.46 & 33.69 & 13.20 & 6.32 & 46.79 \\
\midrule
AMBER & Base & 35.56 & 6.00 & 13.36 & 45.09 & 36.99 & 8.58 & 8.55 & 45.88 \\
 & Original & 28.55 & 15.42 & 23.65 & 32.38 & 39.37 & 6.51 & 5.70 & 48.42 \\
 & Key edit & 26.74 & 18.33 & 26.04 & 28.89 & 33.93 & 12.99 & 6.23 & 46.85 \\
\midrule
Mean & Base & 35.70 & 6.33 & 12.72 & 45.25 & 37.14 & 8.73 & 8.18 & 45.96 \\
 & Original & 29.60 & 14.62 & 22.22 & 33.56 & 39.53 & 6.53 & 4.97 & 48.97 \\
 & Key edit & 27.07 & 19.80 & 23.83 & 29.31 & 33.75 & 13.18 & 6.28 & 46.79 \\
\midrule
\multicolumn{10}{l}{\textbf{LLaVA-1.5-13B}} \\
\midrule
POPE-A-OKVQA & Base & 69.39 & 6.38 & 15.07 & 9.16 & 72.61 & 7.32 & 6.52 & 13.55 \\
 & Original & 54.44 & 15.88 & 26.02 & 3.67 & 75.11 & 5.71 & 5.45 & 13.73 \\
 & Key edit & 35.25 & 37.95 & 25.97 & 0.82 & 71.49 & 9.25 & 4.61 & 14.66 \\
\midrule
POPE-GQA & Base & 69.37 & 6.33 & 15.19 & 9.11 & 72.61 & 7.29 & 6.56 & 13.54 \\
 & Original & 54.43 & 15.78 & 26.11 & 3.68 & 75.14 & 5.72 & 5.43 & 13.71 \\
 & Key edit & 35.23 & 38.14 & 25.80 & 0.83 & 71.50 & 9.20 & 4.66 & 14.64 \\
\midrule
AMBER & Base & 67.96 & 6.18 & 15.93 & 9.93 & 72.38 & 7.12 & 6.84 & 13.66 \\
 & Original & 53.43 & 15.36 & 27.24 & 3.98 & 74.92 & 5.50 & 5.81 & 13.77 \\
 & Key edit & 34.98 & 37.52 & 26.56 & 0.94 & 71.39 & 8.95 & 4.81 & 14.85 \\
\midrule
Mean & Base & 68.91 & 6.30 & 15.40 & 9.40 & 72.53 & 7.24 & 6.64 & 13.59 \\
 & Original & 54.10 & 15.67 & 26.45 & 3.77 & 75.05 & 5.64 & 5.56 & 13.74 \\
 & Key edit & 35.15 & 37.87 & 26.11 & 0.86 & 71.46 & 9.13 & 4.69 & 14.72 \\
\midrule
\multicolumn{10}{l}{\textbf{BakLLaVA-7B}} \\
\midrule
POPE-A-OKVQA & Base & 35.71 & 11.62 & 24.21 & 28.46 & 38.64 & 12.36 & 15.65 & 33.36 \\
 & Original & 24.34 & 25.11 & 35.13 & 15.42 & 46.34 & 7.96 & 11.09 & 34.60 \\
 & Key edit & 24.81 & 34.57 & 26.74 & 13.88 & 33.24 & 20.18 & 12.12 & 34.46 \\
\midrule
POPE-GQA & Base & 35.65 & 11.54 & 24.39 & 28.43 & 38.64 & 12.22 & 15.75 & 33.39 \\
 & Original & 24.30 & 24.95 & 35.38 & 15.36 & 46.34 & 7.85 & 11.14 & 34.67 \\
 & Key edit & 24.75 & 34.36 & 27.03 & 13.86 & 33.26 & 20.03 & 12.17 & 34.55 \\
\midrule
AMBER & Base & 35.78 & 9.70 & 25.11 & 29.40 & 38.92 & 10.70 & 16.52 & 33.87 \\
 & Original & 24.25 & 23.81 & 36.95 & 14.99 & 45.80 & 7.45 & 11.50 & 35.25 \\
 & Key edit & 25.37 & 31.57 & 29.20 & 13.86 & 33.59 & 18.35 & 12.42 & 35.64 \\
\midrule
Mean & Base & 35.71 & 10.96 & 24.57 & 28.76 & 38.73 & 11.76 & 15.97 & 33.54 \\
 & Original & 24.30 & 24.62 & 35.82 & 15.26 & 46.16 & 7.75 & 11.25 & 34.84 \\
 & Key edit & 24.98 & 33.50 & 27.66 & 13.86 & 33.36 & 19.52 & 12.24 & 34.88 \\
\bottomrule
\end{tabular}
\caption{Absolute attention shares (\%) underlying \cref{fig:exploratory-host-attention}. Q./opt. denotes question and option tokens. Mean weights the three benchmarks equally; GIFT and ACG retain their own measurement scopes. Rounding can affect displayed totals.}
\label{tab:host-attention-absolute}
\end{table*}

\paragraph{Standalone \methodname measurements.}
\Cref{tab:shrink-attention-absolute} gives the corresponding absolute shares for Figure~\ref{fig:attention-drift}, on the four improved models and nine benchmarks in \cref{tab:main-results}. The two passes follow the same unedited sampled continuation, capped at 256 answer tokens. Within each question, attention is averaged over the selected layer/head/query cells across traced answer steps in eligible layers; question means are then averaged within each benchmark, and benchmark means are weighted equally. Head selection is recomputed independently in each pass, so the selected populations can differ. These are full attention distributions within each selected population. The disjoint group partition is the same as for the host methods, but Other now includes generated tokens as well as remaining template tokens. We verified the raw group sums and checked Other against both the remaining groups and the complement of sink, image, and question/option attention. No renormalization of the sink, image, and question/option groups is applied.

\begin{table*}[p]
\centering
\small
\setlength{\tabcolsep}{7pt}
\begin{tabular}{lrrrrrrrr}
\toprule
& \multicolumn{4}{c}{Base} & \multicolumn{4}{c}{\methodname} \\
\cmidrule(lr){2-5}\cmidrule(lr){6-9}
Benchmark & Sink & Image & Q./opt. & Other & Sink & Image & Q./opt. & Other \\
\midrule
\multicolumn{9}{l}{\textbf{LLaVA-OneVision-7B}} \\
\midrule
MMBench-EN & 20.11 & 23.02 & 24.87 & 31.99 & 4.06 & 33.56 & 24.71 & 37.67 \\
POPE & 20.88 & 25.68 & 21.63 & 31.80 & 3.95 & 34.44 & 20.92 & 40.69 \\
CV-Bench & 19.26 & 18.61 & 30.42 & 31.70 & 8.88 & 22.31 & 32.06 & 36.76 \\
EmbSpatial & 19.02 & 16.92 & 32.13 & 31.93 & 5.72 & 22.62 & 33.56 & 38.10 \\
MMStar & 19.72 & 22.86 & 25.46 & 31.96 & 5.60 & 30.77 & 26.49 & 37.14 \\
VSR & 19.96 & 22.10 & 26.75 & 31.19 & 4.25 & 31.79 & 25.29 & 38.68 \\
WhatsUp & 18.69 & 25.82 & 22.83 & 32.66 & 8.19 & 31.39 & 23.65 & 36.77 \\
COCO-QA & 19.71 & 28.65 & 19.50 & 32.13 & 8.30 & 34.23 & 20.62 & 36.85 \\
VG-QA & 19.15 & 26.25 & 21.96 & 32.64 & 3.51 & 35.66 & 21.10 & 39.73 \\
\cmidrule(lr){1-9}
Mean & 19.61 & 23.32 & 25.06 & 32.00 & 5.83 & 30.75 & 25.38 & 38.04 \\
\midrule
\multicolumn{9}{l}{\textbf{Qwen2.5-VL-3B}} \\
\midrule
MMBench-EN & 17.33 & 19.31 & 27.20 & 36.16 & 8.98 & 21.68 & 29.37 & 39.97 \\
POPE & 19.31 & 29.49 & 16.12 & 35.08 & 7.88 & 33.74 & 17.18 & 41.20 \\
CV-Bench & 16.69 & 19.21 & 28.58 & 35.52 & 6.77 & 21.74 & 30.40 & 41.09 \\
EmbSpatial & 16.29 & 14.61 & 32.78 & 36.32 & 8.52 & 16.07 & 35.08 & 40.33 \\
MMStar & 17.46 & 20.52 & 26.42 & 35.60 & 9.15 & 23.20 & 28.29 & 39.36 \\
VSR & 17.72 & 20.39 & 26.77 & 35.12 & 4.82 & 24.51 & 27.00 & 43.66 \\
WhatsUp & 16.29 & 23.38 & 24.47 & 35.87 & 6.34 & 27.08 & 25.43 & 41.15 \\
COCO-QA & 17.13 & 27.03 & 19.41 & 36.43 & 6.79 & 30.37 & 20.75 & 42.09 \\
VG-QA & 16.49 & 25.53 & 21.00 & 36.98 & 8.71 & 27.57 & 22.65 & 41.07 \\
\cmidrule(lr){1-9}
Mean & 17.19 & 22.16 & 24.75 & 35.90 & 7.55 & 25.11 & 26.24 & 41.10 \\
\midrule
\multicolumn{9}{l}{\textbf{Qwen2.5-VL-7B}} \\
\midrule
MMBench-EN & 22.75 & 20.59 & 25.75 & 30.91 & 14.56 & 24.93 & 26.49 & 34.02 \\
POPE & 24.63 & 32.12 & 14.22 & 29.02 & 18.23 & 35.60 & 14.83 & 31.34 \\
CV-Bench & 21.44 & 19.89 & 27.62 & 31.05 & 15.83 & 22.02 & 28.40 & 33.75 \\
EmbSpatial & 21.01 & 15.36 & 31.26 & 32.37 & 16.60 & 16.60 & 32.35 & 34.45 \\
MMStar & 21.85 & 19.93 & 26.64 & 31.57 & 15.77 & 22.72 & 27.31 & 34.20 \\
VSR & 23.24 & 23.38 & 24.42 & 28.97 & 17.35 & 26.04 & 25.07 & 31.53 \\
WhatsUp & 21.61 & 24.29 & 22.59 & 31.50 & 17.20 & 26.54 & 22.85 & 33.41 \\
COCO-QA & 21.94 & 28.87 & 17.51 & 31.69 & 16.02 & 31.80 & 18.10 & 34.08 \\
VG-QA & 21.39 & 26.61 & 19.42 & 32.58 & 15.63 & 29.29 & 20.17 & 34.90 \\
\cmidrule(lr){1-9}
Mean & 22.21 & 23.45 & 23.27 & 31.07 & 16.36 & 26.17 & 23.95 & 33.52 \\
\midrule
\multicolumn{9}{l}{\textbf{Gemma-3-4B}} \\
\midrule
MMBench-EN & 20.51 & 23.78 & 22.67 & 33.05 & 9.51 & 28.50 & 24.36 & 37.64 \\
POPE & 22.78 & 36.13 & 11.67 & 29.42 & 2.45 & 49.82 & 10.61 & 37.13 \\
CV-Bench & 19.78 & 24.24 & 24.27 & 31.71 & 5.88 & 30.49 & 25.32 & 38.31 \\
EmbSpatial & 18.77 & 20.65 & 27.62 & 32.96 & 8.40 & 24.56 & 29.39 & 37.65 \\
MMStar & 20.53 & 25.60 & 21.28 & 32.58 & 6.10 & 32.32 & 22.57 & 39.01 \\
VSR & 20.66 & 28.16 & 21.95 & 29.23 & 0.72 & 42.60 & 17.36 & 39.31 \\
WhatsUp & 19.93 & 28.37 & 20.43 & 31.26 & 9.33 & 34.02 & 21.26 & 35.40 \\
COCO-QA & 20.82 & 31.65 & 16.03 & 31.51 & 9.65 & 38.08 & 16.60 & 35.68 \\
VG-QA & 20.57 & 30.46 & 17.24 & 31.73 & 9.51 & 36.65 & 18.03 & 35.81 \\
\cmidrule(lr){1-9}
Mean & 20.48 & 27.67 & 20.35 & 31.50 & 6.84 & 35.23 & 20.61 & 37.33 \\
\bottomrule
\end{tabular}
\caption{Absolute attention shares (\%) for standalone \methodname in \cref{fig:attention-drift}. The scope is traced answer steps in eligible layers and independently selected heads in each pass. Q./opt. denotes question and option tokens; Other includes remaining template and generated tokens. Mean weights the nine benchmarks equally. Displayed totals can differ from 100\% through rounding.}
\label{tab:shrink-attention-absolute}
\end{table*}

\section{Direction Controls and Strength Transfer}
\label{sec:supp-directions}

\subsection{Reference Directions}

The scalar projection rule, layer schedule, native head selector, and decoding protocol are fixed; only the removed direction changes. Each direction is a unit vector per layer and KV head in post-positional key space. From each of the nine validation benchmarks, 10 prompts estimate the directions and 10 disjoint prompts evaluate their geometry. No test labels enter direction construction or strength selection.

The mean-query reference averages unit queries from prompt rows after the sink across the associated query heads. The mean-key reference averages all non-sink prompt keys. The top key PC is the first axis of their centered covariance. A template reference uses the first fixed template token after the sink, before task content. Three seeded random unit vectors are drawn independently, without orthogonalization. The native sink direction is unchanged. Each alternative uses the same model-level strength selected for \methodname: $0.7,0.3,0.3,0.2,0.3$ for LLaVA-OneVision, Qwen-3B, Qwen-7B, Gemma, and InternVL3.

The accuracy comparison uses all nine test benchmarks, with every arm and its baseline evaluated in a single shared run at \methodname's model-level strengths, rather than the per-benchmark selection in Table~\ref{tab:main-results}. Tables~\ref{tab:direction-full} and~\ref{tab:direction-contrasts} report all five models and paired comparisons. CIs resample test questions independently within benchmarks, weighting their means equally (10,000 draws, seed 0).

On LLaVA-OneVision and Qwen-3B, every alternative has a negative paired contrast against the sink direction, with its pointwise interval below zero. The mean-query direction has greater mean absolute query alignment ($0.76$ and $0.75$, versus $0.47$ and $0.43$ for the sink), yet produces smaller accuracy gains. The mean-key direction is also distinct geometrically: absolute cosine with the sink ranges from $0.33$ to $0.58$ across models. These results support the sink reference at the tested strengths. On the other three models, where \methodname's gains are smaller, most direction contrasts have intervals that include zero; InternVL3's mean-query contrast lies below zero.

\begin{table*}[t]
\centering\small
\setlength{\tabcolsep}{3pt}
\renewcommand{\arraystretch}{1.08}
\caption{Accuracy changes from unedited decoding (percentage points), equal-weight nine-benchmark means. Brackets: pointwise paired-question 95\% CIs; $\ast$/$\dagger$: intervals above/below zero. All directions use \methodname-selected model strengths.}
\label{tab:direction-full}
\begin{tabular*}{\textwidth}{@{\extracolsep{\fill}}@{}lccccc@{}}
\toprule
Direction & \shortstack{LLaVA-OV\\7B} & \shortstack{Qwen2.5-VL\\3B} & \shortstack{Qwen2.5-VL\\7B} & \shortstack{Gemma-3\\4B} & \shortstack{InternVL3\\8B} \\
\midrule
Sink (\methodname) & \shortstack{$+4.57$\rlap{$^{\ast}$}\\\scriptsize [+3.58,+5.54]} & \shortstack{$+3.55$\rlap{$^{\ast}$}\\\scriptsize [+2.69,+4.42]} & \shortstack{$+0.16$\\\scriptsize [-0.44,+0.76]} & \shortstack{$+0.33$\\\scriptsize [-0.31,+0.97]} & \shortstack{$-0.12$\\\scriptsize [-0.73,+0.48]} \\
Mean query & \shortstack{$+2.49$\rlap{$^{\ast}$}\\\scriptsize [+1.50,+3.45]} & \shortstack{$+2.33$\rlap{$^{\ast}$}\\\scriptsize [+1.46,+3.20]} & \shortstack{$+0.24$\\\scriptsize [-0.37,+0.88]} & \shortstack{$+0.39$\\\scriptsize [-0.24,+1.04]} & \shortstack{$-0.98$\rlap{$^{\dagger}$}\\\scriptsize [-1.59,-0.37]} \\
Mean key & \shortstack{$-0.39$\\\scriptsize [-1.35,+0.56]} & \shortstack{$+2.26$\rlap{$^{\ast}$}\\\scriptsize [+1.43,+3.11]} & \shortstack{$+0.47$\\\scriptsize [-0.13,+1.09]} & \shortstack{$+0.43$\\\scriptsize [-0.15,+1.01]} & \shortstack{$-0.50$\\\scriptsize [-1.09,+0.10]} \\
Top key PC & \shortstack{$-11.08$\rlap{$^{\dagger}$}\\\scriptsize [-12.05,-10.11]} & \shortstack{$-5.03$\rlap{$^{\dagger}$}\\\scriptsize [-5.89,-4.14]} & \shortstack{$-0.20$\\\scriptsize [-0.78,+0.40]} & \shortstack{$+0.01$\\\scriptsize [-0.54,+0.57]} & \shortstack{$+0.39$\\\scriptsize [-0.15,+0.96]} \\
Template token & \shortstack{$+0.14$\\\scriptsize [-0.79,+1.07]} & \shortstack{$+0.15$\\\scriptsize [-0.63,+0.94]} & \shortstack{$+0.61$\rlap{$^{\ast}$}\\\scriptsize [+0.06,+1.18]} & \shortstack{$+0.19$\\\scriptsize [-0.44,+0.81]} & \shortstack{$-0.09$\\\scriptsize [-0.64,+0.48]} \\
Random 1 & \shortstack{$+0.40$\\\scriptsize [-0.30,+1.10]} & \shortstack{$+0.41$\\\scriptsize [-0.28,+1.10]} & \shortstack{$-0.17$\\\scriptsize [-0.65,+0.30]} & \shortstack{$-0.09$\\\scriptsize [-0.51,+0.33]} & \shortstack{$-0.40$\\\scriptsize [-0.87,+0.06]} \\
Random 2 & \shortstack{$+0.18$\\\scriptsize [-0.52,+0.90]} & \shortstack{$-0.11$\\\scriptsize [-0.77,+0.56]} & \shortstack{$+0.07$\\\scriptsize [-0.41,+0.55]} & \shortstack{$-0.00$\\\scriptsize [-0.42,+0.41]} & \shortstack{$-0.03$\\\scriptsize [-0.51,+0.45]} \\
Random 3 & \shortstack{$-0.24$\\\scriptsize [-0.95,+0.48]} & \shortstack{$-0.76$\rlap{$^{\dagger}$}\\\scriptsize [-1.45,-0.07]} & \shortstack{$-0.17$\\\scriptsize [-0.65,+0.31]} & \shortstack{$-0.05$\\\scriptsize [-0.48,+0.37]} & \shortstack{$-0.23$\\\scriptsize [-0.70,+0.25]} \\
\bottomrule
\end{tabular*}
\end{table*}

\begin{table*}[t]
\centering\small
\setlength{\tabcolsep}{3pt}
\renewcommand{\arraystretch}{1.08}
\caption{Paired accuracy contrasts against \methodname (percentage points), equal-weight nine-benchmark means. Brackets: pointwise paired-question 95\% CIs; $\ast$/$\dagger$: intervals above/below zero. All directions use \methodname-selected model strengths.}
\label{tab:direction-contrasts}
\begin{tabular*}{\textwidth}{@{\extracolsep{\fill}}@{}lccccc@{}}
\toprule
Direction & \shortstack{LLaVA-OV\\7B} & \shortstack{Qwen2.5-VL\\3B} & \shortstack{Qwen2.5-VL\\7B} & \shortstack{Gemma-3\\4B} & \shortstack{InternVL3\\8B} \\
\midrule
Mean query & \shortstack{$-2.08$\rlap{$^{\dagger}$}\\\scriptsize [-3.02,-1.13]} & \shortstack{$-1.22$\rlap{$^{\dagger}$}\\\scriptsize [-2.01,-0.42]} & \shortstack{$+0.08$\\\scriptsize [-0.53,+0.69]} & \shortstack{$+0.06$\\\scriptsize [-0.52,+0.64]} & \shortstack{$-0.86$\rlap{$^{\dagger}$}\\\scriptsize [-1.44,-0.27]} \\
Mean key & \shortstack{$-4.97$\rlap{$^{\dagger}$}\\\scriptsize [-5.93,-4.02]} & \shortstack{$-1.29$\rlap{$^{\dagger}$}\\\scriptsize [-2.11,-0.45]} & \shortstack{$+0.31$\\\scriptsize [-0.28,+0.93]} & \shortstack{$+0.10$\\\scriptsize [-0.51,+0.70]} & \shortstack{$-0.38$\\\scriptsize [-0.98,+0.22]} \\
Top key PC & \shortstack{$-15.66$\rlap{$^{\dagger}$}\\\scriptsize [-16.64,-14.68]} & \shortstack{$-8.58$\rlap{$^{\dagger}$}\\\scriptsize [-9.47,-7.67]} & \shortstack{$-0.36$\\\scriptsize [-0.99,+0.28]} & \shortstack{$-0.33$\\\scriptsize [-0.98,+0.32]} & \shortstack{$+0.51$\\\scriptsize [-0.10,+1.13]} \\
Template token & \shortstack{$-4.43$\rlap{$^{\dagger}$}\\\scriptsize [-5.40,-3.47]} & \shortstack{$-3.40$\rlap{$^{\dagger}$}\\\scriptsize [-4.26,-2.53]} & \shortstack{$+0.45$\\\scriptsize [-0.15,+1.06]} & \shortstack{$-0.14$\\\scriptsize [-0.70,+0.39]} & \shortstack{$+0.03$\\\scriptsize [-0.57,+0.64]} \\
Random 1 & \shortstack{$-4.17$\rlap{$^{\dagger}$}\\\scriptsize [-5.15,-3.17]} & \shortstack{$-3.14$\rlap{$^{\dagger}$}\\\scriptsize [-4.00,-2.29]} & \shortstack{$-0.33$\\\scriptsize [-0.93,+0.25]} & \shortstack{$-0.42$\\\scriptsize [-1.06,+0.23]} & \shortstack{$-0.28$\\\scriptsize [-0.90,+0.33]} \\
Random 2 & \shortstack{$-4.40$\rlap{$^{\dagger}$}\\\scriptsize [-5.37,-3.43]} & \shortstack{$-3.66$\rlap{$^{\dagger}$}\\\scriptsize [-4.52,-2.80]} & \shortstack{$-0.09$\\\scriptsize [-0.69,+0.50]} & \shortstack{$-0.34$\\\scriptsize [-0.98,+0.31]} & \shortstack{$+0.09$\\\scriptsize [-0.53,+0.69]} \\
Random 3 & \shortstack{$-4.81$\rlap{$^{\dagger}$}\\\scriptsize [-5.79,-3.86]} & \shortstack{$-4.31$\rlap{$^{\dagger}$}\\\scriptsize [-5.17,-3.44]} & \shortstack{$-0.33$\\\scriptsize [-0.93,+0.26]} & \shortstack{$-0.39$\\\scriptsize [-1.03,+0.27]} & \shortstack{$-0.10$\\\scriptsize [-0.72,+0.50]} \\
\bottomrule
\end{tabular*}
\end{table*}

\subsection{Strength Transfer}

We reuse the main sampled evaluation, with no additional generation. Benchmark-specific selection maximizes each validation mean; per-model selection maximizes the equal-weight mean over nine validation benchmarks. Global selection maximizes the mean over models using the shared grid $\{0.3,0.5\}$ and chooses $0.3$. Ties favor smaller strengths throughout. Test questions are excluded from every choice.

\Cref{tab:fixed-strength-full} retains all three selection rules and their paired-question CIs. Per-model selection changes the nine-benchmark gain by at most $0.33$ points from benchmark-specific selection. Global selection preserves positive intervals for the four main models; LLaVA-OneVision gains a further $1.26$ points with its per-model setting.

\begin{table*}[t]
\centering\small
\setlength{\tabcolsep}{3pt}
\renewcommand{\arraystretch}{1.08}
\caption{Validation-selected strength transfer: nine-benchmark mean accuracy changes (points). Brackets are pointwise paired-question 95\% CIs. The per-model strengths are $0.7,0.3,0.3,0.2,0.3$ in row order; the global strength is $0.3$. $\ast$: CI above zero.}
\label{tab:fixed-strength-full}
\begin{tabular*}{\textwidth}{@{\extracolsep{\fill}}@{}lccc@{}}
\toprule
Model & Per benchmark & Per model & Global \\
\midrule
LLaVA-OV-7B & $+5.10$\rlap{$^{\ast}$}\; $[+4.55,+5.64]$ & $+5.14$\rlap{$^{\ast}$}\; $[+4.56,+5.71]$ & $+3.88$\rlap{$^{\ast}$}\; $[+3.35,+4.41]$ \\
Qwen2.5-VL-3B & $+2.72$\rlap{$^{\ast}$}\; $[+2.21,+3.23]$ & $+3.05$\rlap{$^{\ast}$}\; $[+2.54,+3.57]$ & $+3.05$\rlap{$^{\ast}$}\; $[+2.54,+3.57]$ \\
Qwen2.5-VL-7B & $+0.53$\rlap{$^{\ast}$}\; $[+0.19,+0.87]$ & $+0.48$\rlap{$^{\ast}$}\; $[+0.12,+0.83]$ & $+0.48$\rlap{$^{\ast}$}\; $[+0.12,+0.83]$ \\
Gemma-3-4B & $+0.84$\rlap{$^{\ast}$}\; $[+0.42,+1.27]$ & $+0.72$\rlap{$^{\ast}$}\; $[+0.31,+1.12]$ & $+0.71$\rlap{$^{\ast}$}\; $[+0.28,+1.14]$ \\
InternVL3-8B & $-0.03$\; $[-0.38,+0.33]$ & $-0.07$\; $[-0.43,+0.28]$ & $-0.07$\; $[-0.43,+0.28]$ \\
\bottomrule
\end{tabular*}
\end{table*}

\FloatBarrier

\section{Multiplicity-Adjusted Accuracy Comparisons}
\label{sec:supp-multiplicity}

We distinguish two test families defined before inspecting their adjusted $p$-values: the five model-level nine-benchmark means, and the 45 model-benchmark cells. Each uses the original validation-selected strengths and three-seed per-question mean changes. Two-sided bootstrap $p$-values are twice the smaller tail probability at zero, floored at $1/10{,}000$. Model means resample questions separately within each benchmark before equal weighting. These corrections address the stated accuracy families.

Holm correction retains all four positive model-level means. Among the 45 cells, 26 have raw $p<0.05$; 18 survive Holm correction (16 gains, two losses), and 23 survive Benjamini--Hochberg correction at $q<0.05$. Both Holm-retained losses occur on InternVL3. Tables~\ref{tab:multiplicity-primary} and~\ref{tab:multiplicity-secondary} report the corrected comparisons. The stars in Table~\ref{tab:main-results} continue to denote pointwise intervals, not these adjusted tests.

\begin{table*}[t]
\centering\small
\setlength{\tabcolsep}{3pt}
\renewcommand{\arraystretch}{1.08}
\caption{Model-level nine-benchmark accuracy changes and two-sided bootstrap $p$-values, with Holm correction over five models. Bootstrap resolution is $10^{-4}$.}
\label{tab:multiplicity-primary}
\begin{tabular*}{\textwidth}{@{\extracolsep{\fill}}@{}lrrr@{}}
\toprule
Model & Change (points) & Raw $p$ & Holm $p$ \\
\midrule
LLaVA-OV-7B & +5.10 & 0.0001 & 0.0005 \\
Qwen2.5-VL-3B & +2.72 & 0.0001 & 0.0005 \\
Qwen2.5-VL-7B & +0.53 & 0.0020 & 0.0040 \\
Gemma-3-4B & +0.84 & 0.0001 & 0.0005 \\
InternVL3-8B & -0.03 & 0.8556 & 0.8556 \\
\bottomrule
\end{tabular*}
\end{table*}

\begin{table*}[t]
\centering\small
\setlength{\tabcolsep}{3pt}
\renewcommand{\arraystretch}{1.08}
\caption{Cells with raw $p<0.05$ from the family of 45 model-benchmark accuracy contrasts. Holm and Benjamini--Hochberg (BH) corrections use all 45 cells, including those not displayed. Changes are percentage points.}
\label{tab:multiplicity-secondary}
\begin{tabular*}{\textwidth}{@{\extracolsep{\fill}}@{}llrrrr@{}}
\toprule
Model & Benchmark & Change & Raw $p$ & Holm $p$ & BH $q$ \\
\midrule
LLaVA-OV-7B & MMBench & +6.73 & 0.0001 & 0.0045 & 0.0003 \\
LLaVA-OV-7B & POPE & +1.94 & 0.0001 & 0.0045 & 0.0003 \\
LLaVA-OV-7B & CV-Bench & +3.63 & 0.0001 & 0.0045 & 0.0003 \\
LLaVA-OV-7B & EmbSpatial & +1.95 & 0.0022 & 0.0594 & 0.0052 \\
LLaVA-OV-7B & MMStar & +5.24 & 0.0001 & 0.0045 & 0.0003 \\
LLaVA-OV-7B & VSR & +10.83 & 0.0001 & 0.0045 & 0.0003 \\
LLaVA-OV-7B & WhatsUp & +2.64 & 0.0302 & 0.6342 & 0.0544 \\
LLaVA-OV-7B & COCO-QA & +4.64 & 0.0001 & 0.0045 & 0.0003 \\
LLaVA-OV-7B & VG-QA & +8.30 & 0.0001 & 0.0045 & 0.0003 \\
Qwen2.5-VL-3B & POPE & +1.01 & 0.0002 & 0.0064 & 0.0006 \\
Qwen2.5-VL-3B & CV-Bench & +2.80 & 0.0001 & 0.0045 & 0.0003 \\
Qwen2.5-VL-3B & EmbSpatial & +2.07 & 0.0006 & 0.0180 & 0.0015 \\
Qwen2.5-VL-3B & VSR & +5.32 & 0.0001 & 0.0045 & 0.0003 \\
Qwen2.5-VL-3B & WhatsUp & +7.36 & 0.0001 & 0.0045 & 0.0003 \\
Qwen2.5-VL-3B & VG-QA & +3.47 & 0.0001 & 0.0045 & 0.0003 \\
Qwen2.5-VL-7B & VSR & +1.17 & 0.0166 & 0.4150 & 0.0356 \\
Qwen2.5-VL-7B & VG-QA & +2.41 & 0.0006 & 0.0180 & 0.0015 \\
Gemma-3-4B & CV-Bench & +1.45 & 0.0268 & 0.5896 & 0.0502 \\
Gemma-3-4B & EmbSpatial & +1.18 & 0.0198 & 0.4554 & 0.0387 \\
Gemma-3-4B & WhatsUp & +5.94 & 0.0001 & 0.0045 & 0.0003 \\
Gemma-3-4B & COCO-QA & +0.93 & 0.0404 & 0.8080 & 0.0699 \\
Gemma-3-4B & VG-QA & -1.49 & 0.0174 & 0.4176 & 0.0356 \\
InternVL3-8B & POPE & -0.66 & 0.0006 & 0.0180 & 0.0015 \\
InternVL3-8B & MMStar & +2.56 & 0.0004 & 0.0124 & 0.0012 \\
InternVL3-8B & WhatsUp & -3.45 & 0.0001 & 0.0045 & 0.0003 \\
InternVL3-8B & VG-QA & +2.01 & 0.0054 & 0.1404 & 0.0122 \\
\bottomrule
\end{tabular*}
\end{table*}

\section{Direct-Answer Evaluation}
\label{sec:supp-direct-answer}

We also repeat the nine-benchmark comparison with greedy decoding and the instruction ``Answer with the option's letter from the given choices directly.'' Scoring retains the strict option decision rule. Strengths are selected on validation separately for each model, benchmark, and image setting, using the original grids, heads, and layers. Each model/setting is evaluated in one shared run.

The native setting restores processor defaults: AnyRes for LLaVA-OneVision, dynamic resolution for Qwen2.5-VL, and tiles for InternVL3. Gemma already uses its native setting. The fixed setting retains the main experiment's preprocessing. Qwen native-resolution runs cover fewer test questions than their fixed counterparts: test counts in benchmark order are 2,425, 5,040, 1,478, 2,040, 851, 854, 448, 1,510, and 818. Other runs retain the main test counts. Each contrast pairs the arms on their common questions; native-versus-fixed differences are not controlled estimates of resolution alone.

\Cref{tab:direct-answer-summary} gives the means; Tables~\ref{tab:direct-answer-native} and~\ref{tab:direct-answer-protocol} retain every benchmark. Mean changes at validation-selected strengths range from $-0.23$ to $+0.10$ points, with some small negative intervals excluding zero. Under this protocol, unedited accuracy is higher than under sampled reasoning for the two most responsive models, and unparsed responses average at most $0.3\%$, so accuracy changes are small.

\paragraph{Interpreting the sampled gains.}
The existing paired decomposition separates pairs with at least one unparsed response from pairs where both responses parse, using all questions as denominator. These parts sum to the accuracy change before rounding. Tables~\ref{tab:test-unparsed} and~\ref{tab:format-answer} report the main results. In the direction study, the sink's LLaVA-OneVision gain is $+4.57$ points: $+5.55$ from pairs involving unparsed responses and $-0.98$ from pairs where both parse; Qwen-3B has $+3.57$ and $-0.03$ contributions to its $+3.55$-point gain (rounded separately). Much of the sampled-reasoning gain therefore comes from responses that reach a clear final answer under \methodname.

\begin{table*}[t]
\centering\small
\setlength{\tabcolsep}{3pt}
\renewcommand{\arraystretch}{1.08}
\caption{Direct-answer greedy evaluation: nine-benchmark mean accuracy (\%), with per-benchmark validation-selected strengths. Fixed denotes the main experiment's image preprocessing. Gemma is already native. Changes and pointwise CIs are in percentage points.}
\label{tab:direct-answer-summary}
\begin{tabular*}{\textwidth}{@{\extracolsep{\fill}}@{}llrrrr@{}}
\toprule
Model & Resolution & Base & \methodname & Change & 95\% CI \\
\midrule
LLaVA-OV-7B & Native & 76.0 & 76.0 & $+0.02$ & $[-0.14,+0.18]$ \\
LLaVA-OV-7B & Fixed & 75.6 & 75.4 & $-0.23$\rlap{$^{\dagger}$} & $[-0.40,-0.06]$ \\
Qwen2.5-VL-3B & Native & 74.6 & 74.6 & $-0.05$ & $[-0.28,+0.18]$ \\
Qwen2.5-VL-3B & Fixed & 74.2 & 73.9 & $-0.22$\rlap{$^{\dagger}$} & $[-0.43,-0.02]$ \\
Qwen2.5-VL-7B & Native & 79.3 & 79.1 & $-0.17$\rlap{$^{\dagger}$} & $[-0.32,-0.02]$ \\
Qwen2.5-VL-7B & Fixed & 79.1 & 79.1 & $+0.04$ & $[-0.11,+0.18]$ \\
Gemma-3-4B & Native & 57.9 & 57.7 & $-0.22$ & $[-0.59,+0.15]$ \\
InternVL3-8B & Native & 82.3 & 82.3 & $+0.04$ & $[-0.13,+0.22]$ \\
InternVL3-8B & Fixed & 82.2 & 82.3 & $+0.10$ & $[-0.07,+0.27]$ \\
\bottomrule
\end{tabular*}
\end{table*}

\begin{table*}[t]
\centering\small
\setlength{\tabcolsep}{3pt}
\renewcommand{\arraystretch}{1.08}
\caption{Direct-answer accuracy with native image preprocessing. Each cell gives Base accuracy (\%) / \methodname change (points); $\ast$/$\dagger$ denote pointwise 95\% CIs above/below zero. Strengths are selected on validation, separately for each benchmark and resolution.}
\label{tab:direct-answer-native}
\begin{tabular*}{\textwidth}{@{\extracolsep{\fill}}@{}lccccc@{}}
\toprule
Benchmark & \shortstack{LLaVA-OV\\7B} & \shortstack{Qwen2.5-VL\\3B} & \shortstack{Qwen2.5-VL\\7B} & \shortstack{Gemma-3\\4B} & \shortstack{InternVL3\\8B} \\
\midrule
MMBench & 86.6 / $+0.23$ & 85.0 / $-0.37$ & 89.2 / $-0.08$ & 80.0 / $-0.52$\rlap{$^{\dagger}$} & 89.8 / $-0.17$ \\
POPE & 89.7 / $+0.08$ & 90.2 / $-0.22$ & 89.0 / $-0.06$ & 84.6 / $-0.46$\rlap{$^{\dagger}$} & 90.8 / $-0.06$ \\
CV-Bench & 78.7 / $-0.19$ & 74.1 / $+0.14$ & 80.4 / $+0.20$ & 55.7 / $-1.18$\rlap{$^{\dagger}$} & 83.2 / $+0.09$ \\
EmbSpatial & 72.4 / $-0.52$\rlap{$^{\dagger}$} & 63.9 / $-0.29$ & 70.7 / $+0.05$ & 49.2 / $+0.96$ & 75.4 / $+0.14$ \\
MMStar & 62.5 / $-0.33$ & 55.2 / $-0.24$ & 61.7 / $-0.82$ & 44.8 / $+0.08$ & 67.8 / $-0.33$ \\
VSR & 76.5 / $-0.25$ & 80.0 / $+0.47$ & 88.1 / $+0.00$ & 67.4 / $-0.08$ & 86.0 / $-0.16$ \\
WhatsUp & 94.8 / $+0.76$\rlap{$^{\ast}$} & 92.0 / $-0.22$ & 95.5 / $-0.45$ & 50.2 / $-0.91$ & 94.1 / $-0.15$ \\
COCO-QA & 60.6 / $-0.23$ & 67.5 / $-0.33$ & 69.7 / $+0.00$ & 48.0 / $+2.00$\rlap{$^{\ast}$} & 72.9 / $+0.00$ \\
VG-QA & 61.8 / $+0.60$\rlap{$^{\ast}$} & 63.8 / $+0.61$ & 69.2 / $-0.37$ & 41.3 / $-1.89$\rlap{$^{\dagger}$} & 80.4 / $+1.03$\rlap{$^{\ast}$} \\
Mean & 76.0 / $+0.02$ & 74.6 / $-0.05$ & 79.3 / $-0.17$\rlap{$^{\dagger}$} & 57.9 / $-0.22$ & 82.3 / $+0.04$ \\
\bottomrule
\end{tabular*}
\end{table*}

\begin{table*}[t]
\centering\small
\setlength{\tabcolsep}{3pt}
\renewcommand{\arraystretch}{1.08}
\caption{Direct-answer accuracy with fixed main-protocol image preprocessing. Each cell gives Base accuracy (\%) / \methodname change (points); $\ast$/$\dagger$ denote pointwise 95\% CIs above/below zero. Strengths are selected on validation, separately for each benchmark and resolution.}
\label{tab:direct-answer-protocol}
\begin{tabular*}{\textwidth}{@{\extracolsep{\fill}}@{}lcccc@{}}
\toprule
Benchmark & \shortstack{LLaVA-OV\\7B} & \shortstack{Qwen2.5-VL\\3B} & \shortstack{Qwen2.5-VL\\7B} & \shortstack{InternVL3\\8B} \\
\midrule
MMBench & 86.4 / $+0.03$ & 85.3 / $-0.40$\rlap{$^{\dagger}$} & 89.3 / $-0.26$\rlap{$^{\dagger}$} & 89.6 / $-0.32$\rlap{$^{\dagger}$} \\
POPE & 88.1 / $+0.06$ & 89.4 / $+0.00$ & 88.9 / $-0.10$ & 90.1 / $-0.19$\rlap{$^{\dagger}$} \\
CV-Bench & 77.4 / $-0.05$ & 72.7 / $+0.00$ & 79.5 / $+0.00$ & 82.4 / $+0.09$ \\
EmbSpatial & 72.7 / $-0.52$\rlap{$^{\dagger}$} & 62.8 / $-0.69$\rlap{$^{\dagger}$} & 73.3 / $-0.62$\rlap{$^{\dagger}$} & 75.5 / $-0.07$ \\
MMStar & 60.0 / $-0.49$ & 53.7 / $-0.74$ & 59.8 / $+0.49$ & 67.3 / $+0.16$ \\
VSR & 76.6 / $-0.49$ & 81.2 / $-0.25$ & 87.8 / $+0.08$ & 86.7 / $+0.08$ \\
WhatsUp & 96.2 / $+0.00$ & 90.9 / $-0.30$ & 95.7 / $+0.15$ & 95.3 / $+0.61$\rlap{$^{\ast}$} \\
COCO-QA & 62.1 / $-0.42$\rlap{$^{\dagger}$} & 66.9 / $-0.05$ & 68.8 / $+0.42$\rlap{$^{\ast}$} & 73.4 / $+0.09$ \\
VG-QA & 61.0 / $-0.17$ & 64.6 / $+0.43$ & 68.6 / $+0.17$ & 79.8 / $+0.43$ \\
Mean & 75.6 / $-0.23$\rlap{$^{\dagger}$} & 74.2 / $-0.22$\rlap{$^{\dagger}$} & 79.1 / $+0.04$ & 82.2 / $+0.10$ \\
\bottomrule
\end{tabular*}
\end{table*}

\begin{table*}[t]
  \centering\small
  \setlength{\tabcolsep}{5pt}
  \caption{Unparsed test responses (\%), unedited $\to$ \methodname, at the validation-selected strengths in \cref{tab:main-results}.}
  \label{tab:test-unparsed}
  \begin{tabular}{lccccc}
    \toprule
    Benchmark & \shortstack{LLaVA-OV\\7B} & \shortstack{Qwen2.5-VL\\3B} & \shortstack{Qwen2.5-VL\\7B} & \shortstack{Gemma-3\\4B} & \shortstack{InternVL3\\8B} \\
    \midrule
    MMBench & $23.5\!\to\!13.5$ & $10.3\!\to\!8.5$ & $0.5\!\to\!0.4$ & $0.0\!\to\!0.0$ & $0.2\!\to\!0.2$ \\
    POPE & $29.2\!\to\!26.8$ & $5.7\!\to\!3.9$ & $0.3\!\to\!0.2$ & $0.0\!\to\!0.0$ & $0.1\!\to\!0.6$ \\
    CV-Bench & $25.7\!\to\!20.2$ & $10.7\!\to\!5.1$ & $0.9\!\to\!0.6$ & $0.0\!\to\!0.1$ & $0.2\!\to\!0.2$ \\
    EmbSpatial & $18.5\!\to\!13.0$ & $12.3\!\to\!9.1$ & $1.4\!\to\!1.1$ & $0.1\!\to\!0.0$ & $0.2\!\to\!0.1$ \\
    MMStar & $27.3\!\to\!17.8$ & $14.9\!\to\!12.5$ & $2.0\!\to\!1.2$ & $2.4\!\to\!2.1$ & $1.9\!\to\!1.5$ \\
    \midrule
    VSR & $44.1\!\to\!27.9$ & $11.9\!\to\!2.5$ & $0.3\!\to\!0.1$ & $0.0\!\to\!0.0$ & $0.0\!\to\!0.0$ \\
    WhatsUp & $18.4\!\to\!14.8$ & $14.5\!\to\!4.1$ & $0.7\!\to\!0.4$ & $0.1\!\to\!0.2$ & $0.4\!\to\!0.1$ \\
    COCO-QA & $34.9\!\to\!27.5$ & $9.3\!\to\!5.8$ & $0.4\!\to\!0.5$ & $0.0\!\to\!0.1$ & $0.3\!\to\!0.2$ \\
    VG-QA & $33.9\!\to\!24.9$ & $13.7\!\to\!9.0$ & $0.3\!\to\!0.1$ & $0.1\!\to\!0.1$ & $0.2\!\to\!0.2$ \\
    \bottomrule
  \end{tabular}
\end{table*}

\begin{table*}[t]
  \centering\small
  \setlength{\tabcolsep}{5pt}
  \caption{Paired format / answer contributions (percentage points) to sampled test accuracy. Both use all test questions as denominator and sum to the total change before rounding. Strengths are those in \cref{tab:main-results}.}
  \label{tab:format-answer}
  \begin{tabular}{lccccc}
    \toprule
    Benchmark & \shortstack{LLaVA-OV\\7B} & \shortstack{Qwen2.5-VL\\3B} & \shortstack{Qwen2.5-VL\\7B} & \shortstack{Gemma-3\\4B} & \shortstack{InternVL3\\8B} \\
    \midrule
    MMBench & $+8.27\,/\,-1.54$ & $+1.21\,/\,-0.51$ & $+0.03\,/\,-0.23$ & $+0.00\,/\,-0.20$ & $-0.04\,/\,-0.08$ \\
    POPE & $+1.99\,/\,-0.05$ & $+1.50\,/\,-0.49$ & $+0.03\,/\,+0.15$ & $+0.00\,/\,-0.09$ & $-0.48\,/\,-0.18$ \\
    CV-Bench & $+3.82\,/\,-0.19$ & $+3.18\,/\,-0.38$ & $+0.16\,/\,+0.25$ & $-0.03\,/\,+1.48$ & $-0.02\,/\,-0.82$ \\
    EmbSpatial & $+3.37\,/\,-1.42$ & $+2.14\,/\,-0.07$ & $+0.22\,/\,-0.01$ & $+0.05\,/\,+1.13$ & $+0.00\,/\,-0.53$ \\
    MMStar & $+5.54\,/\,-0.30$ & $+1.04\,/\,-0.30$ & $+0.52\,/\,+0.82$ & $+0.14\,/\,-0.05$ & $+0.22\,/\,+2.35$ \\
    \midrule
    VSR & $+12.22\,/\,-1.39$ & $+6.68\,/\,-1.36$ & $+0.19\,/\,+0.98$ & $+0.00\,/\,-0.22$ & $+0.00\,/\,+0.76$ \\
    WhatsUp & $+3.50\,/\,-0.86$ & $+8.02\,/\,-0.66$ & $+0.25\,/\,-0.76$ & $-0.10\,/\,+6.04$ & $+0.25\,/\,-3.70$ \\
    COCO-QA & $+4.76\,/\,-0.12$ & $+2.05\,/\,-1.04$ & $-0.12\,/\,-0.11$ & $+0.02\,/\,+0.91$ & $+0.09\,/\,-0.12$ \\
    VG-QA & $+6.06\,/\,+2.24$ & $+2.99\,/\,+0.49$ & $+0.09\,/\,+2.33$ & $+0.00\,/\,-1.49$ & $+0.00\,/\,+2.01$ \\
    \bottomrule
  \end{tabular}
\end{table*}

\FloatBarrier

\end{document}